\documentclass{article}

    \usepackage[preprint]{neurips_2025}

\usepackage{amsmath}
\usepackage{subcaption}
\usepackage[utf8]{inputenc} % allow utf-8 input
\usepackage[T1]{fontenc}    % use 8-bit T1 fonts
\usepackage{hyperref}       % hyperlinks
\usepackage{url}            % simple URL typesetting
\usepackage{booktabs}       % professional-quality tables
\usepackage{amsfonts}       % blackboard math symbols
\usepackage{nicefrac}       % compact symbols for 1/2, etc.
\usepackage{microtype}      % microtypography
\usepackage{xcolor}         % colors
\usepackage{cite}
\usepackage{graphicx}
\usepackage{mathrsfs}

\usepackage{wrapfig}

\usepackage{mathtools}
\usepackage{amssymb}
\usepackage{tabularray}
\DeclareMathAlphabet{\mathbbold}{U}{bbold}{m}{n}

\usepackage{algorithm}
\usepackage{algpseudocode}
\usepackage{amsmath, amssymb}
\usepackage{amsthm}
\usepackage{amsmath}

\usepackage{booktabs}   % for \toprule, \midrule, \bottomrule
\usepackage{makecell}   % for \makecell to wrap lines

\newtheorem{theorem}{Theorem}[section]

\newtheorem{lemma}[theorem]{Lemma}

\newtheorem{definition}[theorem]{Definition}

\title{A Spectral Framework for Evaluating Geodesic Distances Between Graphs}

\author{%
  Soumen Sikder Shuvo, Ali Aghdaei, Zhuo Feng \\
  Stevens Institute of Technology\\
  \texttt{\{sshuvo,aaghdae1,zhuo.feng\}@stevens.edu} \\
}

\begin{document}

\maketitle

\begin{abstract}
This paper presents a spectral framework for quantifying the differentiation between graph data samples by introducing a novel metric named Graph Geodesic Distance (GGD). For two different graphs with the same number of nodes, our framework leverages a spectral graph matching procedure to find node correspondence so that the geodesic distance between them can be subsequently computed by solving a generalized eigenvalue problem associated with their Laplacian matrices. For graphs of different sizes, a resistance-based spectral graph coarsening scheme is introduced to reduce the size of the larger graph while preserving the original spectral properties. We show that the proposed GGD metric can effectively quantify dissimilarities between two graphs by encapsulating their differences in key structural (spectral) properties, such as effective resistances between nodes, cuts, and the mixing time of random walks. Through extensive experiments comparing with state-of-the-art metrics, such as the latest Tree-Mover's Distance (TMD), the proposed GGD metric demonstrates significantly improved performance for graph classification, particularly when only partial node features are available. Furthermore, we extend the application of GGD beyond graph classification to stability analysis of GNNs and the quantification of distances between datasets, highlighting its versatility in broader machine learning contexts.
\end{abstract}

\section{Introduction}
In the era of big data, comparison and distinction between data points are important tasks. A graph is a specific type of data structure that represents the connections between a group of nodes. Comparing two graphs often involves using a pairwise distance measure, where a small distance indicates a high structural similarity and vice versa. To understand the generalization between distribution shifts, it is important to use an appropriate measure of divergence between data distributions, both theoretically and experimentally \citep{chuang2020estimating}. Determining suitable distance metrics for non-Euclidean data, such as graphs with or without node attributes, remains challenging. These metrics are fundamental to many graph learning methods, such as graph neural networks (GNNs), but are not as readily available as those for Euclidean space. The need to develop new analytical techniques that allow the visualization, comparison, and understanding of different graphs has led to a rich field of research study \citep{haslbeck2018well}. This study dives into the exploration of a novel framework for computing geometric distances between graphs, which can be immediately leveraged for many graph-based machine learning (ML) tasks, such as graph classification, dataset distance, or the stability evaluation of GNNs.

Many distance metrics for comparing graphs have previously been proposed \citep{MAL}. Some of them are based only on graph local structures \citep{tam2022multiscale, haussler1999convolution, xu2013distance, zhu2020fewer, fernandez2001graph, bunke1998graph}, whereas others exploit both graph structural properties and node attributes \citep{shervashidze2011weisfeiler, morris2019weisfeiler}. For example, the Graph Edit Distance (GED) has been proposed to measure the distance between graphs considering the number of changes needed to match one graph to another \citep{sanfeliu1983distancenew, gao2010survey,li2017comparative}; Distance metrics based on the graph kernel have also been investigated \citep{shervashidze2011weisfeiler, vishwanathan2010graph}, such as the Wasserstein Weisfeiler-Leman metric (WWL) \citep{morris2019weisfeiler} and the Gromov–Wasserstein metric \citep{memoli2011gromov}, which allow computing graph distances based on low-dimensional graph representations or optimal transport (OT) \citep{titouan2019optimal, chapel2020partial}, leading to the development of the state-of-the-art graph distance metric called TMD \citep{chuang2022tree}.

However, existing graph distance metrics have notable limitations. For example, the GED metric can capture local node or edge changes but struggles with global perturbations \citep{sanfeliu1983distancenew, gao2010survey,li2017comparative}; the WWL and TMD metrics heavily rely on node features (attributes) for calculating the distance between graphs, leading to degraded performance when only partial node features are available \citep{rossi2022unreasonable, chen2020learning}.

To address these limitations, we propose the Graph Geodesic Distance (GGD) metric—an end-to-end graph comparison framework that leverages spectral graph theory, structure-preserving coarsening, and Riemannian geometry to compute meaningful distances between graphs.

To address these limitations of prior methods, we propose the Graph Geodesic Distance (GGD) metric, a novel framework that combines spectral graph theory and Riemannian geometry to quantify topological distances between graphs. Unlike prior SPD-based works \citep{lim2019geometric}, our method operates directly on graph inputs, using spectral graph matching to establish node correspondence before embedding graphs into a Riemannian manifold of modified Laplacian matrices. This allows GGD to effectively capture key structural (spectral) dissimilarities between two graphs, such as mismatches in Laplacian
eigenvalues/eigenvectors, cuts, effective-resistance distances, etc.

% To address these limitations of prior methods, we propose the Graph Geodesic Distance (GGD) metric, a novel approach that leverages spectral graph theory and Riemannian geometry to effectively
% quantify topological distance between graphs. This framework handles graphs of the same size by using spectral graph matching to determine node correspondence and computes distances on a Riemannian manifold of modified Laplacian matrices. We show that the proposed GGD metric can theoretically
% capture key structural (spectral) dissimilarities between two graphs, such as mismatches in Laplacian
% eigenvalues/eigenvectors, cuts, effective-resistance distances, etc.

One distinct advantage of the proposed GGD metric is its capability to compute distances between graphs based on their spectral (structural) properties, while including node features can further improve its accuracy. This makes GGD suitable for analyzing real-world graphs with partial or no node features. Moreover, the proposed framework for computing GGDs is more computationally efficient than existing OT-based metrics, such as the TMD metric.

Our empirical results show that GGD can effectively measure the dissimilarities between graphs: (1) support vector classifiers (SVC) using GGDs perform competitively with state-of-the-art GNN models and graph kernels on graph classification benchmarks; (2) we demonstrate that the GGD metric allows us to quantify the stability of GNN models for graph classification tasks by checking whether two graphs with a small GGD will lead to a significant dissimilarity in the GNN output embeddings; (3) quantify distance between datasets to evaluate the transferability of domain knowledge. We also show that the GGD metric has a better correlation with established GNN outputs compared to the state-of-the-art TMD metric \citep{chuang2022tree} when only partial node features are available: up to a $10\%$ accuracy gain and a $9\times$ runtime speedup have been achieved in various graph classification tasks.

\section{Existing Graph Distance Metrics}
\textbf{Graph Edit Distance (GED)}\quad For non-attributed graph data, a common and simple distance metric is GED. \citep{sanfeliu1983distancenew, gao2010survey}. Given a set of graph edit operations, also known as elementary graph operations, the GED between two graphs \(G_1\) and \(G_2\), written as \(\text{GED}(G_1, G_2)\), can be defined as:
\vspace{-10pt}
\begin{equation}\label{equ:ged}
\text{GED}(G_1, G_2) = \min\limits_{\substack{(e_1, \ldots, e_k) \in \mathcal{P}(G_1, G_2)}} \sum_{i=1}^k c(e_i),
\end{equation}
where \(\mathcal{P}\left(G_1, G_2\right)\) denotes the set of edit operations transforming \(G_1\) into a graph isomorphism of \(G_2\), \(c(e_i)\) is the cost of edit operation \(e_i\). The set of elementary graph edit operators typically includes node insertion, node deletion, node substitution, edge insertion, edge deletion, and edge substitution.

\textbf{Tree Mover's Distance (TMD)}\quad TMD is a pseudometric for measuring distances between simple graphs, extending the concept of WWL to multisets of tree structures \citep{chuang2022tree}. By progressively adding neighboring nodes to the previous node at each level, we obtain the computation tree of a node. These tree structures are crucial in graph analysis \citep{weisfeiler1968reduction, pearson1905problem} and graph kernels \citep{ramon2003expressivity, shervashidze2011weisfeiler}. TMD uses hierarchical optimal transport (HOT) to analyze these computational trees from input graphs. For a graph \(G = (V, E)\) with node features \(f_v \in \mathbb{R}^s\) for node \(v \in V\), let \(T_v^1 = v\), and \(T_v^L\) be the depth-L computation tree of node \(v\). The multiset of these trees for \(G\) is \(T_G^L = \{T_v^L\}_{v \in V}\). The number and shape of trees must match to calculate optimal transport between two multisets of trees. If multisets are uneven, they are augmented with blank nodes. For multisets \(T_p\) and \(T_q\), the augmenting function \(\sigma\) adds blank trees to equalize their sizes. A blank tree \(T_\mathbb{O}\) has a single node with a zero vector feature \(\mathbb{O}_p \in \mathbb{R}^s\):
% \begin{equation}  \label{equ:tree}
% \begin{split}
% \sigma(T_p, T_q) \rightarrow & \Big(T_p \cup T_\mathbb{O}^{\max(|T_q| - |T_p|, 0)}, \\ 
% & T_q \cup T_\mathbb{O}^{\max(|T_p| - |T_q|, 0)}\Big)
% \end{split}
% \end{equation}

\begin{equation}  \label{equ:tree}
\sigma(T_p, T_q) \rightarrow \Big( T_p \cup T_\mathbb{O}^{\max(|T_q| - |T_p|, 0)},\; T_q \cup T_\mathbb{O}^{\max(|T_p| - |T_q|, 0)} \Big)
\end{equation}

Let \(X=\left\{x_i\right\}_{i=1}^k\) and \(Y=\left\{y_i\right\}_{j=1}^k\) be two data multisets and \(C \in \mathbb{R}^{k \times k}\) be the transportation cost for each data pair: \(C_{i j}=d\left(x_i, y_j\right)\), where \(d\) is the distance between \(x_i\) and \(y_j\). The unnormalized Optimal Transport between \(X\) and \(Y\) is defined as follows:
% \vspace{-15pt}
\begin{equation} \label{equ:ot}
\begin{array}{l}
\mathrm{OT}_d(X, Y):=\min _{\gamma \in \Gamma(X, Y)}\langle C, \gamma\rangle \\
\Gamma(X, Y)=\left\{\gamma \in \mathbb{R}_{+}^{m \times m} \mid \gamma \mathbbold{1}_m=\gamma^{\top} \mathbbold{1}_m=\mathbbold{1}_m\right\}.
\end{array}
\end{equation}
Here \(\Gamma\) is the set of transportation plans that satisfies the flow constrain \(\gamma \mathbbold{1}_m=\gamma^{\top} \mathbbold{1}_m=\mathbbold{1}_m\). \citep{chuang2022tree}. Now, the distance between two trees \(T_p\) and \(T_q\) with roots \(r_p\) and \(r_q\) is defined recursively:
\begin{equation}\label{equ:tmd}
\begin{aligned}
\text{TD}_w\left(T_p, T_q\right) := 
\begin{cases}
    \left\|f_{r_p}-f_{r_q}\right\|+w(L)\cdot \mathrm{OT}_{\mathrm{TD}_w}
    \left(\sigma\left(T_{r_p}, T_{r_q}\right)\right), \text{if } L>1 \\
    \left\|f_{r_p}-f_{r_q}\right\|,\quad \text{otherwise}
\end{cases}
\end{aligned}
\end{equation}

where \(L\) is the maximum depth of \(T_p\) and \(T_q\), and \(w\) is a depth-dependent weighting function.\\
Subsequently, the concept of distance from individual trees is extended to entire graphs. For graphs \(G_1\) and \(G_2\), with multisets \(\mathbf{T}_{G_1}^L\) and \(\mathbf{T}_{G_2}^L\) of depth-L computation trees, the Tree Mover's Distance is:
% \vspace{-8pt}
\begin{equation} 
\text{TMD}_w^L(G_1, G_2) = \text{OT}_{\mathrm{TD}_w}(\sigma(\mathbf{T}_{G_1}^L, \mathbf{T}_{G_2}^L)).
\end{equation}

\section{GGD:  Geodesic Distance Between  Graphs }
\paragraph{Modified Laplacian matrices on the Riemannian manifold} One way to represent a simple connected graph is through its Laplacian matrix, which is a Symmetric Positive Semidefinite (SPSD) matrix. Graph representation using adjacency and Laplacian matrices is briefly discussed in Appendix \ref{app:adj_lap}. Adding a small positive value to each diagonal element allows us to transform the original Laplacian matrix into a Symmetric Positive Definite (SPD) matrix, which is referred to as the \textbf{Modified Laplacian Matrix} in this work. In Appendix \ref{app:epsilon_ap}, we describe the effect of this small value on the GGD calculation. We can then consider the cone of such modified Laplacian matrices as a natural Riemannian manifold \citep{lim2019geometric}, where each modified Laplacian, having the same dimensions (same number of rows/columns), can be regarded as a data point on this Riemannian manifold \citep{vemulapalli2015riemannian, pennec2006riemannian}. Details about the Riemannian manifold are provided in Appendix \ref{App:Riemannain_manifold}. The geodesic distance is defined as the shortest path on the Riemannian manifold, providing a more appropriate comparison than Euclidean space \citep{lim2019geometric, crane2020survey, huang2015log}. We will later demonstrate (Section \ref{sec:mismatch}) that such a geodesic distance metric can effectively capture structural (spectral) mismatches between graphs.

% \subsection{Overview of the GGD Framework}
\begin{figure}[h]
  \centering
  \includegraphics[width=\linewidth]{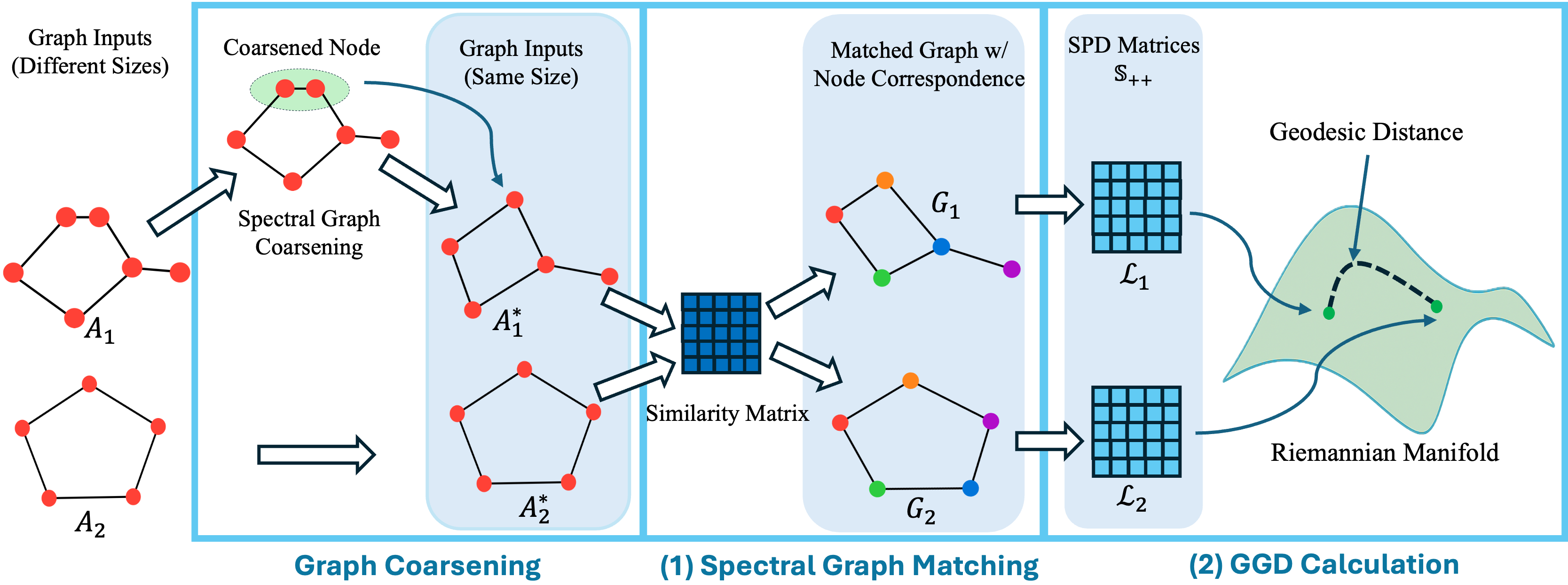}
  \caption{A high-level illustration of the GGD pipeline, including spectral graph coarsening, (Phase 1) spectral graph matching, and (Phase 2) geodesic distance calculation for the Riemannian manifold of modified Laplacian matrices.}
  \label{fig:overview}
\end{figure}

\textbf{A two-phase spectral framework for computing GGDs} Before computing GGDs, it is necessary to establish the node-to-node correspondence between two graphs. This can be achieved by leveraging existing graph-matching techniques \citep{livi2013graph, emmert2016fifty, caetano2009learning}. 
% In this work, we will leverage a recent spectral graph matching framework that has been shown to recover accurate matching with high probability \citep{fan2020spectral}.
The proposed GGD metric can be computed in the following two phases. \textbf{Phase 1} consists of a spectral graph matching step, using combinatorial optimization with the eigenvalues/eigenvectors of the graph adjacency matrices to identify the approximate node-to-node correspondence. \textbf{Phase 2} computes the GGD between the modified Laplacian matrices of the matched graphs by exploiting generalized eigenvalues. A high-level overview of the pipeline is illustrated in Figure \ref{fig:overview}, and a detailed algorithmic flow is provided in Appendix \ref{app:algo} to ensure a clear understanding of the process. The proposed GGD metric differs from previous OT-based graph distance metrics in its ability to accurately represent structural discrepancies between graphs, enabling us to uncover the topological variations between them more effectively.
% Since only the graph Laplacian (adjacency) matrix is required to calculate the GGD, our metric can even work effectively for graphs without node feature information.

\begin{wraptable}{r}{0.53\textwidth}
\centering
\caption{Normalized distance between graphs with simple perturbations.}
\label{g1g2g3}
\vskip -0.35in
\begin{center}
\begin{small}
\begin{sc}
% \begin{tabular}{ccccc}
\begin{tblr}{
  column{even} = {c},
  column{3} = {c},
  column{5} = {c},
  column{1} = {c},
  hline{1,5} = {-}{0.08em},
  hline{2} = {-}{0.05em},
}
% \toprule
{Graph\\Pairs} & GGD & {TMD w/\\NF, L = 4 } & {TMD w/o\\NF, L = 4 } & GED \\
% \midrule
$G_1, G_2$  & 0.623 & 0.689 & 0.970 & 1.000 \\
$G_1, G_3$  & 0.855 & 0.711 & 1.000 & 1.000 \\
$G_2, G_3$  & 1.000 & 1.000 & 0.333 & 1.000 \\
% \bottomrule
% \end{tabular}
\end{tblr}
\end{sc}
\end{small}
\end{center}
\vskip -0.3in
\end{wraptable}

\begin{wrapfigure}{r}{0.49\linewidth}
  \centering  
  \vspace{-20pt}
  \includegraphics[width=\linewidth]{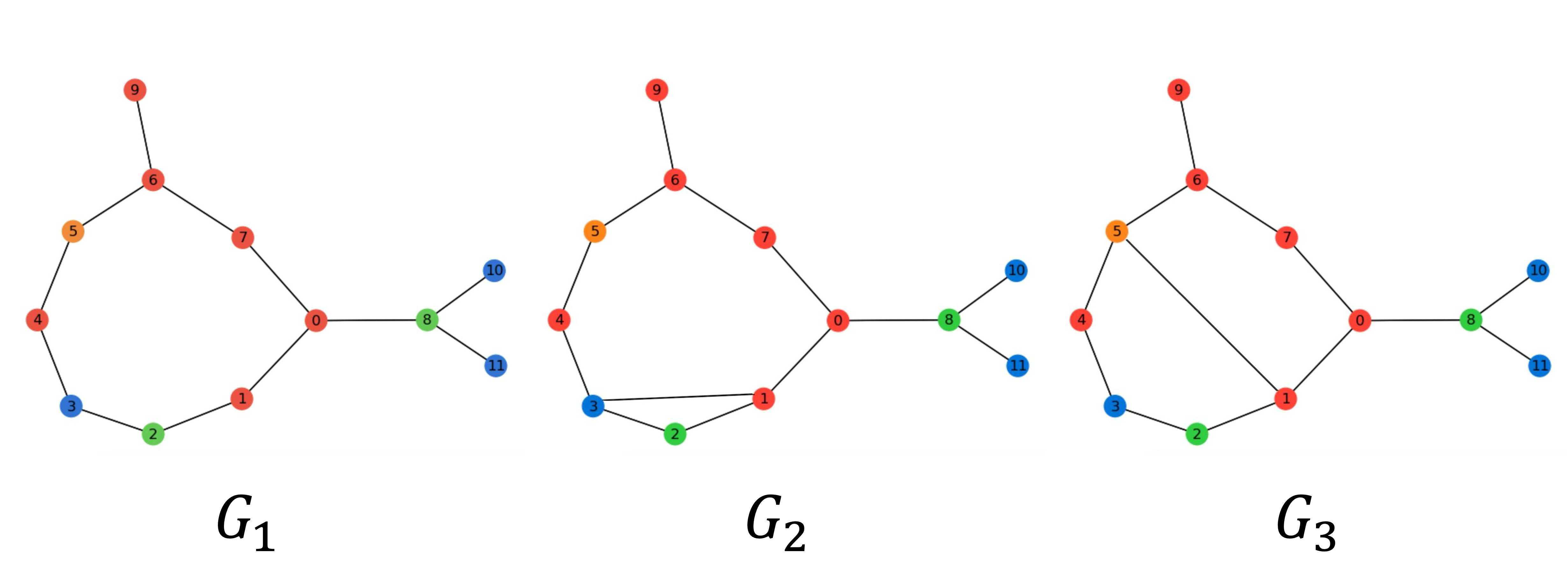}
  \caption{Graphs with simple perturbations}
  \vspace{-10pt}
  \label{g1g2g3_img}
\end{wrapfigure}

\textbf{A motivating example} Let's consider a simple graph \(G_1\), characterized by an almost ring-like topology, as shown in Figure \ref{g1g2g3_img}.  We also create two other graphs \(G_2\) and \(G_3\) by inserting an extra edge into \(G_1\) in different ways. Note that the additional edge in \(G_3\) will have a greater impact on \(G_1\)'s global structure since it connects two further nodes.

We compute the normalized distances (the largest distance always equals one) between the aforementioned three graphs using different metrics (GED, TMD, and GGD) and report the results in Table \ref{g1g2g3}. As observed, \(G_2\) and \(G_3\) have distances similar to \(G_1\) when the TMD metric is adopted without using node features (NFs). On the other hand, the TMD metric can produce similar results as the proposed GGD metric when node features are fully utilized. Not surprisingly, the GED always produces the same distances since only one edge has been added. The above results imply that the GED and TMD (without using NFs) metrics may not properly capture the dissimilarities in the spectral properties of the graphs.

\section{Computing GGDs Between Graphs of the Same Size}\label{sec:ggdsame}
\subsection{Phase 1: Spectral Graph Matching for Finding Node Correspondence}\label{sec:grampa}
Computing the GGD metric between two input graphs requires solving a graph-matching problem in advance., to ensure the minimum possible distance between modified Laplacian matrices.
% In calculating the GGD, we need to measure the geodesic distance between the modified Laplacian matrices of the graphs. 
% Without knowing the node-to-node correspondence that can be achieved through a graph-matching step, the distance between modified Laplacian matrices may be significantly higher than the minimum possible distance. 
In this work, we aim to find the infimum between two SPD matrices on the Riemannian manifold, which can be accomplished through a graph-matching phase. Graph matching techniques can be used to establish node-to-node correspondence by seeking a bijection between node sets to maximize the alignment of edge sets \citep{livi2013graph, emmert2016fifty, caetano2009learning}. This combinatorial optimization problem can be cast into a Quadratic Assignment Problem, which is NP-hard to solve or approximate \citep{fan2020spectral, wang2020learning}.

% \begin{figure}[ht]
% % \vskip 0.2in
% \begin{center}
% \centerline{\includegraphics[width=\columnwidth]{imgs/stiched.png}}
% \caption{Graphs with simple perturbations.}
% \label{g1g2g3_img}
% \end{center}
% \vskip -0.2in
% \end{figure}

In this study, we will exploit a spectral graph matching method called GRAMPA (\textbf{GRA}ph \textbf{M}atching by \textbf{P}airwise eigen-\textbf{A}lignments) \citep{fan2020spectral} to find the approximate node correspondence between two graphs. GRAMPA starts with comparing the eigenvectors of the adjacency matrices of the input graphs. Instead of comparing only the eigenvectors corresponding to the largest eigenvalues, it considers all pairs of eigenvectors/eigenvalues to generate a similarity matrix. This similarity matrix can be constructed by summing up the outer products of eigenvector pairs, weighted by a Cauchy kernel \citep{fan2020spectral}. Subsequently, a rounding procedure will be performed to determine the optimal match between nodes employing the similarity matrix.

\begin{definition}[Similarity Matrix]
Let \(G_1\) and \(G_2\) be two undirected graphs with \(n\) nodes, and let their weighted adjacency matrices be \(A_1\) and \(A_2\), respectively. The spectral decompositions of \(A_1\) and \(A_2\) are expressed as follows:
$A_1 = \sum_{i=1}^{n} \zeta_i u_i u_i^\top \quad \text{and} \quad A_2 = \sum_{j=1}^{n} \mu_j v_j v_j^\top$,
where the eigenvalues are ordered such that \(\zeta_1 \geq \ldots \geq \zeta_n\) and \(\mu_1 \geq \ldots \geq \mu_n\). The similarity matrix \(\widehat{X} \in \mathbb{R}^{n \times n}\) is defined as:

% \begin{equation} \label{similarity_matrix}
%     \widehat{X}=\sum_{i, j=1}^n w\left(\zeta_i, \mu_j\right) \cdot u_i u_i^{\top} \mathbf{J} v_j v_j^{\top} ,\quad\text{where} \quad w(x, y)=\frac{1}{(x-y)^2+\eta^2}
% \end{equation}
% \begin{equation} \label{similarity_matrix}
%     \widehat{X}=\sum_{i, j=1}^n w\left(\zeta_i, \mu_j\right) \cdot u_i u_i^{\top} \mathbf{J} v_j v_j^{\top} ,\text{where  } w(x, y)=\frac{1}{(x-y)^2+\eta^2}.
% \end{equation}

% \begin{align} 
%     \widehat{X} &= \sum_{i, j=1}^n w\left(\zeta_i, \mu_j\right) \cdot u_i u_i^{\top} \mathbf{J} v_j v_j^{\top}, \notag \\
%     w(x, y) &= \frac{1}{(x-y)^2+\eta^2}.
% \end{align}
\vspace{-12pt}
\begin{equation} \label{similarity_matrix}
\widehat{X} = \sum_{i, j=1}^n w\left(\zeta_i, \mu_j\right) \cdot u_i u_i^{\top} \mathbf{J} v_j v_j^{\top}, \quad
    w(x, y) = \frac{1}{(x-y)^2+\eta^2}.
\end{equation}
Here, $\mathbf{J} \in \mathbb{R}^{n \times n}$ denotes an all-one matrix and \(w\) is the Cauchy kernel of bandwidth \(\eta\).
\end{definition}

The permutation estimate matrix \(\widehat{\pi}\) can be obtained by rounding \(\widehat{X}\), typically achieved by solving the Linear Assignment Problem (LAP):
\begin{equation} \label{lap_similarity}
    % \widehat{\pi}=\underset{\pi \in \mathcal{S}_n}{\operatorname{argmax}} \sum_{i=1}^n \widehat{X}_{i, \pi(i)}
    \widehat{\pi}={\operatorname{argmax}} \sum_{i=1}^n \widehat{X}_{i, \pi(i)},
\end{equation}
% \vspace{-3pt}
which can be efficiently solved using the Hungarian algorithm \citep{fan2020spectral}. However, one simpler rounding procedure was advocated in \citep{fan2020spectral} with theoretical results supporting the rounding procedure,
which is given by the following equation:
% This LAP can be solved efficiently using the Hungarian Algorithm \citep{fan2020spectral, kuhn1955hungarian}, where:
\begin{equation} \label{lap_easy}
\widehat{\pi}(i)=\underset{j}{\operatorname{argmax}}  \widehat{X}_{i j},
\end{equation}

here the permutation estimate matrix is constructed by selecting the largest index from each row. While LAP provides optimal matching, its computational complexity can become expensive for very large graphs. By carefully choosing $\eta$, the same match recovery holds if rounding is performed using equation \ref{lap_similarity} instead of solving the LAP in equation \ref{lap_easy} \citep{fan2020spectral}.
\begin{lemma}[Graph Matching Recovery]
    Given symmetric matrices \(A_1\), \(A_2\) and \(Z\) from the Gaussian Wigner model, where \( A_{2 \pi^*} = A_1 + \sigma Z \), there exist constants \(c, c' > 0\) such that if \(1/n^{0.1} \leq \eta \leq c / \log n\) and \(\sigma \leq c' \eta\), then with probability at least $1 - n^{-4}$, GRAMPA Algorithm correctly recovers the permutation matrix \(\pi^*\) from the Similarity matrix \(\widehat{X}\) \citep{fan2020spectral}. Its proof can be found in the supporting documents \ref{grampa_proof}.
\end{lemma}
Once \(\widehat{\pi}\) is obtained, the best-matched mirrors of the input graphs are:
% \begin{equation} \label{match_a_b}
%     \textnormal{Best Match to } A_2 = \widehat{\pi}  A_1 
%  \widehat{\pi}^\top, \quad \textnormal{Best Match to } A_1 = \widehat{\pi}^\top  A_2  \widehat{\pi}.
% \end{equation}
\begin{equation}  \label{match_a_b}
\textnormal{Best Match to } A_2 = \widehat{\pi} A_1 \widehat{\pi}^\top, \quad
\textnormal{Best Match to } A_1 = \widehat{\pi}^\top A_2 \widehat{\pi}
\end{equation}

In practice, the graph matching performance is not too sensitive to the choice of tuning parameter $\eta$. For small-sized graphs, such as the MUTAG dataset\citep{morris2020tudataset}, setting $\eta = 0.5$ yields satisfactory results in matching. In \ref{eta_appendix}, the effect of \(\eta\) for computing GGDs has been comprehensively analyzed.
\subsection{Phase 2: Computing Geodesic Distances Between Graph Laplacians}
The GGD metric can be formally defined as the infimum length of geodesics connecting two data points in the Riemannian manifold formed by the cone of the modified graph Laplacian matrices \citep{lim2019geometric}. This distance metric can be imagined as a matrix representation of the geometric distance \(|\log(a/b)|\)  between two positive numbers $a,b$ \citep{bonnabel2010riemannian, shamai2017geodesic, owen2010fast}.

\begin{definition}[Graph Geodesic Distance]
    Let $\mathcal{L}_1$ and $\mathcal{L}_2 \in \mathbb{S}_{++}^n$ denote two modified Laplacian matrices corresponding to two matched graphs \(G_1\) and \(G_2\) both having \(n\) nodes, then their Graph Geodesic Distance   denoted by $GGD(G_1, G_2): \mathbb{S}_{++}^n \times \mathbb{S}_{++}^n \rightarrow \mathbb{R}_{+}$, is defined as:
    \begin{equation} \label{ggd}
        GGD(G_1, G_2) = \left[\sum_{i=1}^{n} \log^2(\lambda_i(\mathcal{L}_1^{-1}\mathcal{L}_2))\right]^{1/2},
    \end{equation}
where \(\lambda_i\) are the generalized eigenvalues computed with the matrix pencil ($\mathcal{L}_1$, $\mathcal{L}_2$).
\end{definition}

The above GGD formulation for computing distances between SPD matrices is based on an Affine-Invariant Riemannian Metric (AIRM) \citep{lim2019geometric}, while another well-known metric, the Log-Euclidean Riemannian Metric (LERM) \citep{ilea2018covariance, thanwerdas2023n, chen2024adaptive} is also discussed in Appendix \ref{app:ai_vs_le}.
% of the supplementary section.

\subsection{Connection between GGD and Graph Structural Mismatches}\label{sec:mismatch}

Consider two graphs, $G_1$ and $G_2$, that have the same node set $V$, with a known correspondence between their nodes. Let $L_1$ and $L_2$ be the Laplacian matrices of these graphs, respectively. Suppose we take a subset of nodes, denoted by $S$ and its complement, $S'$. We assign the value $1$ to the nodes in $S$ and the value $0$ to those in $S'$. This defines the set $S$ as:
\[ 
S \stackrel{\text{def}}{=} \{ v \in V : x(v) = 1 \}.
\]

% \begin{figure}[h]
% \vskip 0.2in
% \begin{center}
% \centerline{\includegraphics[width=0.75\columnwidth]{imgs/cmd_new.png}}
% \caption{The cut mismatch (for the node set $S$) between two simple graphs is $\frac{6}{2}=3$.}
% \label{cmd2}
% \end{center}
% \vskip -0.2in
% \end{figure}

For graph $G_1$, the cut for the node subset $S$ (which is the number of edges that cross between $S$ and $S'$) can be computed as: \(\text{cut}_{G_1}(S, S') = x^T L_1 x.\)

\begin{wrapfigure}{r}{0.4\linewidth}
% \vspace{-10pt}
  \centering
  \includegraphics[width=\linewidth]{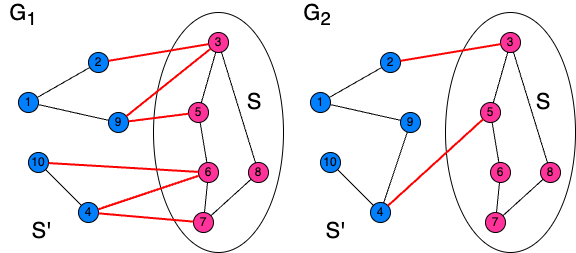}
  \caption{The cut mismatch (for the node set $S$) between two graphs is $\frac{6}{2}=3$}
  \label{cmd2}
\end{wrapfigure}

As shown in Figure \ref{cmd2}, for a node subset $S$, six edges cross between $S$ and $S'$ in graph $G_1$, whereas two edges cross in graph $G_2$. This ratio of edge counts in the two graphs is referred as a cut mismatch. The relationship between this cut mismatch and the generalized eigenvalue problem for the matrix pair $(L_1, L_2)$ can be formalized using the Generalized Courant-Fischer Minimax Theorem \citep{golub2013matrix, feng2020grass}.

\begin{lemma}[The Generalized Courant-Fischer Minimax Theorem]
Given two Laplacian matrices $L_1, L_2 \in \mathbb{R}^{n \times n}$ such that $\operatorname{null}\left(L_2\right) \subseteq \operatorname{null}\left(L_1\right)$, the $k$-th largest generalized eigenvalue of $L_1$ and $L_2$ can be computed as follows for $1 \leq k \leq \operatorname{rank}\left(L_2\right)$:
\begin{equation} \label{general_cf_minmax}
\lambda_k=\min _{\substack{\operatorname{dim}(U)=k \\ U \perp \operatorname{null}\left(L_2\right)}} \max _{x \in U} \frac{x^{\top} L_1 x}{x^{\top} L_2 x}.
\end{equation}
\end{lemma}

This theorem provides a method for bounding the maximum cut mismatch between two graphs by calculating the largest generalized eigenvalue. Specifically, we can use the following optimization problem to compute the dominant eigenvalue $\lambda_{max}$ \citep{feng2020grass}:
% \begin{equation} \label{dominant_eig_val}
%     \lambda_{\max }=\max _{\substack{|x| \neq 0 \\ x^{\top} \mathbb{1}=0}} \frac{x^{\top} L_1 x}{x^{\top} L_2 x} \geq \max _{\substack{|x| \neq 0 \\ x(v) \in\{0,1\}}} \frac{x^{\top} L_1 x}{x^{\top} L_2 x} = \max \frac{cut_{G_1}(S, S')}{cut_{G_2}(S, S')}.
% \end{equation}
% \begin{align} \label{dominant_eig_val}
%     \lambda_{\max } &= \max _{\substack{|x| \neq 0 \\ x^{\top} \mathbb{1}=0}} 
%     \frac{x^{\top} L_1 x}{x^{\top} L_2 x} \notag \geq \max _{\substack{|x| \neq 0 \\ x(v) \in\{0,1\}}} 
%     \frac{x^{\top} L_1 x}{x^{\top} L_2 x} \notag \\
%      &= \max \frac{cut_{G_1}(S, S')}{cut_{G_2}(S, S')}.
% \end{align}

\begin{equation} \label{dominant_eig_val}
\lambda_{\max} = \max_{\substack{|x| \neq 0 \\ x^{\top} \mathbb{1} = 0}} \frac{x^{\top} L_1 x}{x^{\top} L_2 x} \geq \max_{\substack{|x| \neq 0 \\ x(v) \in \{0,1\}}} \frac{x^{\top} L_1 x}{x^{\top} L_2 x} = \max \frac{\mathrm{cut}_{G_1}(S, S')}{\mathrm{cut}_{G_2}(S, S')}
\end{equation}

From equation (\ref{dominant_eig_val}), we can see that the dominant generalized eigenvalue $\lambda_{max}$ corresponds to the most significant cut mismatch between $G_1$ and $G_2$. In particular, $\lambda_1 = \lambda_{max}$ sets an upper bound on the cut mismatch between $G_1$ and $G_2$, while $\lambda_n = \lambda_{min}$ defines the upper bound of the mismatch in the reverse direction, between $G_2$ and $G_1$. Appendix \ref{app:eig_vs_cut} illustrates this relationship with practical examples. Additionally, we illustrate the relationship between the approximate GGD values with extreme eigenvalues, compared with the accurate GGD values in Appendix \ref{ap:extre_eig}.

\section{Computing GGDs for Graphs with Different Sizes} 
\paragraph{Submatrix selection methods} To calculate geodesic distances between SPD matrices of different sizes, prior studies have proposed a submatrix adaptation method \citep{lim2019geometric}. In this approach, a principle submatrix with the same size as the smaller matrix is obtained from the larger matrix \citep{ye2016schubert}, and subsequently used to calculate the GGD. Furthermore, this method can be extended to project the smaller matrix into a larger one with the same size as the larger matrix \citep{lim2019geometric}. While these methods are efficient for handling SPD matrices, for our application taking the submatrix of the modified Laplacian can lose important nodes/edges, compromising critical graph structural properties.
\vspace{-12pt}
\paragraph{Graph coarsening methods} 
% In this work, we will leverage spectral graph coarsening to address the issue. 
Spectral graph coarsening is a widely adopted process \citep{loukas2019graph, aghdaei2022hyperef} for reducing graph sizes while preserving key spectral (structural) properties, such as the Laplacian eigenvalues/eigenvectors. Recent spectral graph coarsening methods aim to decompose an input graph into many distinct node clusters, so that a reduced graph can be formed by treating each node cluster as a new node, with a goal of assuring that the reduced graph will approximately retain the original graph's structure \citep{loukas2019graph, han2024topology, aghdaei2022hyperef}. 
% Given our need to maintain the eigenvalues of matrix representations, spectral methods are particularly relevant for us. 
Therefore, when computing GGDs for graphs of different sizes, we can first adopt spectral graph coarsening to transform the bigger graph into a smaller one, so that our framework in Section \ref{sec:ggdsame} can be subsequently utilized.
However, existing state-of-the-art graph coarsening methods do not allow us to precisely control the size of the reduced graphs. 

\vspace{-5pt}
\subsection{Our Approach: Spectral Graph Coarsening by Effective Resistances}

In this work, we exploit a spectral graph coarsening method leveraging effective-resistance clustering \citep{aghdaei2022hyperef}, specifically designed to reduce graph size while preserving key spectral characteristics. Unlike prior work \citep{lim2019geometric}, which addresses dimension mismatch directly in SPD space, our coarsening method operates at the structural level before transformation into SPD matrices, making it more suitable for computing distances between graphs with unequal sizes. Our approach begins by estimating the effective resistances of all edges in the original graph. If node features are available, we also incorporate feature differences as an additional parameter. During coarsening, edges are ranked by resistance distance, and only the top few edges with the smallest effective resistances are merged into new nodes. This strategy enables precise control over the size of the reduced graph while preserving crucial structural properties, such as the eigenvalues and eigenvectors of the adjacency matrices—essential for the subsequent spectral graph matching step (Phase 1 in Section \ref{sec:grampa}).

% In this work, we introduce a spectral graph coarsening method using effective-resistance clustering \citep{aghdaei2022hyperef}. Our approach starts with estimating the effective resistances of all edges in the original graph. We can also incorporate the difference between node features (if available) as an additional parameter. In the graph coarsening phase, our method will rank edges according to their resistance distances and only the top few edges with the smallest resistances will be coarsened into new nodes. This approach enables precise control over the size of the reduced graphs while preserving crucial structural properties, such as the eigenvalues/eigenvectors of the adjacency matrices, which are essential for the subsequent spectral graph matching step (Phase 1 in Section \ref{sec:grampa}).

Consider a connected, weighted, undirected graph $G = (V, E, w)$ with $|V| = n$. The effective resistance between nodes $(p, q) \in V$ plays a crucial role in various graph analysis tasks including spectral sparsification algorithms \citep{spielman2011spectral}. The effective resistance distances can be accurately computed using the equation:
\begin{equation}\label{eq:eff_resist0}
R_{eff}(p,q) = \sum\limits_{i= 2}^{n} \frac{(u_i^\top b_{pq})^2}{\sigma_i},
\end{equation}
where ${b_{p}} \in \mathbb{R}^{n}$ denote the standard basis vector with all zero entries except for the $p$-th entry being $1$, and ${b_{pq}}=b_p-b_q$. $u_{i} \in \mathbb{R}^{n}$ for $i=1, \dots, n$ denote the  unit-length, mutually-orthogonal eigenvectors corresponding to Laplacian eigenvalues $\sigma_i$ for $i=1, \dots, n$. Background on effective resistance is presented in Appendix \ref{app:effective_resistance}, with its estimation detailed in Appendix \ref{app:estimatation_eff_res}.
% \vspace{-10pt}

\vspace{-5pt}
\section{GGD as a Distance Metric}
Assuming the graph matching problem can always find the exact correspondence between nodes, we prove that the GGD metric (based on AIRM) between any two nonempty graphs is a metric that satisfies the following conditions:
\begin{itemize}
    \item The distance between a graph and itself or between two isomorphic graphs is zero: \( GGD(G, G) = 0 \).
    \item (Positivity) The distance between two distinct graphs is positive: \( GGD(G_1, G_2) \geq 0 \).
    \item (Symmetry) The distance between $G_1$ and $G_2$ is the same of the one between $G_2$ and $G_1$: \( GGD(G_1, G_2) = GGD(G_2, G_1) \).
    \item The triangle inequality: \(GGD(G_1, G_3) \leq GGD(G_1, G_2) + GGD(G_2, G_3)\).
\end{itemize}
Detailed proofs of the above four properties are provided in Appendix \ref{ggdproofs}.

\section{Experiments}

\subsection{Application of GGDs in GNN  Stability Analysis}

\begin{figure}
  \centering
  \includegraphics[width=\linewidth]{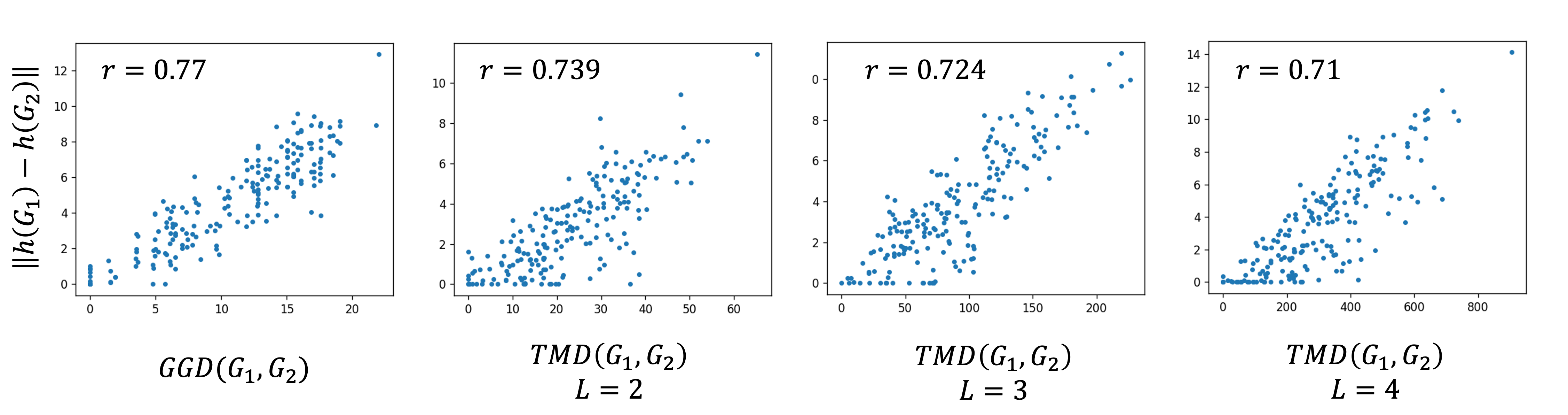}
  \caption{Correlation between graph distance metrics and GNN model outputs.}
  \label{corr_riem_tmd}
  \vspace{-10pt}
\end{figure}

To analyze the stability of GNN models \citep{bronstein2017geometric, garg2020generalization, duvenaud2015convolutional}, we conducted multiple experiments with the GGD and TMD metrics. GNNs typically operate by a message-passing mechanism \citep{gilmer2017neural}, where at each layer, nodes send their feature representations to their neighbors. The feature representation of each node is initialized to its original features and is updated by repeatedly aggregating incoming messages from neighbors. In our experiment, we relate GGD to the Graph Isomorphism Networks (GIN) \citep{xu2018how}, one of the most widely applied and powerful GNNs, utilizing the MUTAG dataset \citep{morris2020tudataset} as our reference graph dataset. The objective is to analyze the relationship between the input distance $GGD(G_1, G_2)$ and the distance between the output GIN vectors, $\| h(G_1) - h(G_2) \|$ for randomly selected pairs of graphs. 

We employed a three-layer GIN network as described in \citep{xu2018how}. This network uses GIN convolutional layers to update tensors of nodes based on their neighboring nodes and then aggregates those outputs in a vector representation, followed by linear layers for classification tasks. Thus it outputs a single vector $h(G)$ for the entire graph $G$. The result is illustrated in Figure \ref{corr_riem_tmd}.

We observe a strong correlation between GGD and the output distance, as indicated by a high Pearson correlation coefficient. This finding implies the effectiveness of the proposed GGD metric for analyzing the stability of GNN models \citep{chuang2022tree}. To compare GGD with existing metrics, we repeat this experiment using TMD without considering node attributes (features). As shown in Figure \ref{corr_riem_tmd}, GGD demonstrates a better correlation with GIN outputs than the TMD metric across different levels. These findings indicate that when dealing with graphs without node features, GGD should be adopted for the stability analysis of graph learning models. The performance of GGD under partially missing node features is further discussed in Appendix \ref{app:partial_nf}.

\subsection{Application of GGDs in Graph Classification Tasks}

\begin{table*}[t]
\caption{Classification accuracies for various models on graph datasets.}
\label{accuracy_table}
\centering
\setlength{\tabcolsep}{3.8pt} % Adjust column spacing
\renewcommand{\arraystretch}{0.95} % Adjust row spacing
\begin{footnotesize} % Slightly smaller font than \begin{small}
\begin{sc}
\begin{tabular}{lcccc}
\toprule
Dataset & MUTAG & PC-3H & SW-620H & BZR \\
\midrule
GGD & \underline{86.24$\pm$7.89} & \textbf{78.34$\pm$1.60} & \textbf{77.6$\pm$3.50} & \textbf{83.23$\pm$6.25} \\
TMD, L = 2 & 76.19$\pm$5.26 & -- & -- & -- \\
TMD, L = 3 & 77.34$\pm$5.26 & 71.24$\pm$2.45 & 70.22$\pm$2.29 & 73.43$\pm$2.44 \\
TMD, L = 4 & 78.20$\pm$5.26 & 71.37$\pm$1.42 & 70.84$\pm$2.29 & 73.96$\pm$4.88 \\
TMD, L = 5 & 78.20$\pm$5.26 & 71.89$\pm$2.40 & 71.20$\pm$1.88 & 75.13$\pm$2.44 \\
GCN~\citep{Kipf:2016tc} & 77.37$\pm$3.95 & 70.56$\pm$1.66 & 69.44$\pm$0.94 & 72.56$\pm$3.66 \\
GIN~\citep{xu2018how} & 82.60$\pm$4.60 & \underline{75.34$\pm$1.10} & 73.36$\pm$2.32 & \underline{77.09$\pm$3.66} \\
DGCNN~\citep{dgcnnzhang2018end} & 76.66$\pm$3.19 & 73.79$\pm$0.75 & \underline{74.37$\pm$1.54} & 72.38$\pm$1.08 \\
WWL~\citep{togninalli2019wasserstein} & 72.39$\pm$2.63 & 65.46$\pm$1.11 & 68.06$\pm$0.86 & 72.37$\pm$1.22 \\
WL Subtree~\citep{shervashidze2011weisfeiler} & 76.81$\pm$6.30 & 68.43$\pm$0.76 & 69.36$\pm$1.20 & N/A \\
FGW~\citep{titouan2019optimal} & \textbf{88.33$\pm$5.26} & 61.77$\pm$1.11 & 58.28$\pm$1.02 & 51.03$\pm$2.63 \\
\bottomrule
\end{tabular}
\end{sc}
\end{footnotesize}
\end{table*}
% \vspace{-20pt}

% \begin{table*}
% \caption{Classification accuracies for various models on graph datasets.}
% \label{accuracy_table}
% \centering
% \begin{small}
% \begin{sc}
% \begin{tabular}{lcccc}
% \toprule
% Dataset & MUTAG & PC-3H & SW-620H & BZR \\
% \midrule
% GGD & \underline{86.24$\pm$7.89} & \textbf{78.34$\pm$1.60} & \textbf{77.6$\pm$3.50} & \textbf{83.23$\pm$6.25} \\
% TMD, L = 2 & 76.19$\pm$5.26 & -- & -- & -- \\
% TMD, L = 3 & 77.34$\pm$5.26 & 71.24$\pm$2.45 & 70.22$\pm$2.29 & 73.43$\pm$2.44 \\
% TMD, L = 4 & 78.20$\pm$5.26 & 71.37$\pm$1.42 & 70.84$\pm$2.29 & 73.96$\pm$4.88 \\
% TMD, L = 5 & -- & 71.89$\pm$2.40 & 71.20$\pm$1.88 & 75.13$\pm$2.44 \\
% GCN~\citep{Kipf:2016tc} & 77.37$\pm$3.95 & 70.56$\pm$1.66 & 69.44$\pm$0.94 & 72.56$\pm$3.66 \\
% GIN~\citep{xu2018how} & 82.60$\pm$4.60 & \underline{75.34$\pm$1.10} & 73.36$\pm$2.32 & \underline{77.09$\pm$3.66} \\
% DGCNN~\citep{dgcnnzhang2018end} & 76.66$\pm$3.19 & 73.79$\pm$0.75 & \underline{74.37$\pm$1.54} & 72.38$\pm$1.08 \\

% WWL~\citep{togninalli2019wasserstein} & 72.39$\pm$2.63 & 65.46$\pm$1.11 & 68.06$\pm$0.86 & 72.37$\pm$1.22 \\
% WL Subtree~\citep{shervashidze2011weisfeiler} & 76.81$\pm$6.30 & 68.43$\pm$0.76 & 69.36$\pm$1.20 & N/A \\
% FGW~\citep{titouan2019optimal} & \textbf{88.33$\pm$5.26} & 61.77$\pm$1.11 & 58.28$\pm$1.02 & 51.03$\pm$2.63 \\
% \bottomrule
% \end{tabular}
% \end{sc}
% \end{small}
% \end{table*}

We evaluate whether the GGD metric aligns with graph labels in graph classification tasks using datasets from TUDatasets \citep{morris2020tudataset}. We employ a Support Vector Classifier (SVC) $(C=1)$ with an indefinite kernel $e^{-\gamma * GGD(G_1, G_2)}$, which can be viewed as a noisy observation of the true positive semidefinite kernel \citep{luss2007support}. The parameter $\gamma$ is selected through cross-validation from the set \{0.01, 0.05, 0.1\}. For comparative analysis with existing methods, we include graph kernels based on graph subtrees: the WL subtree kernel \citep{shervashidze2011weisfeiler}; and two widely adopted GNNs: graph isomorphism network (GIN) \citep{xu2018how} and graph convolutional networks (GCN) \citep{Kipf:2016tc}.

Table \ref{accuracy_table} presents the mean and standard deviation over five independent trials with a 90\%-10\% train-test split. For most cases, GGD consistently outperforms the performance of state-of-the-art GNNs, graph kernels, and metrics when node attributes are missing. Additionally, we observe that GGD allows us to obtain better results for larger datasets than smaller ones.
\vspace{-15pt}
\subsection{Application of GGDs in Dataset Distance}
To extend the application of GGD beyond structured graph datasets, we explore its utility in measuring distances between datasets. In transfer learning, finding the distance between datasets helps to quantify how similar or dissimilar the source and target domains are, guiding the adaptation of knowledge from one domain to another \citep{otdd}. We calculated this distance by first converting standard datasets into graph representations and then computing distances between these graphs using GGD framework.

To construct a graph from a dataset, we treat data points as nodes and establish edges based on a k-nearest neighbor (k-NN) approach. Nodes representing similar data points are connected, forming an initial dense graph. To further refine the structure, we apply a spectral graph sparsification procedure to reduce the number of edges while preserving key connectivity properties. The process of creating graphs from the dataset is further explained in Appendix \ref{app:datasettograph}.

Once the graphs are obtained, we apply GGD framework to compute pairwise distances. These distances are then compared with established dataset distance measures based on Optimal Transport \citep{otdd}. We also explore how these distances correlate with dataset \textit{Transferability}, which is the improvement in model performance when pretrained on a source dataset and fine-tuned on a target dataset. Transferability $\mathcal{T}$ of a source domain $\mathcal{D}_S$ to a target domain $\mathcal{D}_T$ is defined as the relative decrease in classification error when adapting compared to training only on the target domain \citep{otdd}.
\vspace{-5pt}
\begin{equation} \label{transferability}\mathcal{T}\left(\mathcal{D}_S \rightarrow \mathcal{D}_T\right)=100 \times \frac{\operatorname{error}\left(\mathcal{D}_S \rightarrow \mathcal{D}_T\right)-\operatorname{error}\left(\mathcal{D}_T\right)}{\operatorname{error}\left(\mathcal{D}_T\right)}.
\end{equation}

\begin{wrapfigure}{r}{0.55\linewidth}
\vspace{-22pt}
  \centering
  \includegraphics[width=\linewidth]{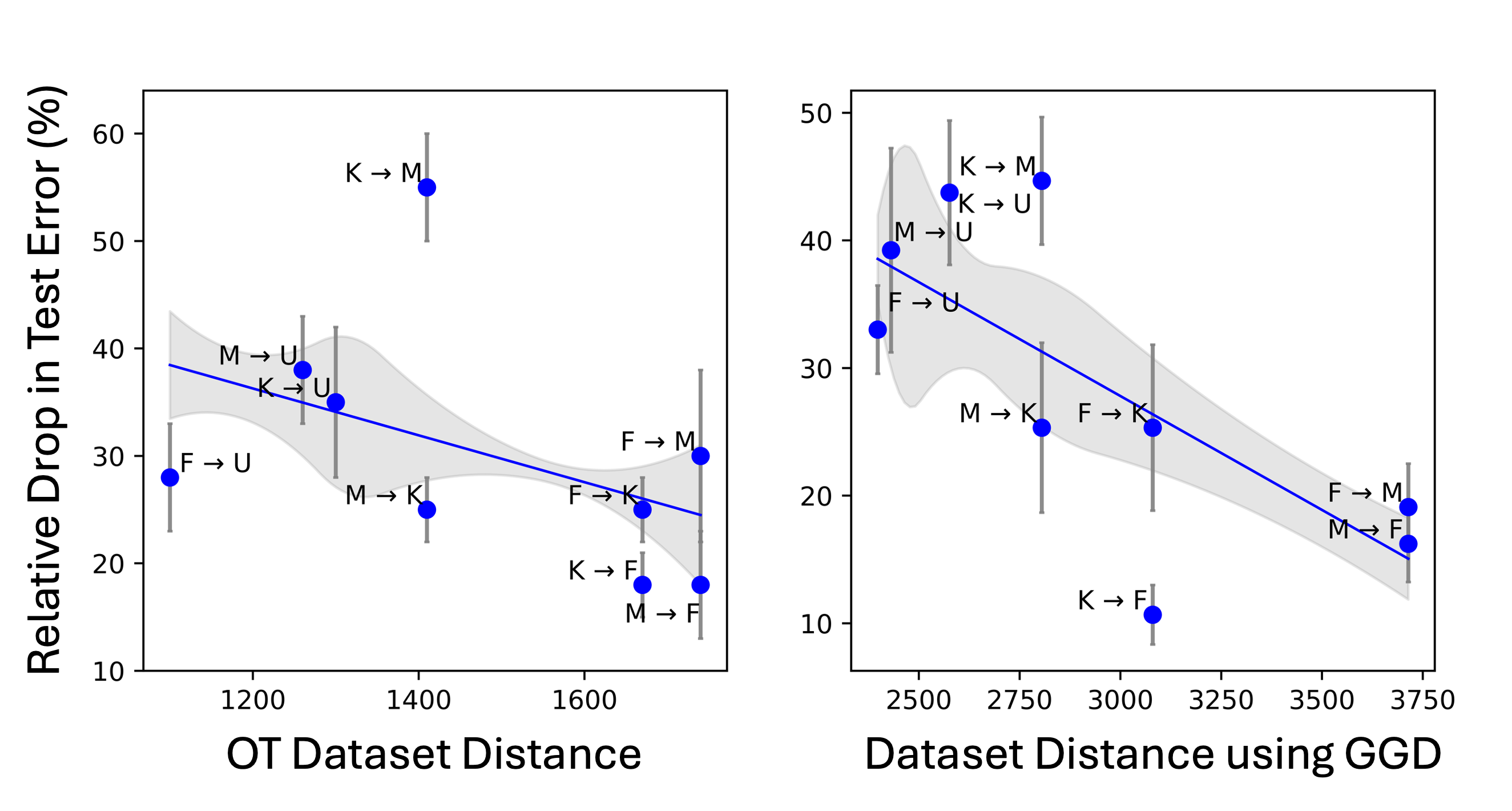}
  \caption{Dataset distance vs. Adaptation for *NIST datasets  (M: MNIST, K: KMNIST, F: Fashion-MNIST, U:
USPS).}
  \label{img:dataset_distance}
\end{wrapfigure}

Our results indicate that GGD-based dataset distances provide meaningful insights into dataset relationships and transfer learning ability, matching the performance of conventional distance metrics, demonstrated in Figure \ref{img:dataset_distance}.

\subsection{Runtime Complexity Analysis and Comparison}
When comparing various graph distance metrics, a primary consideration is their computational complexity. Conventional approaches usually require intricate computations that frequently have cubic or higher complexities. For our problem, the spectral graph matching step requires the eigenvalue decomposition of adjacency matrices and solving the linear assignment problem (LAP). Eigenvalue decomposition of an \(n \times n\) matrix has a complexity of \(O(n^3)\) \citep{borodin1975computational, flamary2021pot}, while solving the LAP using the Hungarian algorithm also has a runtime complexity of \(O(n^3)\). Similarly, calculating the generalized eigenvalue of two SPD matrices entails a cubic complexity. Consequently, the overall complexity of GGD calculation is \(O(n^3)\). On the other hand, TMD is an OT-based distance metric with a complexity of \(O(n^3\log(n))\) \citep{chuang2022tree, flamary2021pot}. Therefore, GGD exhibits slightly better (lower) runtime complexity than TMD.

% \vspace{-5pt}
% \begin{wraptable}[11]{r}{0.55\textwidth}
% \centering
% \caption{Runtime comparison for different distance metrics on various datasets.}
% \label{runtimes}
% % \vskip -0.3in
% \begin{center}
% \begin{small}
% \begin{sc}
% \begin{tblr}{
%   rowsep=0pt,
%   colsep=3pt,
%   column{1} = {l},
%   column{2} = {c},
%   column{3} = {c},
%   column{4} = {c},
%   column{5} = {c},
%   hline{1,6} = {-}{0.08em},
%   hline{2} = {-}{0.05em},
% }
%  & MUTAG & PC-3H & SW-620H & BZR \\
% GGD & \textbf{4.87s} & \textbf{31.89s} & \textbf{45.37s} & \textbf{5.80s} \\
% TMD, L = 3 & 5.29s & 88.60s & 98.69s & 7.22s \\
% TMD, L = 4 & 7.90s & 112.12s & 134.38s & 10.34s \\
% TMD, L = 6 & 11.27s & 273.31s & 287.92s & 14.98s \\
% \end{tblr}
% \end{sc}
% \end{small}
% \end{center}
% % \vskip -0.025in
% \end{wraptable}

\begin{wraptable}{r}{0.53\textwidth}
\centering
\caption{Runtime comparison for different distance metrics on various datasets.}
\label{runtimes}
% \vskip -0.3in
\begin{center}
\begin{small}
\begin{sc}
\begin{tblr}{
  rowsep=0pt,
  colsep=3pt,
  column{1} = {l},
  column{2} = {c},
  column{3} = {c},
  column{4} = {c},
  column{5} = {c},
  hline{1,6} = {-}{0.08em},
  hline{2} = {-}{0.05em},
}
 & MUTAG & PC-3H & SW-620H & BZR \\
GGD & \textbf{4.87s} & \textbf{31.89s} & \textbf{45.37s} & \textbf{5.80s} \\
TMD, L = 3 & 5.29s & 88.60s & 98.69s & 7.22s \\
TMD, L = 4 & 7.90s & 112.12s & 134.38s & 10.34s \\
TMD, L = 6 & 11.27s & 273.31s & 287.92s & 14.98s \\
\end{tblr}
\end{sc}
\end{small}
\end{center}
% \vskip -0.025in
\end{wraptable}
To evaluate runtime performance, we compare GGD and TMD on both small graphs (MUTAG, BZR) and large graphs (PC-3H, SW-620H) from the TUDataset \citep{morris2020tudataset}. Table \ref{runtimes} reports the average time (in seconds) to compute $100$ pairwise distances, averaged over five runs. GGD consistently outperforms TMD across all datasets, especially on larger graphs with more nodes. Since TMD requires deeper levels to capture structural information, its runtime increases rapidly. In contrast, GGD is approximately 6-9 times faster, making it significantly more efficient, particularly on large graphs. More details are provided in Appendix \ref{exp_set} and Appendix \ref{app:scalability}.

\section{Limitations}

Our current pipeline has three primary limitations. First, computationally, the end-to-end complexity is dominated by eigenvalue decomposition and the linear assignment step (both $O(n^3)$). Although this is comparable to or slightly better than some OT-based alternatives, scaling to very large graphs remains nontrivial. Appendix \ref{app:scalability} shows that using a small fraction of extreme eigenvalues preserves a high correlation with exact GGD and greatly reduces computation time, but this remains an approximation. Second, GGD is most reliable when comparing graphs of comparable size. When sizes differ dramatically, the size matching via graph coarsening introduces additional approximation error, though we expect such extreme mismatches to be rare in practical applications. Finally, like many graph distance metrics, our theory and experiments assume simple, undirected graphs. Extending GGD to directed, attributed, or higher-order graphs is a valuable direction for future work. We include these limitations to clarify the intended scope and reliability of GGD in practice and to guide when approximations or extensions may be required.

\section{Conclusion}

In this work, we introduce Graph Geodesic Distance (GGD), a novel spectral graph distance metric based on graph matching and the infimum on a Riemannian manifold. GGD captures key structural mismatches crucial for graph classification tasks. We also demonstrate that GGD is an effective metric for analyzing GNN model stability and graph classification, delivering superior performance even with partial node features. Additionally, GGD can evaluate transferability by calculating distances between datasets.

\newpage

\bibliographystyle{unsrt}
\bibliography{example_paper}

%%%%%%%%%%%%%%%%%%%%%%%%%%%%%%%%%%%%%%%%%%%%%%%%%%%%%%%%%%%%

\newpage

\appendix
\section{Appendix}

% \subsection{Overview of the GGD Framework}
% \begin{figure}[H]
%   \centering
%   \includegraphics[width=\linewidth]{imgs/ggd_f.png}
%   \caption{A high-level illustration of the GGD pipeline, including spectral graph coarsening, spectral graph matching, and geodesic distance computation on the Riemannian manifold of modified Laplacian matrices.}
%   \label{fig:overview}
% \end{figure}

\subsection{Algorithm Flow} \label{app:algo}

\begin{algorithm}[h!]
\small
\caption{GGD: Geodesic Graph Distance}
\label{alg:ggd}
\begin{algorithmic}[1]
\State \textbf{Input:} Graphs $G_1 = (V_1, E_1, w_1)$, $G_2 = (V_2, E_2, w_2)$, tuning parameter $\eta > 0$, small diagonal value $0 < \epsilon \ll 1$, node feature weight $\alpha$
\State \textbf{Output:} GGD Value
\State Compute the adjacency matrices $A_1$, $A_2$
\If{$\text{shape}(A_1) \neq \text{shape}(A_2)$}
    \State Assign the larger graph to $G_1$, and the smaller graph to $G_2$
    \While{$\text{shape}(A_1) \geq \text{shape}(A_2)$}
        \State Compute the effective resistance $R_{\text{eff}}(p,q)$ of each edge $(p, q) \in E_1$
        \State Compute the modified effective resistance $R^*_{\text{eff}}(p, q) = R_{\text{eff}}(p,q) + \alpha \lVert NF_p - NF_q \rVert$
        \State Coarsen the edge with the lowest $R^*_{\text{eff}}(p, q)$
        \State Update $A_1$
    \EndWhile
\EndIf
\State Compute eigenvectors $u_i$, $v_i$ and eigenvalues $\zeta_i$, $\mu_i$ of $A_1$ and $A_2$, respectively
\State Compute the similarity matrix $\hat{X} \in \mathbb{R}^{n \times n}$
\State Solve Linear Assignment Problem to compute the permutation estimate matrix $\hat{\pi}$
\State Update $A_2$ by multiplying with $\hat{\pi}$ to get best match with $A_1$
\State Derive $L_1$ and $L_2$ from $A_1$ and $A_2$
\State Add $\epsilon$ to diagonal values of $L_1$ and $L_2$
\State Compute GGD value using the generalized eigenvalues
\State \Return GGD
\end{algorithmic}
\end{algorithm}

\subsection{Graph Adjacency and Laplacian Matrices} \label{app:adj_lap}
For an undirected graph $G=(V, E, w)$, where $V$ represents the set of nodes (vertices), $E$ represents the set of edges, and $w$ denotes the associated edge weights, the adjacency matrix $A$ is defined as follows:
\begin{equation} \label{equ_adjacency}
    A(i, j)= \begin{cases}w(i, j), & \text { if }(i, j) \in E. \\ 0, & \text { otherwise. }\end{cases}
\end{equation}

Let $D$ denote the diagonal matrix where $D(i, i)$ is equal to the (weighted) degree of node $i$. The graph Laplacian matrix is then given by $L=D-A$.  
The rank of the Laplacian matrix of a graph $G$ is $n - c(G)$, where $n$ is the number of nodes and $c(G)$ is the number of connected components in the graph. For a connected graph, this implies that the rank of the Laplacian matrix is $n-1$, meaning Laplacian matrices are not full-rank \citep{bondy1976graph}.

\subsection{Riemannian Manifold} \label{App:Riemannain_manifold}

% A Riemannian manifold is a smooth manifold whose inner product on the tangent space for a point p varies smoothly. In other words, the Riemannian Manifold is a smooth, curved space that locally resembles Euclidean space but has its own unique geometric properties, such as distances and angles defined by a Riemannian metric. Unlike flat spaces, a Riemannian manifold can have curvature, allowing the study of diverse geometric shapes, from spheres to more abstract surfaces. This structure enables the computation of shortest paths (geodesics), volumes, and curvature, making it a fundamental tool in areas like differential geometry, physics, and data science.

A manifold is a type of topological space that resembles Euclidean space in small, local regions around each point. In other words, for every point on a manifold, there is a neighborhood that is similar to a flat multidimensional space. A Riemannian manifold is a type of manifold equipped with a smoothly varying inner product on the tangent spaces at each point. This means that for every point on the manifold, the tangent space has a way of measuring distances and angles, and these measurements change smoothly from point to point \citep{lee2018introduction}. In simpler terms, a Riemannian manifold is a smooth, curved space that locally behaves like Euclidean space but has its own geometric properties, such as how distances, angles, and volumes are defined. These properties are determined by a Riemannian metric, which generalizes the concept of measuring lengths and angles in flat space to curved spaces \citep{lee2018introduction}.

A Riemannian manifold can have curvature, unlike a flat space. This curvature allows the study of geometric shapes ranging from spheres and cylinders to more abstract surfaces. The Riemannian structure enables us to compute geodesics, volumes, and various types of curvature. This makes Riemannian manifolds fundamental in fields like differential geometry and physics, and increasingly important in data science, where curved spaces are used to model complex datasets \citep{you2021re}.

\subsection{Effective Resistance in Graph Theory} \label{app:effective_resistance}
Effective resistance, also known as resistance distance, is a concept in spectral graph theory that draws an analogy between electrical networks and graphs, helping to quantify how easily current can flow between two nodes, where the edges are treated as resistors. The effective resistance between nodes provides insight into the connectivity between the network. This means two nodes with lower effective resistance values have higher connectivity \citep{ellens2011effective}.

\subsection{Scalable estimation of effective resistances} \label{app:estimatation_eff_res}To address the computational complexity associated with directly computing eigenvalues and eigenvectors required for estimating edge effective resistances, we leverage a scalable framework for approximating the eigenvectors of the graph Laplacian matrix using the Krylov subspace \citep{saad2011numerical}.
Let $A$ denote the adjacency matrix of a graph \(G\), consider its order-$m$ Krylov subspace $\mathbf{K}_m(A, x)$ that is a vector space spanned by the vectors computed through power iterations $x, A x, A^2 x, \ldots, A^{m-1} x$ \citep{article}. By enforcing orthogonality among the above vectors in the Krylov subspace, a new set of mutually orthogonal vectors of unit lengths can be constructed for approximating the original Laplacian eigenvectors in \ref{eq:eff_resist0}, which are denoted as $\Tilde{u}_{1}$, $\Tilde{u}_{2}$, $\ldots$, $\Tilde{u}_{m}$. To estimate the effective resistance between two nodes $p$ and $q$, we can exploit the approximated eigenvectors:
\begin{equation}\label{eq:ER_estimation}
R_{eff}(p,q) \approx \sum\limits_{i= 1}^{m} \frac{(\Tilde{u}^\top_{i}  b_{pq})^2}{\Tilde{u}^\top_{i} L \Tilde{u}_{i}},
\end{equation}
 % + \alpha \|f_p - f_q\|
where  $\Tilde{u}_{i}$ represents the approximated eigenvector corresponding to the $i$-th eigenvalue of $L$.
% \end{lemma}
\vspace{-5pt}
\paragraph{Graph coarsening with node features} In order to account for the variation in node features along with edge resistive distance, we can use the following modified effective resistance formulation:
% We expand this effective resistance with the difference between the Node features of the nodes, the modified effective resistance \(R^*_{e f f}(p, q)\) stands:
\begin{equation} \label{mod_effective_res}
    R^*_{e f f}(p, q) = R_{e f f}(p, q) + \alpha \|f_p - f_q\|,
\end{equation}
% \[\]
where \(f_p\) and \(f_q\) are node feature vectors of nodes \(p\) and \(q\), respectively, while \(\alpha\) is a weighting factor that determines the effect of node feature information in the graph coarsening process. For instance, if the weight is sufficiently large, the modified effective resistance between nodes with different features will always exceed that of nodes with similar features, effectively preventing their coarsening.

\subsection{Detailed Proofs Showing GGD is a Metric} \label{ggdproofs}

\subsubsection{Identity Property} 
\begin{proof}

Let the corresponding SPD matrix of the graph \(G\) be \(\mathcal{L} \in \mathbb{S}_{++}^n\). From Equation \ref{ggd}, we have:
\[
GGD(G, G) = \left[\sum_{i=1}^{n} \log^2(\lambda_i(\mathcal{L}^{-1}\mathcal{L}))\right]^{1/2} = \left[\sum_{i=1}^{n} \log^2(\lambda_i(I))\right]^{1/2}.\]
The identity matrix has only one eigenvalue, which is 1. So, \(GGD(G, G) = \left[\log^2(1)\right]^{1/2} = 0\).
\end{proof}

\subsubsection{Positivity Property}
\begin{proof}

Let the corresponding SPD matrices of the graphs \(G_1\) and \(G_2\) be \(\mathcal{L}_1\), \(\mathcal{L}_2 \in \mathbb{S}_{++}^n\). Let the generalized eigenvalues of (\(\mathcal{L}_1^{-1}\mathcal{L}_2\)) be \(\lambda_1, \lambda_2, \lambda_3, \ldots, \lambda_n\). From Equation \ref{ggd}, we get:
\[GGD(G_1, G_2) = \left[\log^2(\lambda_1) + \log^2(\lambda_2) + \log^2(\lambda_3) + \ldots + \log^2(\lambda_n) \right]^{1/2}.\]
Now, \(\log^2(\lambda_1) + \log^2(\lambda_2) + \log^2(\lambda_3) + \ldots + \log^2(\lambda_n) \geq 0\), for any values of \(\lambda_i\).

We can conclude, \(GGD(G_1, G_2) \geq 0\).
\end{proof}
% Let's say 

\subsubsection{Symmetry Property}
\begin{proof}
Let the corresponding SPD matrices of the graphs \(G_1\) and \(G_2\) be \(\mathcal{L}_1\), \(\mathcal{L}_2 \in \mathbb{S}_{++}^n\). Let the generalized eigenvalues of (\(\mathcal{L}_1^{-1}\mathcal{L}_2\)) be \(\lambda_1, \lambda_2, \lambda_3, \ldots, \lambda_n\). From Equation \ref{ggd}, we get:
    \[GGD(G_1, G_2) = \left[\sum_{i=1}^{n} \log^2(\lambda_i)\right]^{1/2}.\]

Suppose \( \lambda_i \) is an eigenvalue of \( \mathcal{L}_1^{-1}\mathcal{L}_2 \) with corresponding eigenvector \( v_i \), i.e.,
\[
\mathcal{L}_1^{-1} \mathcal{L}_2 v_i = \lambda_i v_i.
\]
Multiplying both sides by \( \mathcal{L}_1 \), we get:
\[
\mathcal{L}_2 v_i = \lambda_i \mathcal{L}_1 v_i.
\]
Now multiply both sides by \( \mathcal{L}_2^{-1} \):
\[
v_i = \lambda_i \mathcal{L}_2^{-1} \mathcal{L}_1 v_i.
\]
Rearranging, we obtain:
\[
\mathcal{L}_2^{-1} \mathcal{L}_1 v_i = \frac{1}{\lambda_i} v_i.
\]
So, the eigenvalues of \( \mathcal{L}_2^{-1}\mathcal{L}_1 \) are \( \frac{1}{\lambda_1}, \frac{1}{\lambda_2}, \ldots, \frac{1}{\lambda_n} \), with the same eigenvectors.

% So, the eigenvalues of (\(\mathcal{L}_2^{-1}\mathcal{L}_1\)) will be \(\frac{1}{\lambda_1}, \frac{1}{\lambda_2}, \frac{1}{\lambda_3}, \ldots, \frac{1}{\lambda_n}\).

\[GGD(G_2, G_1) = \left[\sum_{i=1}^{n} \log^2(\frac{1}{\lambda_i})\right]^{1/2}.\]

Now, \(\log\left(\frac{1}{\lambda_i}\right) = -\log(\lambda_i)\); so, \(\log^2\left(\frac{1}{\lambda_i}\right) = \log^2(\lambda_i)\).

So, we can conclude \(GGD(G_1, G_2) = GGD(G_2, G_1)\).
\end{proof}

\subsubsection{Triangle Inequality}
\begin{proof}
    Let, \(\mathcal{L}_1, \mathcal{L}_2, \mathcal{L}_3 \in \mathbb{S}_{++}^n\) are three SPD matrices corresponding to graphs \(G_1, G_2, G_3\).
    
    Now, The Frobenius norm $\|X\|_F$ is the geodesic length at $d(\exp X, I)=\|X\|_F$ \citep{bonnabel2010riemannian}.
    Hence at identity, $d(\mathcal{L}, I)=\|\log \mathcal{L}\|_F$. 
    
    From \citep{bonnabel2010riemannian, you2021re} we get,
    \begin{equation}
    GGD(G_1, G_2)=GGD\left(G_1^{-1 / 2} G_2 G_1^{-1 / 2}, I\right)=\left\|\log \left(\mathcal{L}_1^{-1 / 2} \mathcal{L}_2 \mathcal{L}_1^{-1 / 2}\right)\right\|_F=\left\|\log \left(\mathcal{L}_1^{-1} \mathcal{L}_2\right)\right\|_F.
% =\left( \sum_i \log^2(\lambda_i(A^{-1}B)) \right)^{1/2}
\end{equation}
    We know,
\[
\mathcal{L}_1^{-1} \mathcal{L}_3 = \mathcal{L}_1^{-1} (\mathcal{L}_2 \mathcal{L}_2^{-1}) \mathcal{L}_3 = (\mathcal{L}_1^{-1} \mathcal{L}_2)(\mathcal{L}_2^{-1} \mathcal{L}_3).
\]
Now using the Frobenius norm inequality, we get:
\[\| \mathcal{L}_1^{-1} \mathcal{L}_3 \| = \| (\mathcal{L}_1^{-1} \mathcal{L}_2)(\mathcal{L}_2^{-1} \mathcal{L}_3) \| \leq \| \mathcal{L}_1^{-1} \mathcal{L}_2 \| \| \mathcal{L}_2^{-1} \mathcal{L}_3 \|.\]
Now taking logarithms on both sides:
\[\| \log(\mathcal{L}_1^{-1} \mathcal{L}_3) \|  \leq \| \log(\mathcal{L}_1^{-1} \mathcal{L}_2) \|+ \| \log(\mathcal{L}_2^{-1} \mathcal{L}_3) \|.\]
Using Equation \ref{ggd}, we conclude:
\[GGD(G_1, G_3) \leq GGD(G_1, G_2) + GGD(G_2, G_3).\]
\end{proof}

\subsection{Graph Matching Recovery} \label{grampa_proof}

Given symmetric matrices \(A_1\), \(A_2\) and \(Z\) from the Gaussian Wigner model, where \( A_{2\pi^*} = A_1 + \sigma Z \), there exist constants \(c, c' > 0\) such that if \(1/n^{0.1} \leq \eta \leq c / \log n\) and \(\sigma \leq c' \eta\), then with probability at least $1 - n^{-4}$ for all large n, the matrix $\widehat{X}$ in equation \ref{similarity_matrix} satisfies,
\begin{equation}
\label{equ:theorem_matching}
    \min _{i \in[n]} \widehat{X}_{i, \pi^*(i)}>\max _{i, j \in[n]: j \neq \pi^*(i)} \widehat{X}_{i j}
\end{equation}

and hence, the GRAMPA algorithm correctly recovers the permutation estimation matrix \(\pi^*\).

% From \ref{similarity_matrix}, the similarity matrix \(\widehat{X}\) is defined as:
% \[
%     \widehat{X} = \sum_{i,j=1}^n \frac{u_i u_i^T \mathbf{J} v_j v_j^T}{(\zeta_i - \mu_j)^2 + \eta^2},
%     \]
% where \(u_i\) and \(v_j\) are eigenvectors of \(A_1\) and \(A_2\) respectively, and \(\zeta_i\) and \(\mu_j\) are their corresponding eigenvalues. \(\mathbf{J}\) is a all-one matrix, and \(\eta\) is the tuning parameter.

Now, this proof is divided into two parts:

\begin{lemma}[Noiseless Setting Diagonal Dominance] \label{grampa_noiseless_lemma} In a noiseless situation, means replacing $A_2$ with $A_1$, similarity matrix \(\widehat{X}^*\) is defined as:
\begin{equation} \label{noiseless_diag_dom}
    \widehat{X}^* = \widehat{X} (A_1, A_1) = \sum_{i,j=1}^n \frac{u_i u_i^T \mathbf{J} u_j u_j^T}{(\zeta_i - \zeta_j)^2 + \eta^2}.
\end{equation}

For some constants $C, c>0$, if $1 / n^{0.1}<\eta<c / \log n$, then with probability at least $1-5 n^{-5}$ for large $n$, it can be proved that the diagonal components of \(\widehat{X}^*\) are dominant by showing \citep{fan2019spectral}:
\begin{equation} 
    \min_{i \in [n]} (\widehat{X}^*)_{ii} > \frac{1}{3 \eta^2}
    \end{equation}
    and
    \begin{equation}
    \label{bound_noise}
    \max_{i, j \in[n]: i \neq j} (\widehat{X}^*)_{ij} < C \left(\frac{\sqrt{\log n}}{\eta^{3/2}} + \frac{\log n}{\eta}\right).
\end{equation}

\end{lemma}

\begin{lemma}[Bounding the Noise Impact] \label{grampa_noisebound_lemma}
The difference between the similarity matrix \(X\) in the presence of noise and the noiseless situation is bounded. If $\eta>1 / n^{0.1}$, then for a constant $C>0$, with probability at least $1-2 n^{-5}$ for large $n$, it can be shown \citep{fan2019spectral}:

\begin{equation} \label{equ:noise_impact}
    \max_{i,j \in [n]} |\widehat{X}_{ij} - (\widehat{X}^*)_{ij}| < C \sigma \left(\frac{1}{\eta^3} + \frac{\log n}{\eta^2} \left(1 + \frac{\sigma}{\eta}\right)\right).
\end{equation}

\end{lemma}

Assuming lemma \ref{grampa_noiseless_lemma} and \ref{grampa_noisebound_lemma}, for some $c, c^{\prime}>0$ sufficiently small, and by setting \(\eta < c / \log n\) and \(\sigma < c' \eta\), the algorithm ensures that the right sides of both equations \ref{bound_noise} and \ref{equ:noise_impact} are at most $1/(12\eta^2)$. Then when $\pi^* = id$ (the identity permutation), these lemmas combine to imply: 

\begin{equation}
    \min _{i \in[n]} \widehat{X}_{i i}>\frac{1}{4 \eta^2}>\frac{1}{6 \eta^2}>\max _{i, j \in[n]: i \neq j} \widehat{X}_{i j}
\end{equation}

with probability at least $1 - n^{-4}$. So, all diagonal entries of \(\widehat{X}\) are larger than all off-diagonal entries, thereby achieving exact recovery \citep{fan2019spectral}.

\subsection{Relation between Generalized Eigenvalues with Cut Mismatch} \label{app:eig_vs_cut}

\begin{figure}
  \centering
  \includegraphics[width=0.6\linewidth]{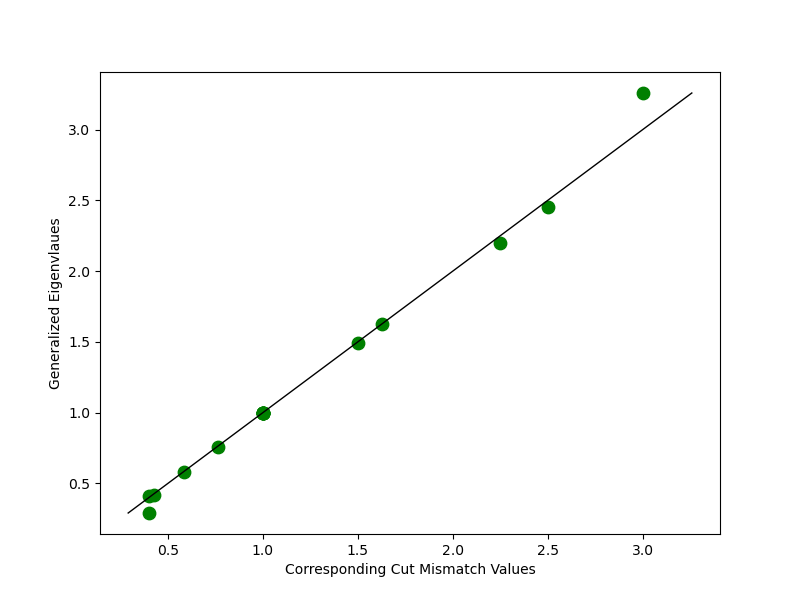}
  \caption{Generalized eigenvalues and their corresponding cut mismatches}
  \label{img_cut}
\end{figure}

We selected two graphs from the MUTAG dataset and computed their generalized eigenvalues following the procedure for calculating the Generalized Graph Distance (GGD), which involves determining the node-to-node correspondence. Subsequently, we considerd all possible subsets of nodes and evaluate their corresponding cut mismatches. As shown in Figure \ref{img_cut}, each generalized eigenvalue is closely associated with a cut mismatch. This empirical observation supports our hypothesis that the GGD between two input graphs is strongly correlated with structural mismatches in graphs.

% Still editing the image Figure \ref{img_cut}.

\subsection{Choosing \texorpdfstring{\(\epsilon\)}{epsilon} for converting Laplacians to SPD matrices} \label{app:epsilon_ap}
Laplacian matrices are symmetric positive semi-definite (SPSD) matrices. To convert these to symmetric positive definite (SPD) matrices, we added a diagonal matrix with very small values ($\epsilon$). We used $0.0001$ as the small value ($\epsilon$) in our experiments. When working with Laplacian matrices of a weighted or unweighted graph, values significantly smaller than the edge weights of that graph have a minimal effect on the transformation. We conducted additional experiments with different small values and included the results in Tables \ref{epsilon1} and \ref{epsilon2}. In Table \ref{epsilon1}, we observed that in our specific case with the MUTAG graph dataset \citep{morris2020tudataset}, where all graphs are unweighted, any value less than 0.001 has an almost negligible influence on the performance of the graph classification task. Additionally, when using values equal to or less than 0.001, the GGD value remains almost the same, as shown in Table \ref{epsilon2}.

\begin{table}
    \centering
    \caption{Classification accuracy using MUTAG dataset with different values of $\epsilon$.}
    \label{epsilon1}
    \begin{tabular}{lccc}
        \hline
        Value of $\epsilon$ & Classification accuracy & Value of $\epsilon$ & Classification accuracy \\
        \hline
        0.1 & 76.38 $\pm$ 7.89 & 1e-4 & 85.96 $\pm$ 5.26\\ 
        5e-2 &  79.02 $\pm$ 6.58     &  1e-5 & 84.21 $\pm$ 5.26\\
        1e-2 & 79.02 $\pm$ 5.26 & 1e-6 & 85.96 $\pm$ 5.26\\
        5e-3 &  81.57 $\pm$ 7.89 & 1e-7 & 85.96 $\pm$ 7.89\\ 
        1e-3 & 81.57 $\pm$ 7.89 & &  \\
        \hline
    \end{tabular}
\end{table}

\begin{table}
    \centering
    \caption{GGD values using MUTAG dataset for different values of $\epsilon$.}
    \label{epsilon2}
    \begin{tabular}{ccc}
        \hline
        Value of & Normalized GGD of a random  & Average normalized \\
         $\epsilon$& graph pair (MUTAG[85], MUTAG[103]) & GGD of 1000 pairs\\
        \hline
        0.1 & 0.712 & 0.727 \\
        5e-2 & 0.827 & 0.834 \\
        1e-2 & 0.952 & 0.959 \\
        5e-3 & 0.978 & 0.979 \\
        1e-3 & 0.996 & 0.995 \\
        1e-4 & 0.9996 & 0.9995 \\
        1e-5 & 0.99995 & 0.99996 \\
        1e-6 & 0.999996 & 0.999996 \\
        1e-7 & 1 & 1 \\
        \hline
    \end{tabular}
\end{table}

% This is equivalent to adding a self-loop in each of the nodes of the input graphs. We experimented with this value to check the effect on the value of GGD calculated. 

\subsection{Partial Node Features} \label{app:partial_nf}

Cutting-edge graph distance metrics like TMD rely on node attributes to compute the dissimilarity between graphs, resulting in more accurate outcomes when all attributes are available. However, acquiring datasets with complete node attributes is often unattainable in real-world scenarios, leading to partially missing features In such scenarios when only partial node features are available, we compare TMD with GGD to better understand their differences. Table \ref{partial_table} shows that the TMD metric outperforms GGD at various levels when node features are fully accessible. However, when node features are randomly removed from the MUTAG dataset, the accuracy of TMD degrades substantially. 

\begin{table}
    \centering
    \caption{Comparison of correlation with GNN outputs and distance metrics with partial node features.}
            \label{partial_table}
    \begin{small}
            \begin{sc}
            \begin{tabular}{lccccc}
                \toprule
                Dist Metric & 0\% & 20\% & 50\% & 80\% & 100\% \\
                \midrule
                GGD        & 0.78 & \textbf{0.78} & \textbf{0.77} & \textbf{0.77} & \textbf{0.77} \\
                TMD, L = 3 & \textbf{0.84} & \textbf{0.78} & 0.72 & 0.63 & 0.61 \\
                TMD, L = 4 & 0.81 & 0.77 & 0.62 & 0.58 & 0.57 \\
                TMD, L = 5 & 0.80 & 0.75 & 0.65 & 0.58 & 0.53 \\
                \bottomrule
            \end{tabular}
            \end{sc}
            \end{small}
            \end{table}

%     \begin{minipage}{\linewidth}
%         \begin{minipage}{0.7\linewidth}
%             \centering
%             \begin{small}
%             \begin{sc}
%             \begin{tabular}{lccccc}
%                 \toprule
%                 Dist Metric & 0\% & 20\% & 50\% & 80\% & 100\% \\
%                 \midrule
%                 GGD        & 0.78 & \textbf{0.78} & \textbf{0.77} & \textbf{0.77} & \textbf{0.77} \\
%                 TMD, L = 3 & \textbf{0.84} & \textbf{0.78} & 0.72 & 0.63 & 0.61 \\
%                 TMD, L = 4 & 0.81 & 0.77 & 0.62 & 0.58 & 0.57 \\
%                 TMD, L = 5 & 0.80 & 0.75 & 0.65 & 0.58 & 0.53 \\
%                 \bottomrule
%             \end{tabular}
%             \end{sc}
%             \end{small}
%         \end{minipage}
%         \hfill
%         \begin{minipage}{0.29\linewidth}
%             \captionof{table}{Comparison of correlation with GNN outputs and distance metrics with partial node features.}
%             \label{partial_table}
%         \end{minipage}
%     \end{minipage}
% \end{table}

\subsection{Approximate GGD on Small Graphs Using Extreme Eigenvalues} \label{ap:extre_eig}

% \begin{table}
% \caption{Performance of GGD using extreme eigenvalues only.}
% \label{riem_only24}
% \centering
% \begin{small}
% \begin{sc}
% \begin{tabular}{lccc}
% \toprule
% Task & \multicolumn{3}{c}{Number of extreme eigenvalues} \\
% \cmidrule(lr){2-4}
%  & 2 & 4 & All \\
% \midrule
% {Correlation\\with GNN} & 0.74 & 0.76 & 0.77 \\
% {Classification\\accuracy} & $81.50\pm6.85$ & $83.87\pm7.56$ & $86.00\pm7.50$ \\
% \bottomrule
% \end{tabular}
% \end{sc}
% \end{small}
% \end{table}

\begin{table}
\centering
\caption{Performance of GGD using extreme eigenvalues only.}
\label{riem_only24}
\begin{small}
\begin{tabular}{lccc} % First column allows wrapping
\toprule
Task & \multicolumn{3}{c}{Number of extreme eigenvalues} \\
\cmidrule(lr){2-4}
& 2 & 4 & All \\
\midrule
Correlation with GNN & 0.74 & 0.76 & \textbf{0.77} \\
Classification accuracy & 81.50$\pm$6.85 & 83.87$\pm$7.56 & \textbf{86.00$\pm$7.50} \\
\bottomrule
\end{tabular}
\end{small}
\end{table}

\begin{figure}
  \centering
  \includegraphics[width=0.6\linewidth]{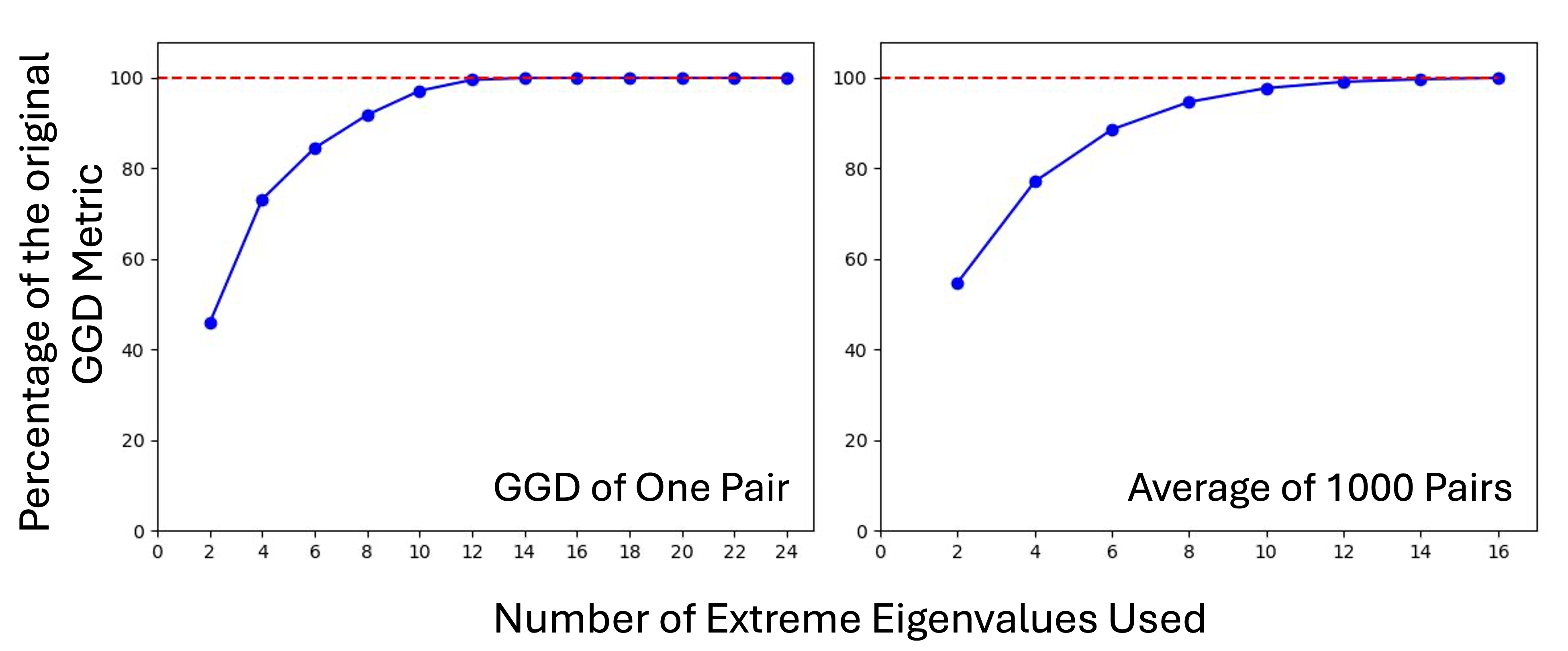}
  \caption{Percentage of the original GGD using numbers of extreme eigenvalues.}
  \label{ext_eig}
\end{figure}

The largest and the smallest eigenvalues correspond to the most dominant mismatches in graph cuts and effective resistance distances, contributing the most to the total GGD value. Similarly, the second largest and smallest eigenvalues correspond to the next significant mismatched cuts. In our experiment, we obtain approximate GGDs using a few extreme eigenvalue pairs and compare them with the ground truth. Figure \ref{ext_eig} illustrates the relative accuracy of the approximate GGDs, in which we observe that the top four pairs of extreme eigenvalues contribute $80\%$ of the total GGD values. In addition, we conduct the SVC classification task and GNN correlation study using GGD with only $2$ and $4$ extreme eigenvalue pairs, respectively, and present the associated findings in Table \ref{riem_only24}.

% \vspace{-5pt}

\subsection{Scalability of GGD Approximation on Very Large Graphs} \label{app:scalability}
For very large graphs, computing the full spectrum of eigenvalues becomes computationally expensive. To address this, we use approximate GGD strategy using only a small fraction of the extreme eigenvalues. Our results in Table \ref{table:ff} show that even with just 2\% of the extreme eigenvalues, the approximation remains highly correlated with the original GGD value. Moreover, the percentage required decreases with increasing graph size, significantly improving scalability. Table \ref{table:gg} demonstrates the resulting runtime improvements.

\begin{table}
\caption{Correlation between GGD values and approximate GGD values using a portion of extreme eigenvalues for very large graphs, where n represents the number of nodes.}
\label{table:ff}
    \centering
    \begin{tabular}{c c c c c c}
        \hline
        Extreme Eigenvalues Used   &   0.5\%  &   1\%   &   2\%   &    4\%   &   All \\
        \hline

Correlation (n $\in$ [5000, 5200]) &  0.8323 & 0.9361 & 0.9772 & 0.9902 & 1                
\\
Correlation (n $\in$ [10000, 10500]) & 0.8666 & 0.9501 & 0.9808 & 0.9928 & 1

\\
Correlation (n $\in$ [15000, 16000]) & 0.8892 & 0.9599 & 0.9879 & 0.9952 & 1

 \\
        \hline
    \end{tabular}
    
\end{table}

\begin{table}
\centering
\caption{Runtime Comparison of Exact vs. Approximate GGD on Very Large Graphs}
\label{table:gg}
\begin{tblr}{
  cells = {c},
  cell{1}{1} = {r=2}{},
  cell{1}{2} = {c=3}{},
  % vlines,
  hline{1,3,6} = {-}{},
  hline{2} = {2-4}{},
}
{Node \\Numbers} & Calculation Time &                           &                           \\
                 & GGD              & Aprx GGD (2\% Eigs) & Aprx GGD (4\% Eigs) \\
n $\in$ [5000, 5200]               & 29.13s           & 7.11s                         & 7.98s                         \\
n $\in$ [10000, 10500]               & 241.33s                &    55.56s                      & 56.29s                        \\
n $\in$ [15000, 16000]               & 920.77s                & 162.30s                         & 163.64s                        
\end{tblr}
\end{table}

\subsection{Comparison of Two Different Riemannian Metrics for SPD Matrices} \label{app:ai_vs_le}
The two most commonly used Riemannian metrics on the SPD manifold are the Affine Invariant Riemannian Metric (AIRM) and the Log-Euclidean Riemannian Metric (LERM) \citep{ilea2018covariance, thanwerdas2023n, chen2024adaptive}. AIRM is a Riemannian metric that remains invariant under affine transformations, meaning the metric is unaffected when matrices are transformed by any invertible operation. The geodesic distance between two SPD matrices, A and B, using AIRM is given by \citep{you2021re, moakher2005differential}:

\begin{equation} \label{equ_ai}
    d_{\text{AIRM}}(A, B) = \| \log(A^{-1} B) \|_F  = \left[\sum_{i=1}^{n} \log^2(\lambda_i(A^{-1}B))\right]^{1/2}.
\end{equation}

On the other hand, LERM addresses some of the computational complexity challenges associated with AIRM by mapping SPD matrices to an Euclidean space through the matrix logarithm operation. In this Euclidean space, computations are simplified. The geodesic distance between two SPD matrices, A and B, using LERM is defined as \citep{huang2014learning}:

\begin{equation} \label{equ_le}
    d_{\text{LERM}}(A, B) = \| \log(A)-\log(B)\|_F.
\end{equation}

\begin{figure}
  \centering
  \includegraphics[width = 0.55\textwidth]{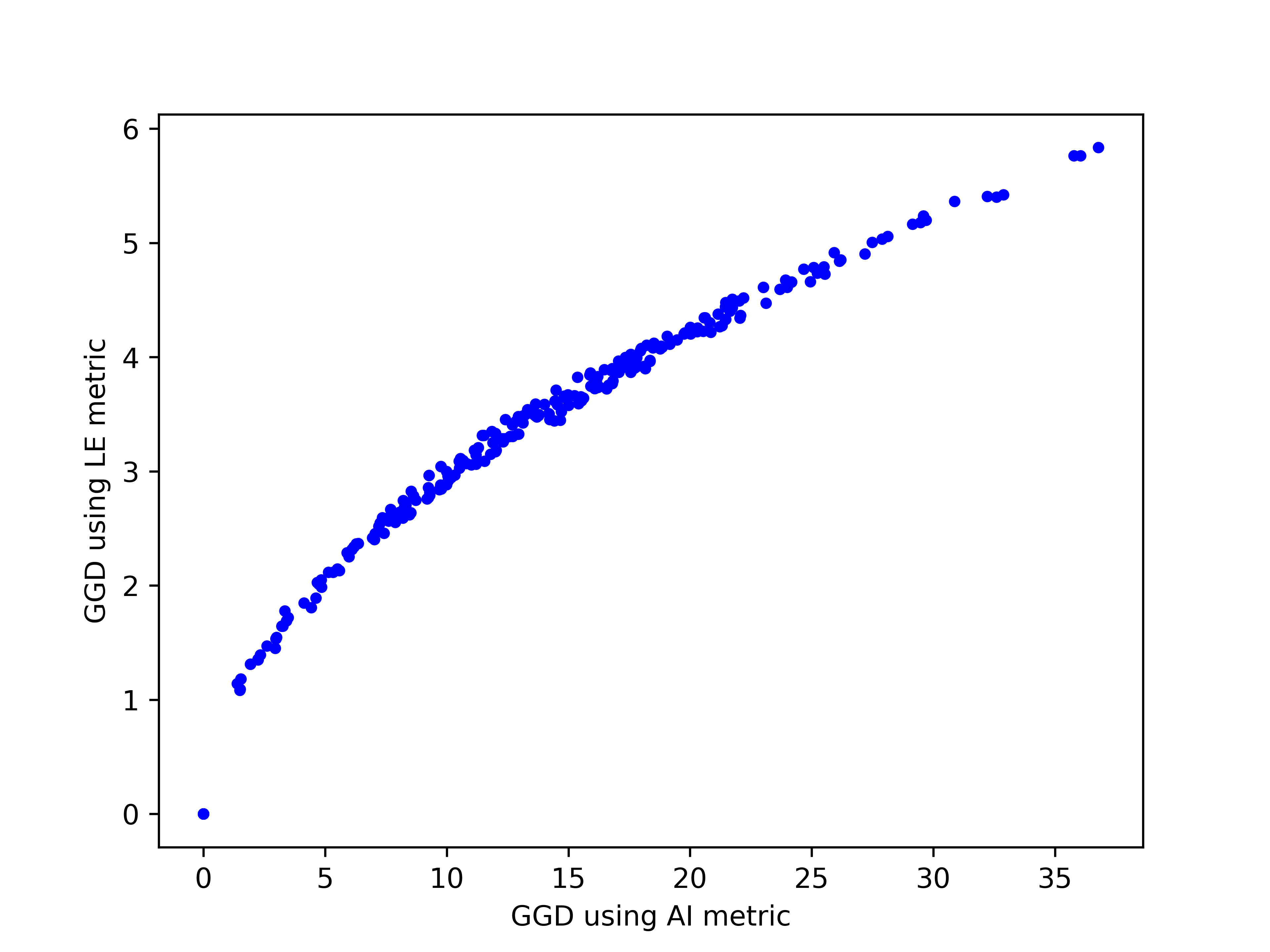}
  \caption{GGD between graph pairs using AI and LE Riemannian metric.}
  \label{ai_vs_le}
\end{figure}

In this paper, we primarily used AIRM to compute geodesics because of its stronger theoretical foundation and its ability to better explain graph cut mismatches. However, for comparison, we also conducted experiments using LERM. Figure \ref{ai_vs_le} shows that the Graph Geodesic Distances computed using the LERM metric are highly correlated with those obtained using AIRM, though the GGD using AIRM demonstrates better performance overall. A detailed performance comparison of these two metrics is provided in Table \ref{ai_le_table}.

\begin{table}
\centering
\caption{Comparison between Riemannian metrics for GGD calculation.}
\label{ai_le_table}
\begin{tabular}{lcc} 
\hline
\multicolumn{1}{c}{Riemannian metric} & Correlation with GNN output & Classification accuracy  \\ 
\hline
Affine-Invariant               & \textbf{0.7786}                      & \textbf{86.00±7.50\%}                 \\
Log-Euclidean                  & 0.7634                      & 84.38\%                  \\
\hline
\end{tabular}
\end{table}

\subsection{Using Normalized Laplacians for GGD Calculation}
\label{app:norm_lap}
In many existing studies, the normalized Laplacian matrix is commonly used to study spectral graph properties \citep{chung1997spectral}. The normalized Laplacian matrix of a graph $G$ is defined as: $L_{norm} = I - A_{norm}$, where $A_{norm}$ is the normalized adjacency matrix. The normalized adjacency matrix is expressed as \citep{chung1997spectral}:

\begin{equation} \label{norm_adj}
    A_{norm} = D^{-1/2}AD^{-1/2},
\end{equation}

where  $D$ represents the diagonal degree matrix, and $A$ denotes the adjacency matrix of the graph.

Form equation \ref{norm_adj}, we can derive:
\begin{equation} \label{norm_laplacian}
    L_{norm} = I - D^{-1/2}AD^{-1/2} = D^{-1/2}(D - A)D^{-1/2} = D^{-1/2}LD^{-1/2}.
\end{equation}

Similar to the Laplacian matrices of graphs, normalized graph Laplacian matrices are also symmetric and positive semi-definite. Therefore, it is necessary to add small values to the diagonal elements of these matrices. However, our experiments reveal that the GGD calculation is highly sensitive to this small value ($\epsilon$), resulting in significant fluctuations across different values, as demonstrated in Table \ref{epsilon_on_norm}. Additionally, the geodesic distances computed with the modified normalized Laplacian matrices exhibit poor accuracy in both classification tasks and their correlation with GNN outputs.

% \begin{table}
% \centering
% \caption{Effect of epsilon in the calculation of GGD using normalized Laplacian matrices}
% \label{epsilon_on_norm}
% \begin{tabular}{lcccc} 
% \hline
% Epsilon                               & 0.01   & 0.001  & 0.0001  & 0.00001  \\ 
% \hline
% GGD using Laplacian matrix            & 16.235 & 16.775 & 16.832  & 16.838   \\
% GGD using normalized Laplacian matrix & 384.1  & 254.44 & 188.345 & 165.332  \\
% \hline
% \end{tabular}
% \end{table}

\begin{table}
\centering
\caption{Effect of epsilon in the calculation of GGD using normalized Laplacian matrices.}
\label{epsilon_on_norm}
\begin{tabular}{lcccc} 
\hline
\multicolumn{1}{c}{Value of $\epsilon$ }                 & 0.01   & 0.001  & 0.0001  & 0.00001  \\ 
\hline
GGD using Laplacian matrices            & 16.235 & 16.775 & 16.832  & 16.838   \\
GGD using normalized Laplacian matrices & 384.097  & 254.440 & 188.345 & 165.332  \\
\hline
\end{tabular}
\end{table}

\subsection{Effect of tuning parameter \texorpdfstring{$\eta$}{eta} on Graph Matching} \label{eta_appendix}
\begin{figure}
  \centering
  \includegraphics[width = 0.45\textwidth]{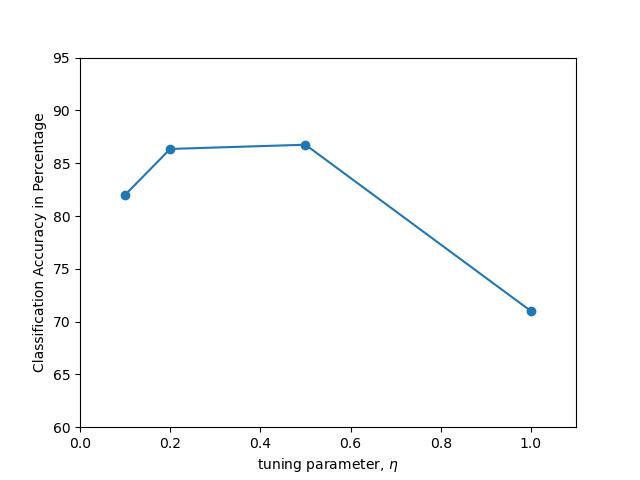}
  \caption{Classification accuracy vs GRAMPA tuning parameter.}
  \label{eta_vs_acc}
\end{figure}

In the original work, it was suggested that the regularization parameter $\eta$ needs to be chosen so that $\sigma \vee n^{-0.1} \lesssim \eta \lesssim 1 / \log n$ \citep{fan2020spectral}. It is also mentioned that for practical cases, computing permutation matrix for different values of $\eta$ in an iterative way can result in better accuracy. The GRAMPA uses $\eta = 0.2$ for all their experiments \citep{fan2020spectral}.\\
We used a few values of $\eta$ in the classification problem using the MUTAG dataset and got that the best accuracy is obtained at $\eta = 0.5$. In Figure \ref{eta_vs_acc}, the performance of the tuning parameter is demonstrated.

\subsection{Sensitivity of GGD to Simple Graph Perturbations} \label{app:perturbation} 

Due to the simplicity and general applicability of the GGD, it can be readily computed between any pair of undirected graphs. To evaluate the robustness of GGD under simple structural changes, we conducted a series of experiments involving basic graph perturbations across both small and large graphs.

\begin{figure}[H] 
    \centering
    
    \begin{minipage}{0.32\textwidth}
        \centering
        \includegraphics[width=\linewidth]{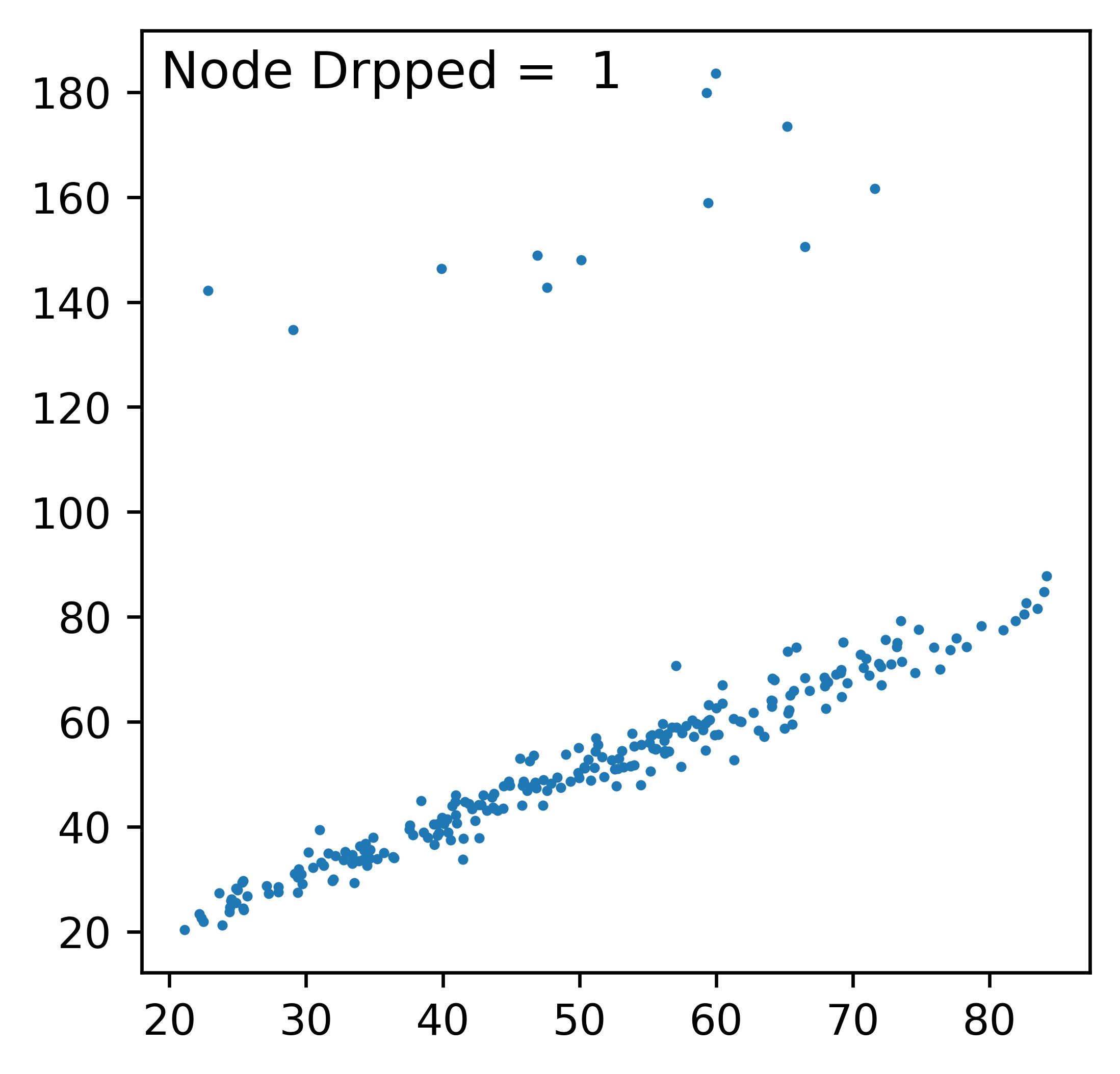}
    \end{minipage}
    \hfill
    \begin{minipage}{0.32\textwidth}
        \centering
        \includegraphics[width=\linewidth]{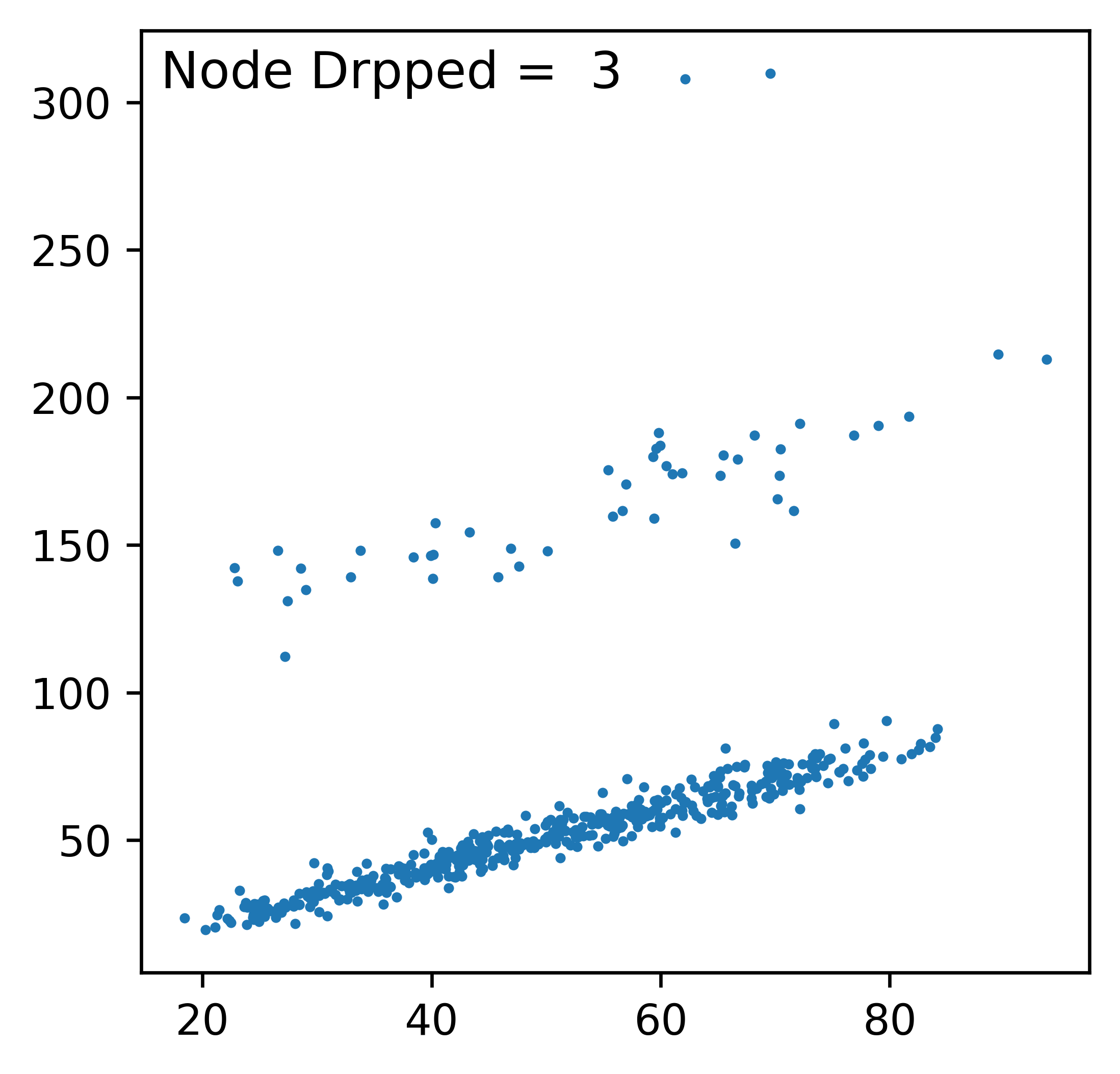}
    \end{minipage}
    \hfill
    \begin{minipage}{0.32\textwidth}
        \centering
        \includegraphics[width=\linewidth]{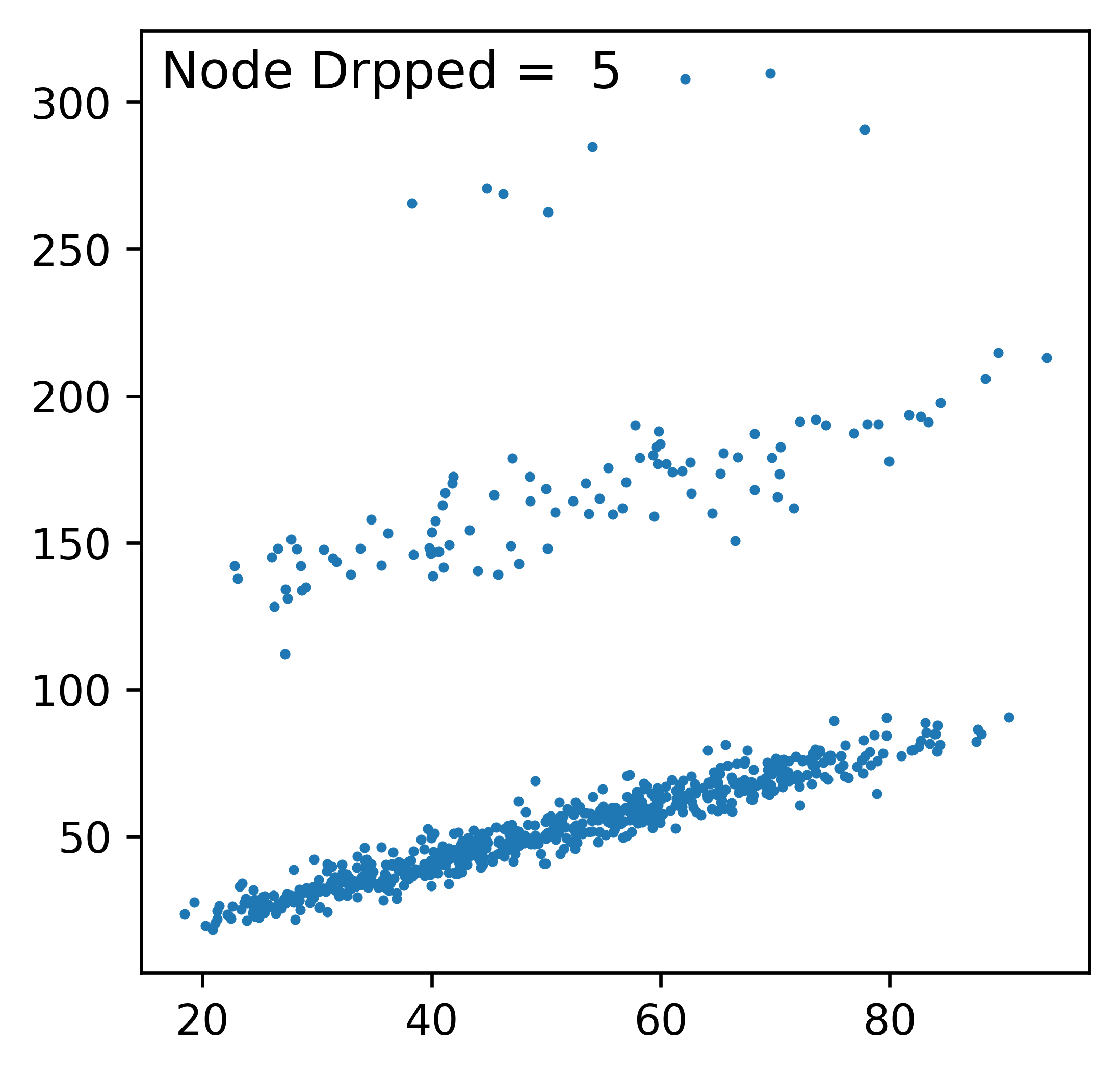}
    \end{minipage}
    \caption{GGD (x-axis) vs GGD after \textbf{Node Drop} perturbation (y-axis) for smaller graphs.}
    \label{fig:per_1}
\end{figure}

\begin{figure}[H]
    \centering
    \begin{minipage}{0.32\textwidth}
        \centering
        \includegraphics[width=\linewidth]{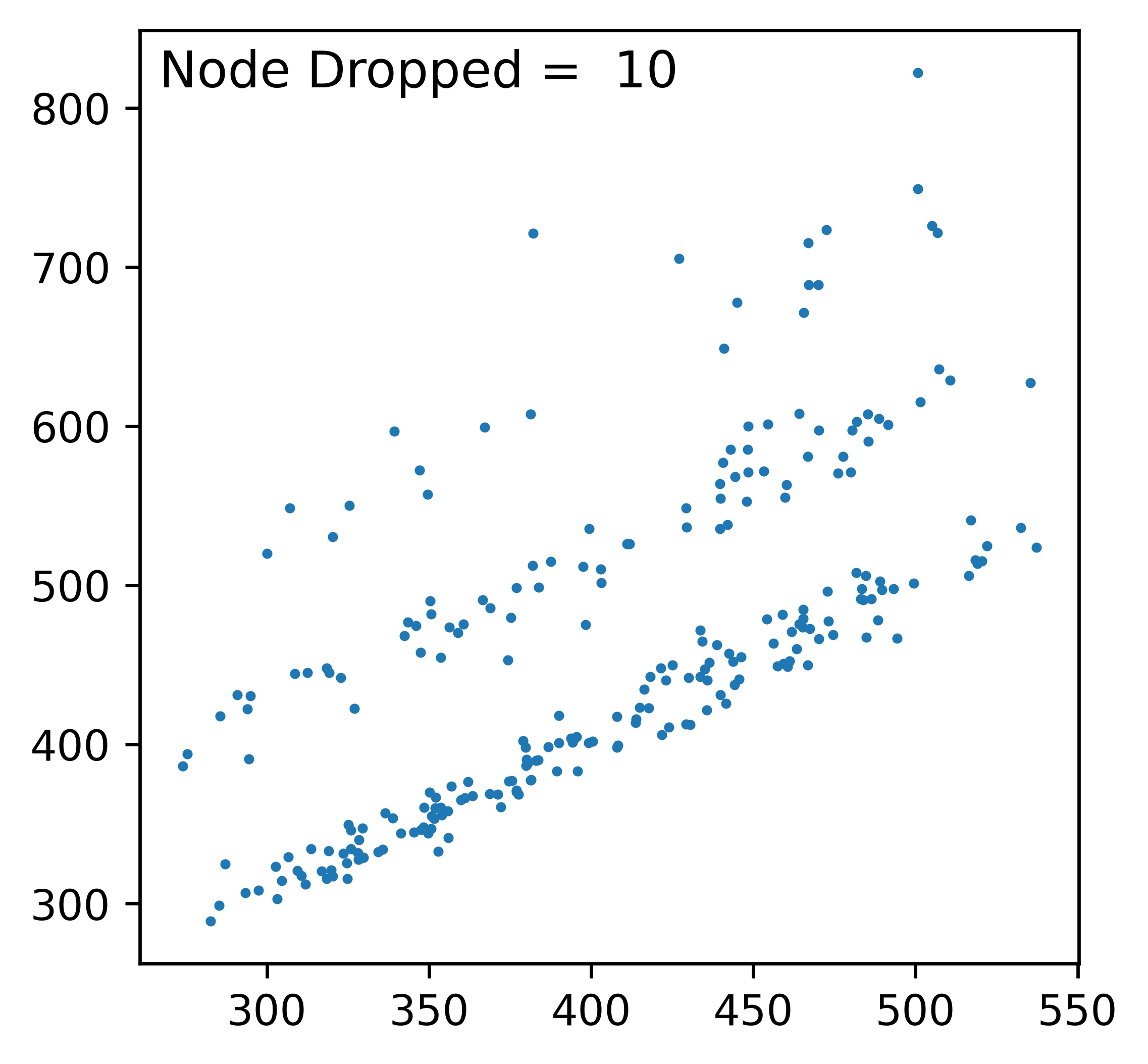}
    \end{minipage}
    \hfill
    \begin{minipage}{0.32\textwidth}
        \centering
        \includegraphics[width=\linewidth]{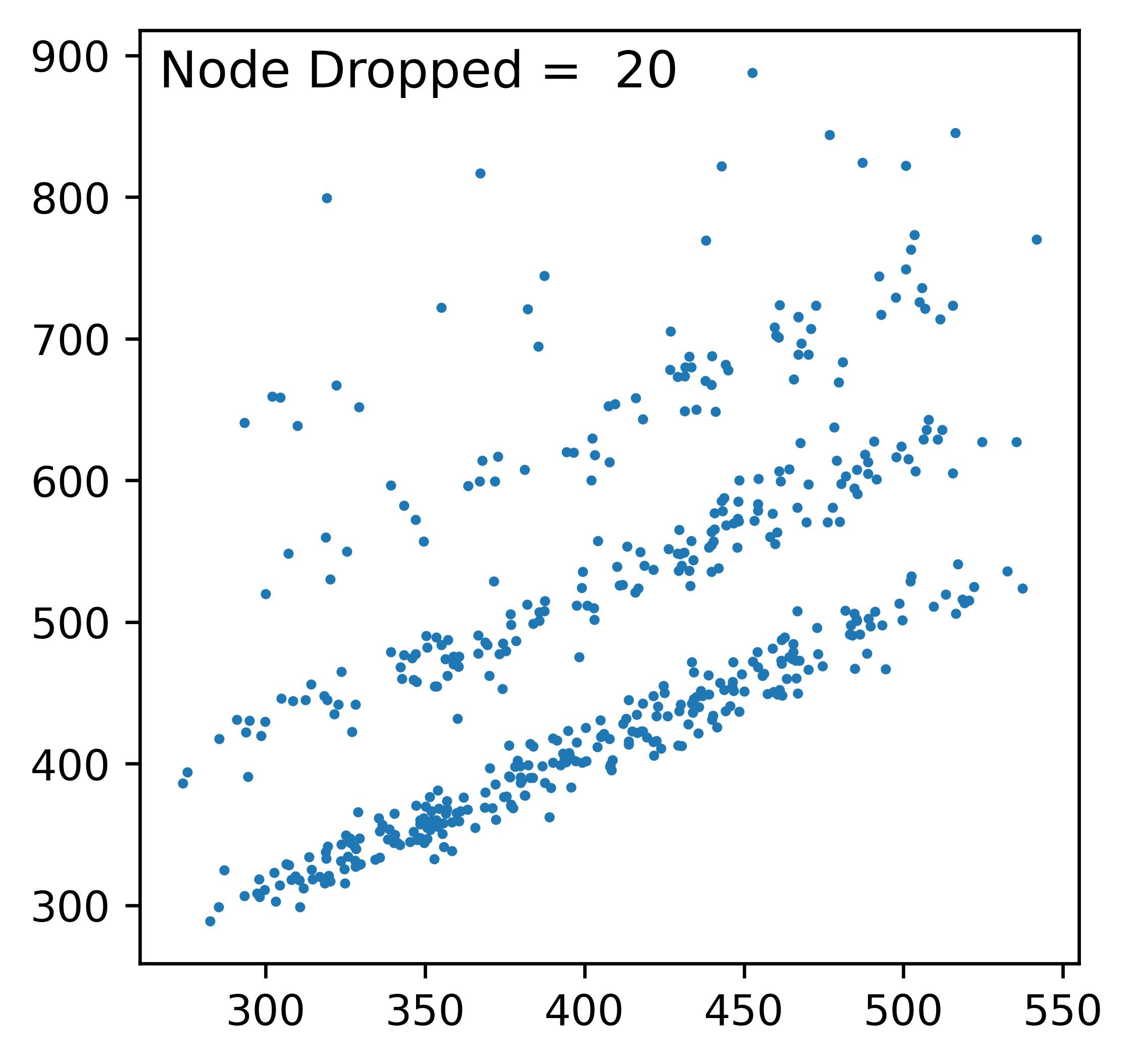}
    \end{minipage}
    \hfill
    \begin{minipage}{0.32\textwidth}
        \centering
        \includegraphics[width=\linewidth]{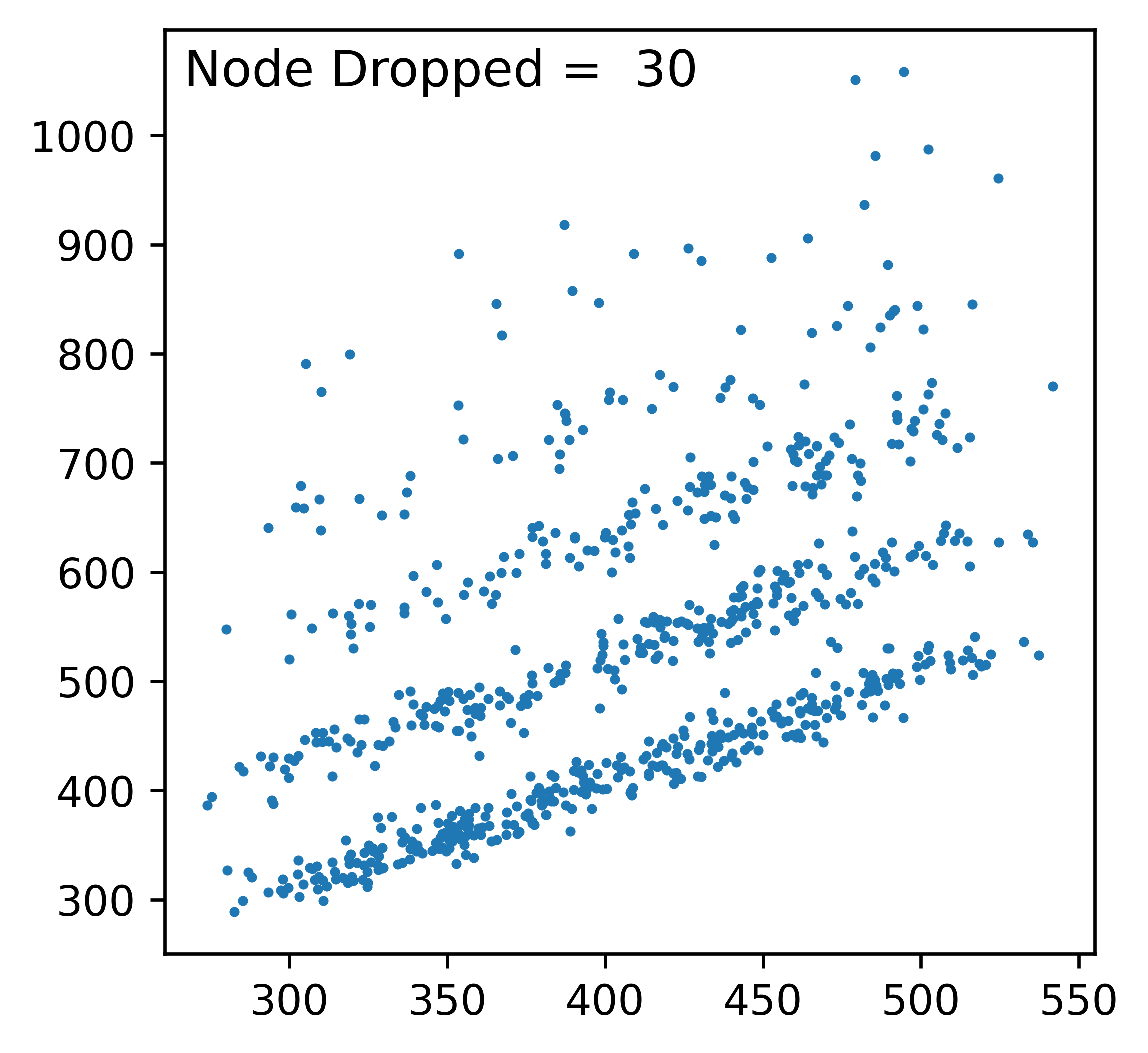}
    \end{minipage}
     
    \caption{GGD (x-axis) vs GGD after \textbf{Node Drop} perturbation (y-axis) for larger graphs.}
    \label{fig:per_2}
\end{figure}

\begin{figure}[H]
    \centering
    \begin{minipage}{0.32\textwidth}
        \centering
        \includegraphics[width=\linewidth]{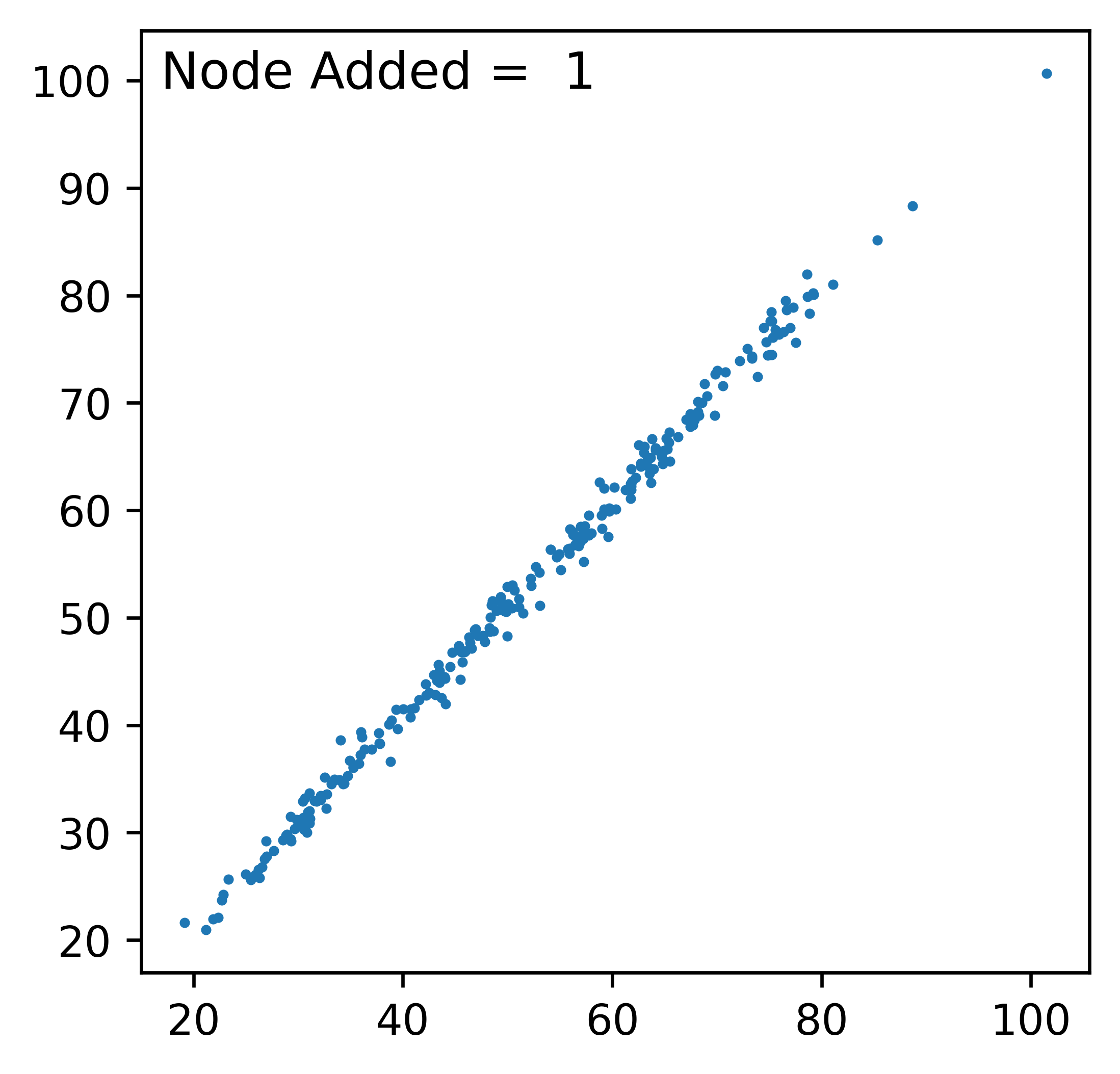}
    \end{minipage}
    \hfill
    \begin{minipage}{0.32\textwidth}
        \centering
        \includegraphics[width=\linewidth]{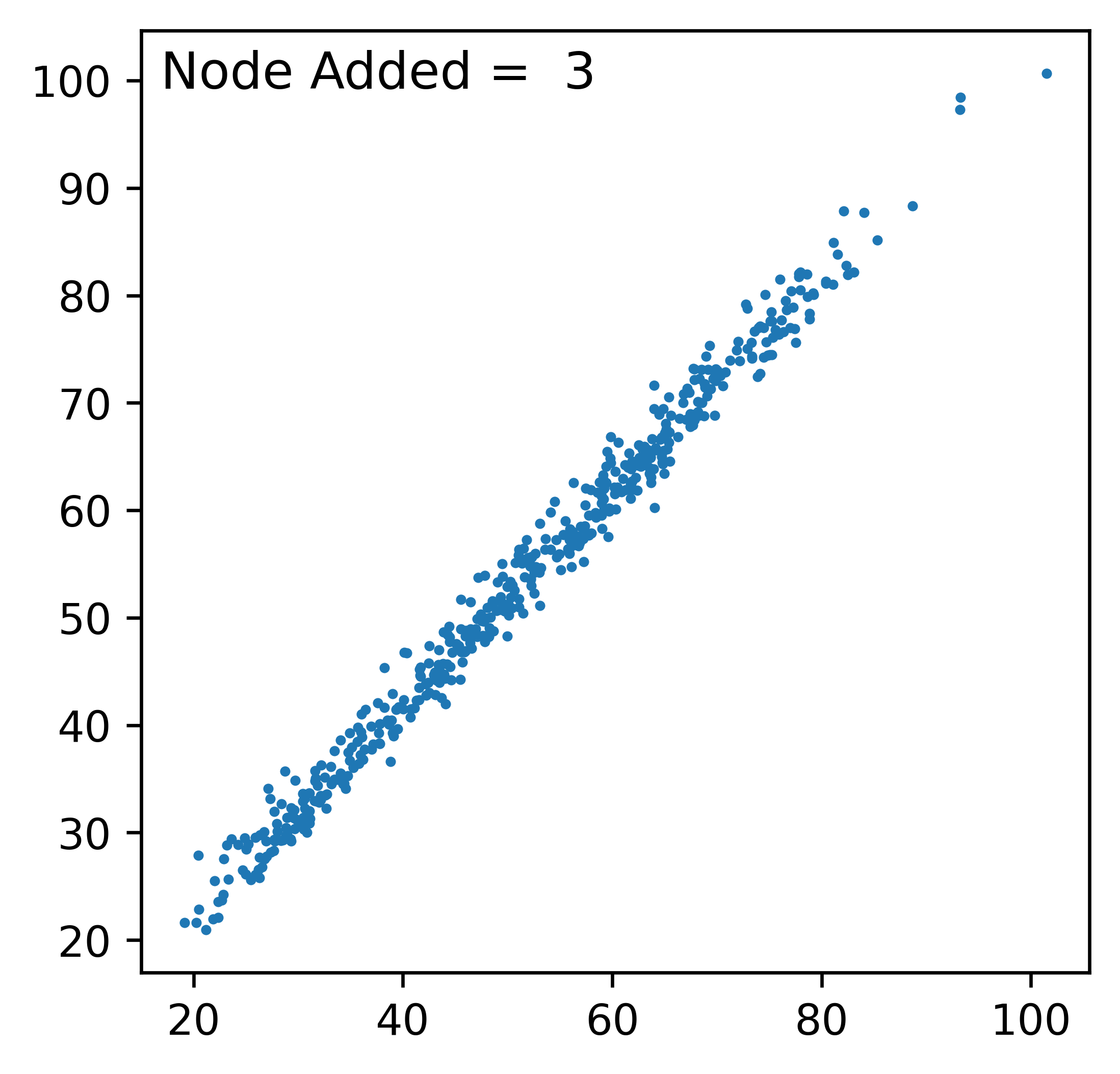}
    \end{minipage}
    \hfill
    \begin{minipage}{0.32\textwidth}
        \centering
        \includegraphics[width=\linewidth]{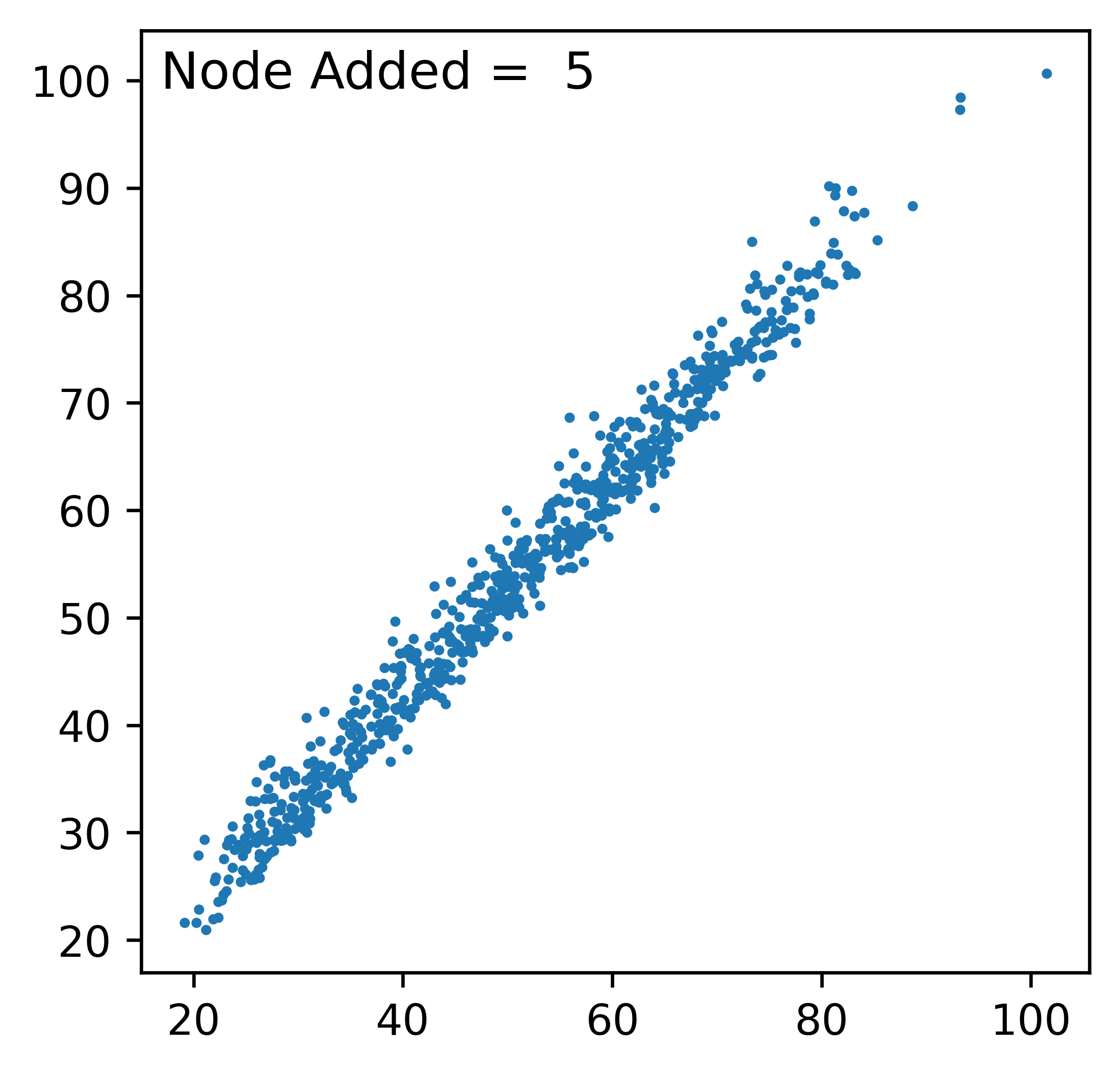}
    \end{minipage}
    \caption{GGD (x-axis) vs GGD after \textbf{Node Addition} perturbation (y-axis) for smaller graphs.} \label{fig:per_3}
\end{figure}

\begin{figure}[H]
    \centering
    \begin{minipage}{0.32\textwidth}
        \centering
        \includegraphics[width=\linewidth]{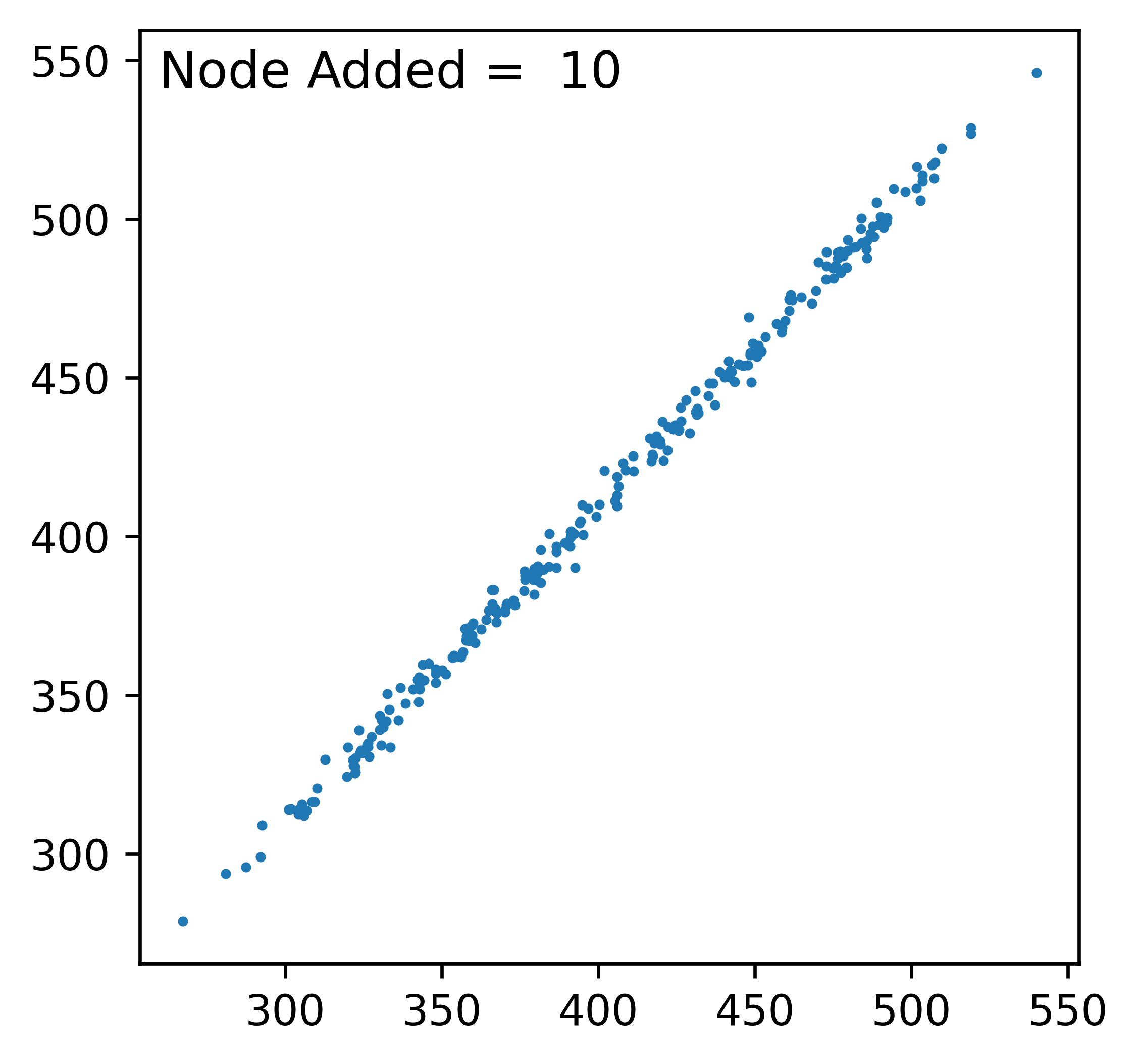}
    \end{minipage}
    \hfill
    \begin{minipage}{0.32\textwidth}
        \centering
        \includegraphics[width=\linewidth]{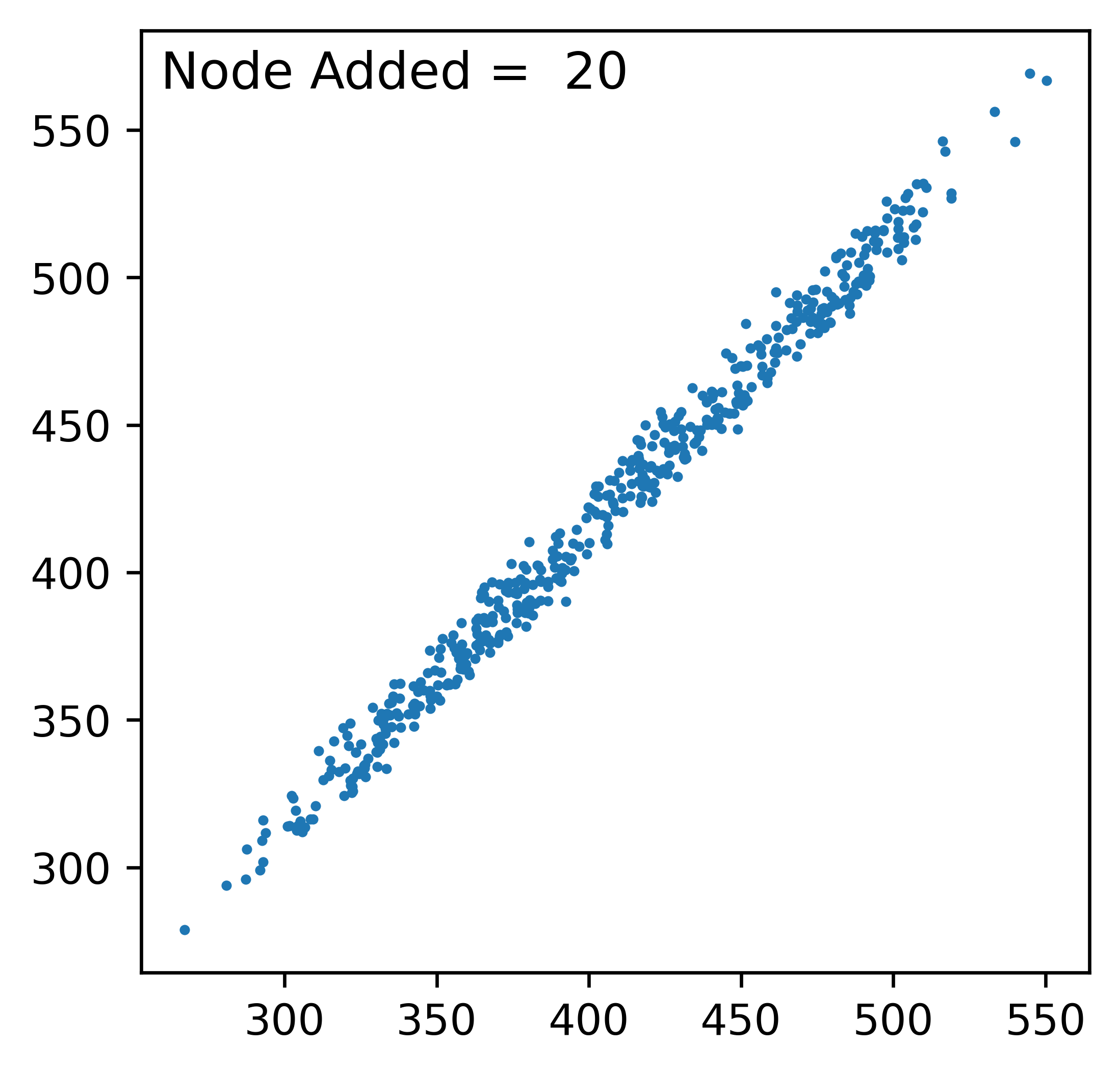}
    \end{minipage}
    \hfill
    \begin{minipage}{0.32\textwidth}
        \centering
        \includegraphics[width=\linewidth]{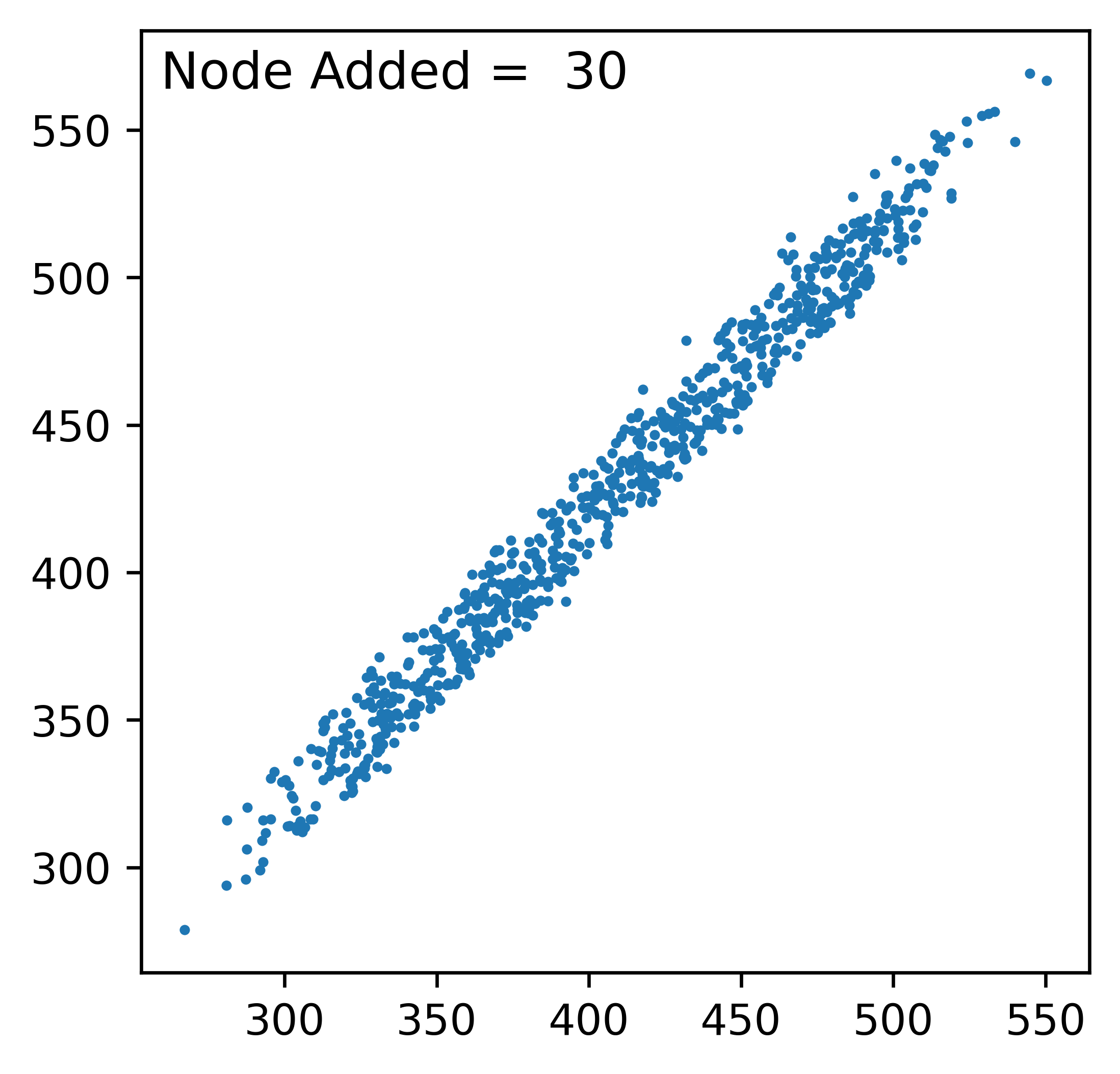}
    \end{minipage}
    \caption{GGD (x-axis) vs GGD after \textbf{Node Addition} perturbation (y-axis) for larger graphs.} \label{fig:per_4}
\end{figure}

\begin{figure}[H]
    \centering
    \begin{minipage}{0.32\textwidth}
        \centering
        \includegraphics[width=\linewidth]{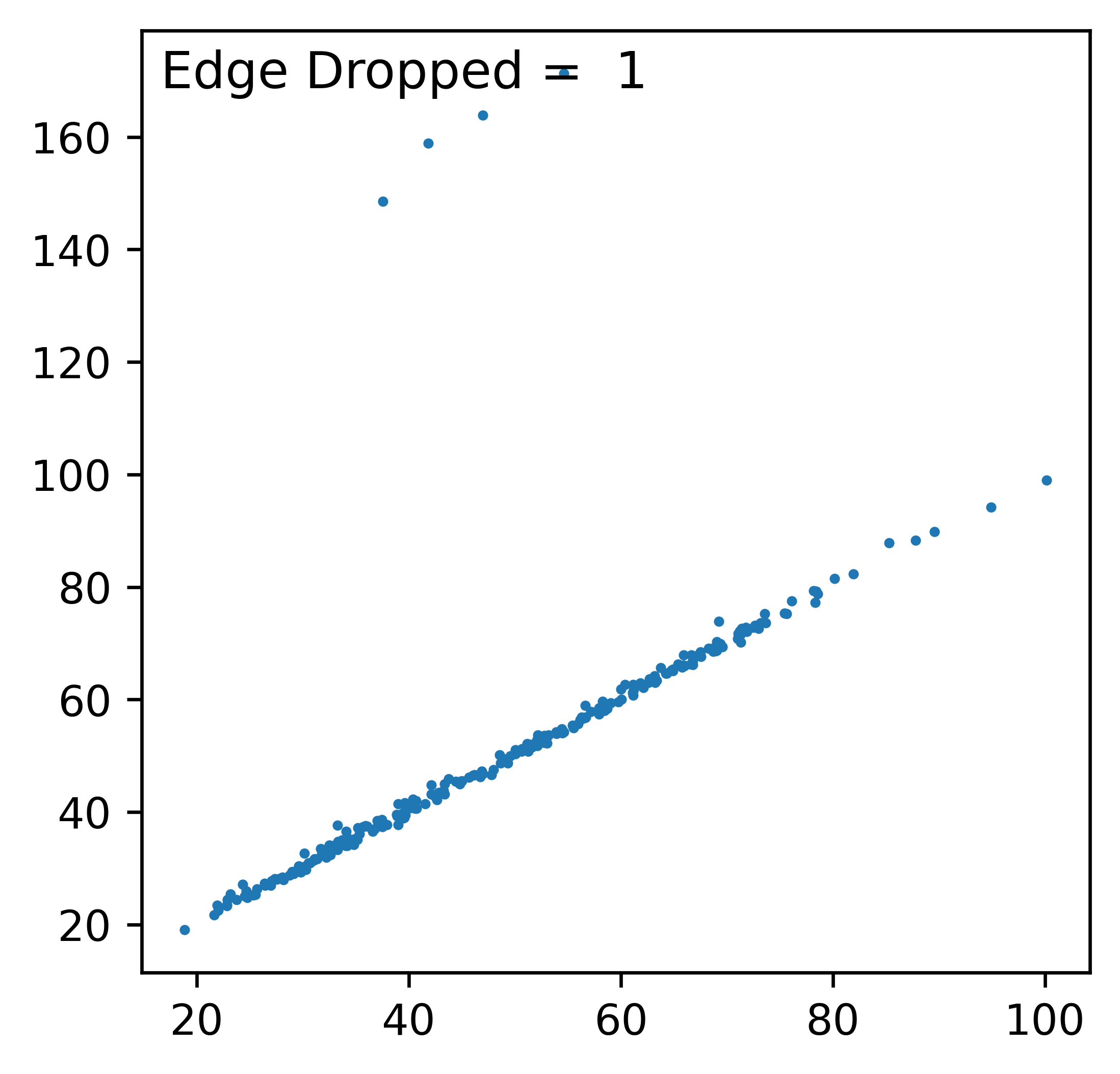}
    \end{minipage}
    \hfill
    \begin{minipage}{0.32\textwidth}
        \centering
        \includegraphics[width=\linewidth]{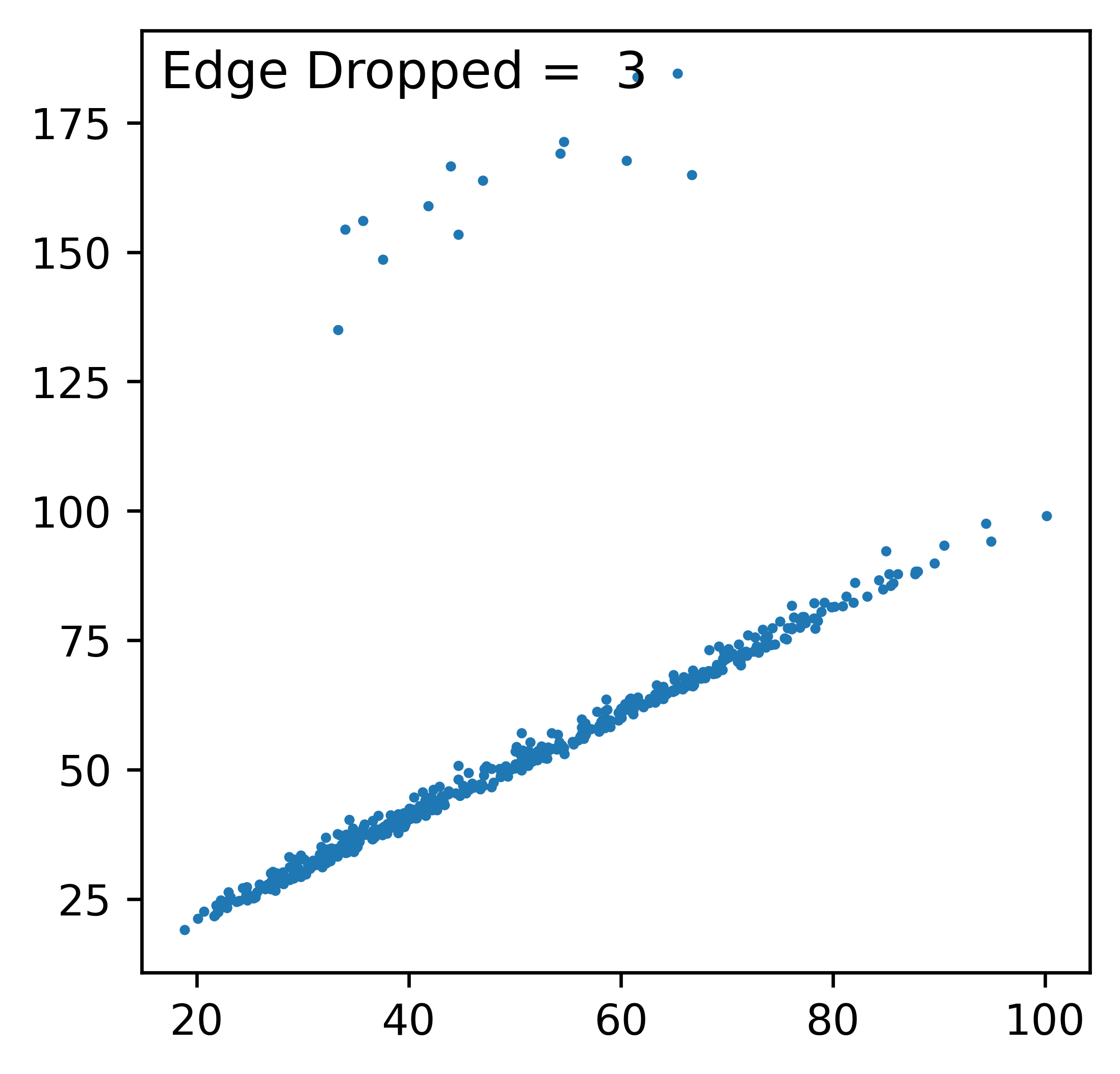}
    \end{minipage}
    \hfill
    \begin{minipage}{0.32\textwidth}
        \centering
        \includegraphics[width=\linewidth]{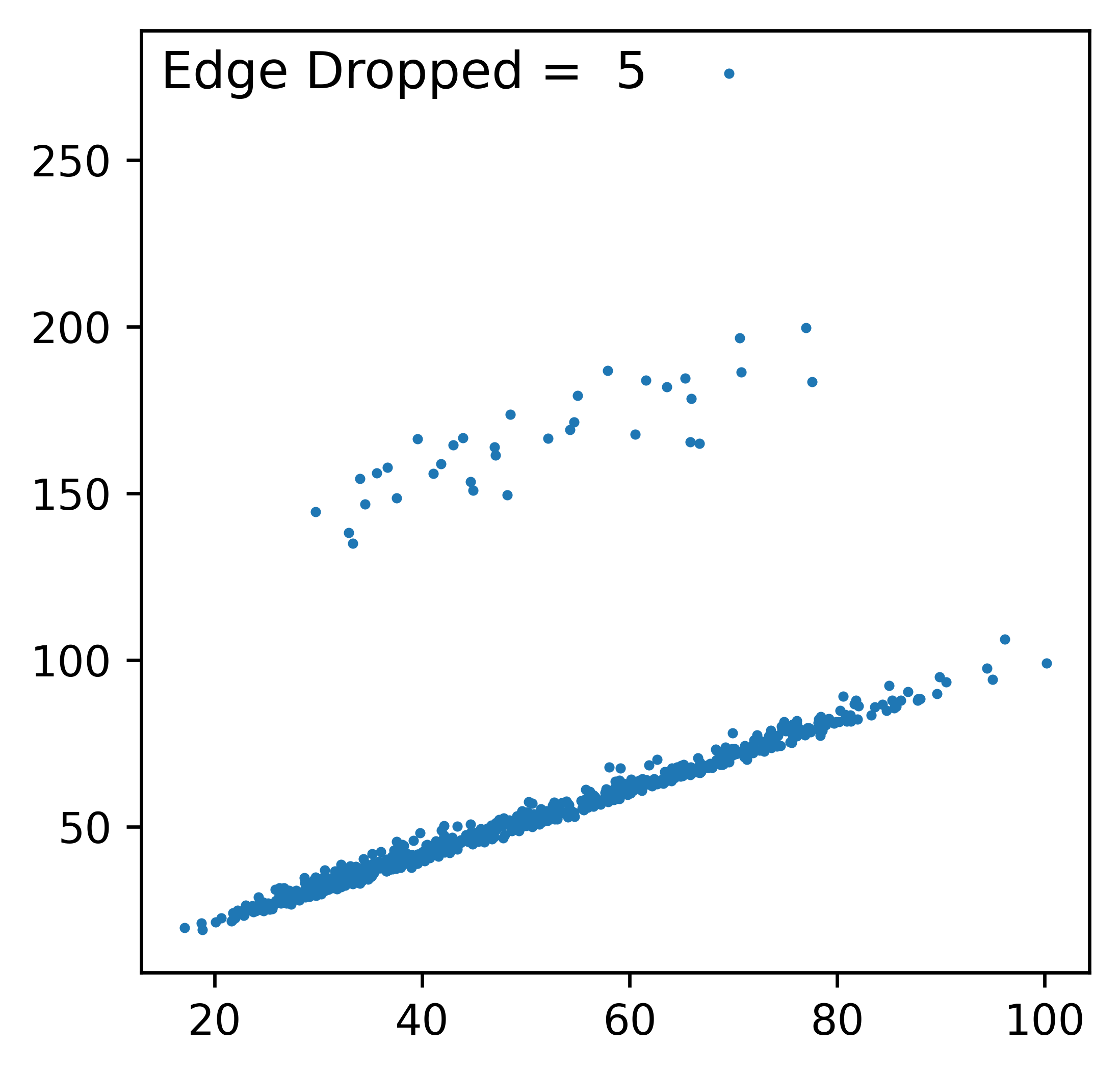}
    \end{minipage}
    \caption{GGD (x-axis) vs GGD after \textbf{Edge Drop} perturbation (y-axis) for smaller graphs.} \label{fig:per_5}
\end{figure}

\begin{figure}[H]
    \centering
    \begin{minipage}{0.32\textwidth}
        \centering
        \includegraphics[width=\linewidth]{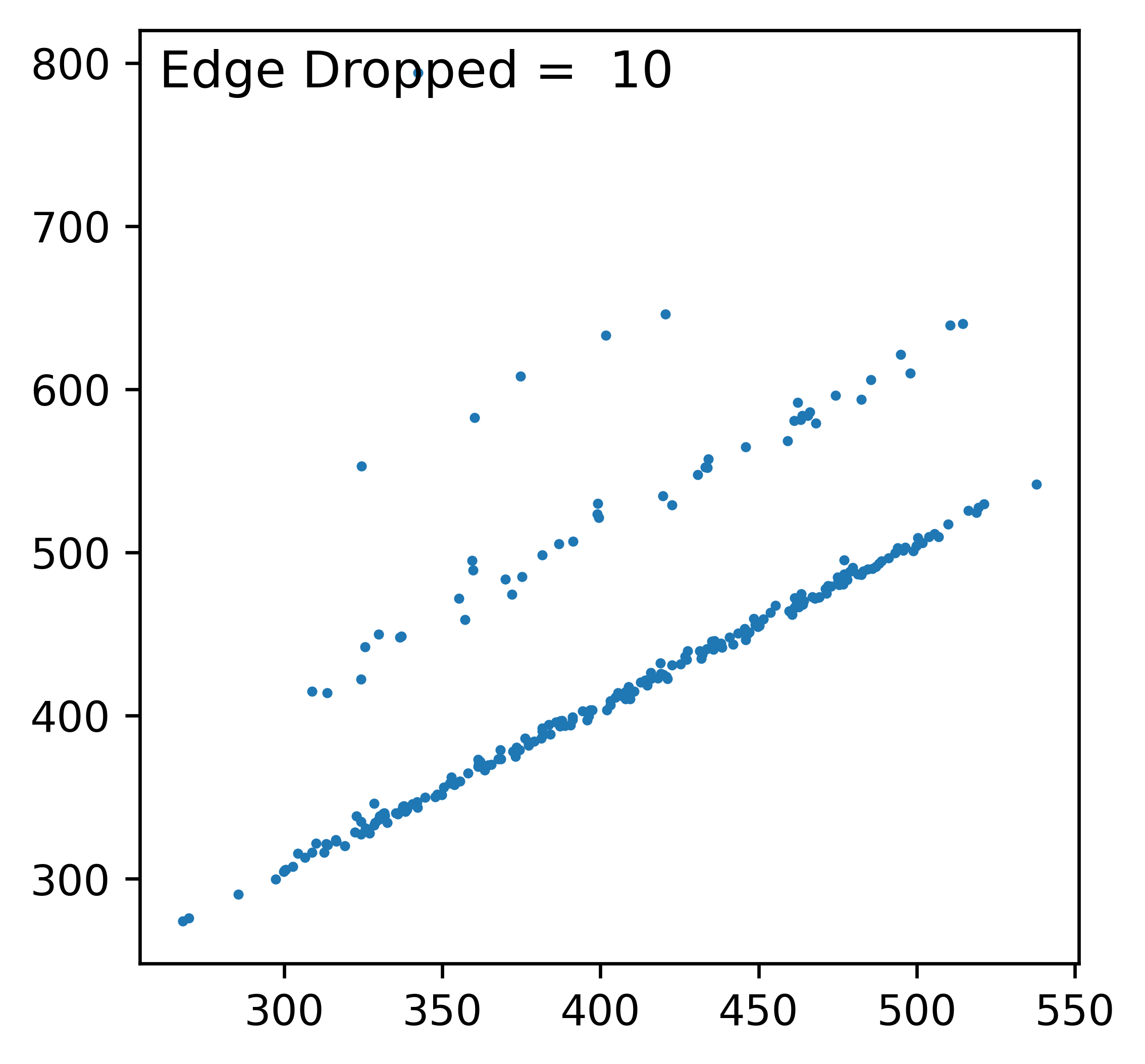}
    \end{minipage}
    \hfill
    \begin{minipage}{0.32\textwidth}
        \centering
        \includegraphics[width=\linewidth]{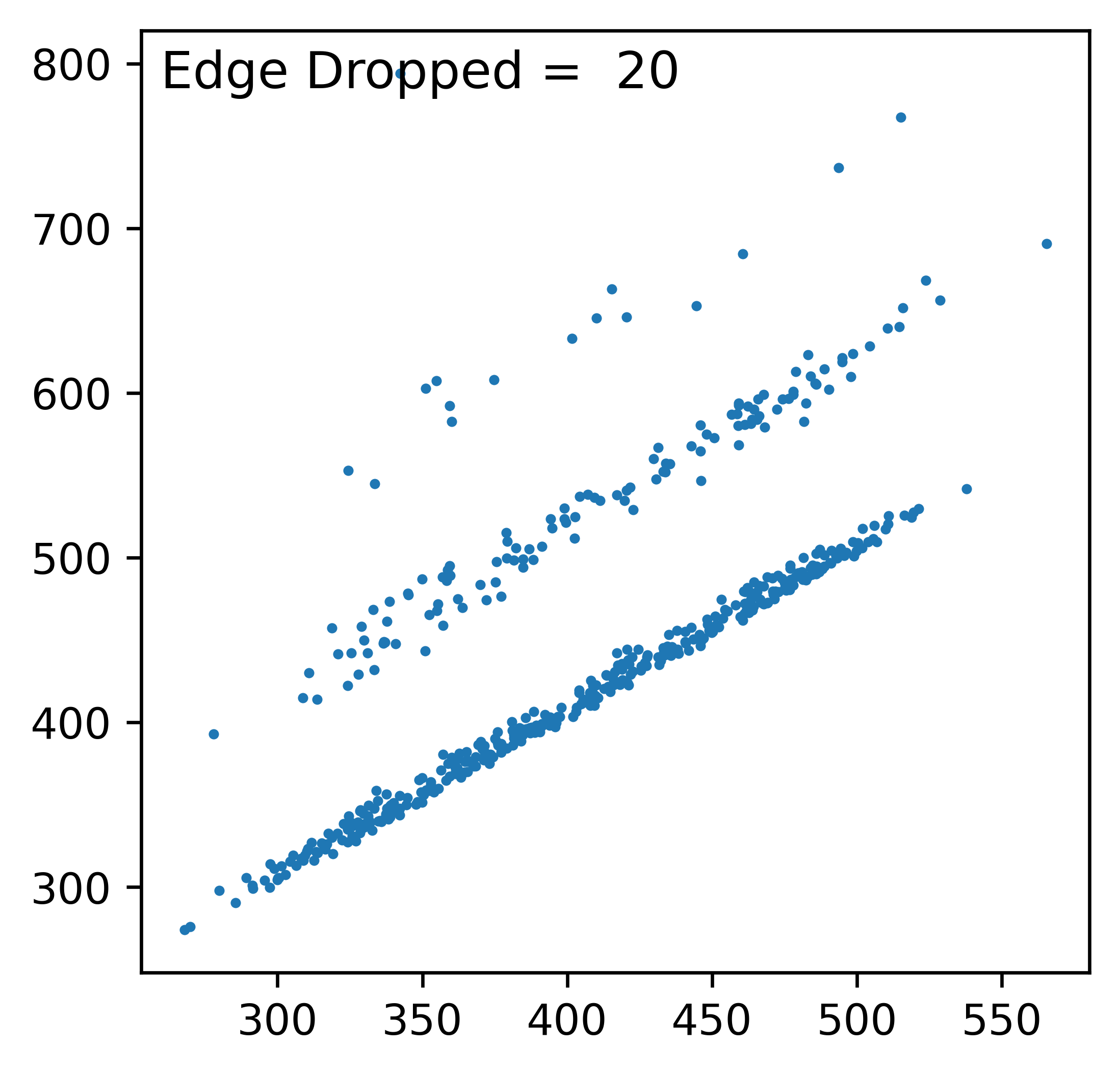}
    \end{minipage}
    \hfill
    \begin{minipage}{0.32\textwidth}
        \centering
        \includegraphics[width=\linewidth]{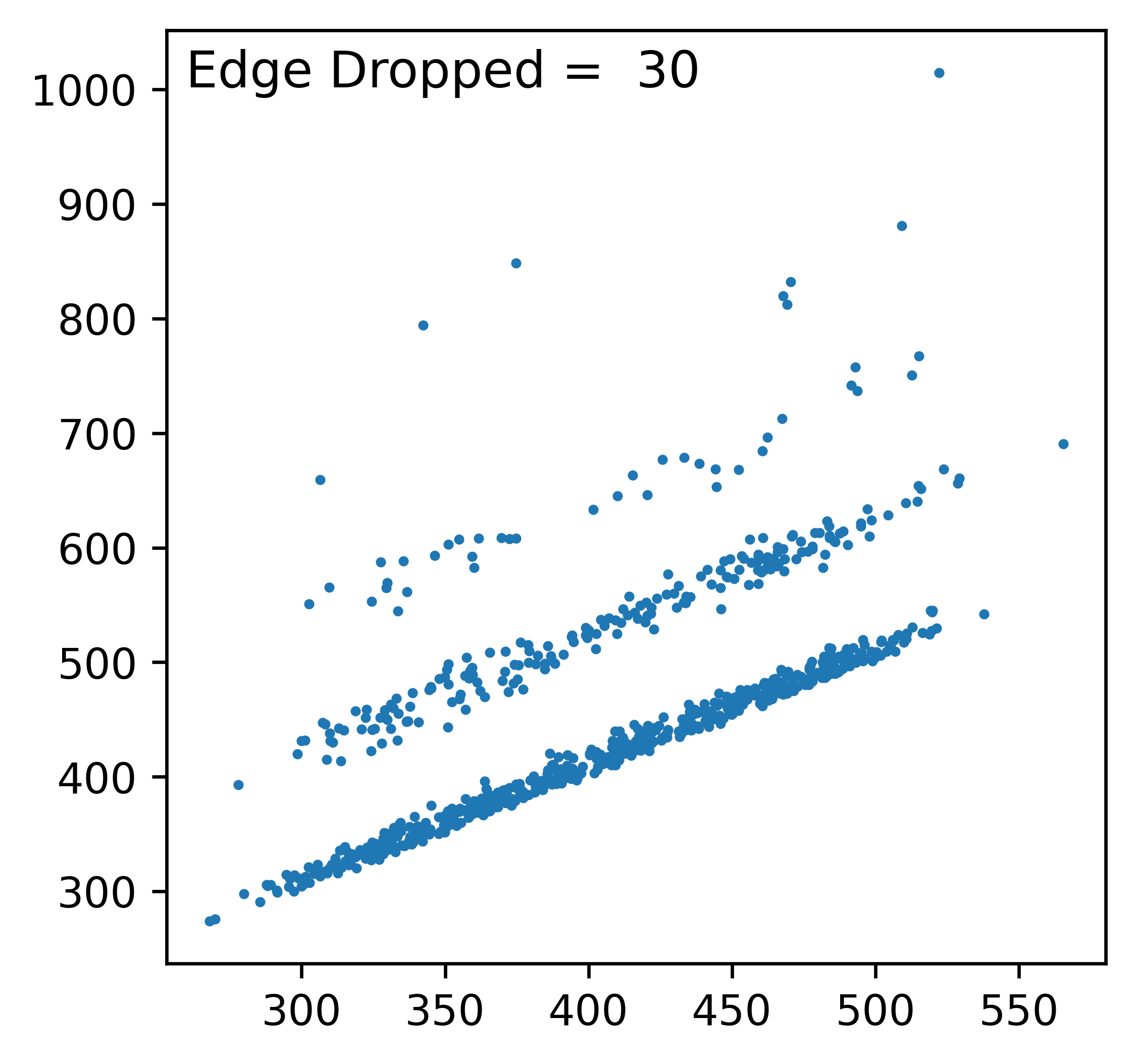}
    \end{minipage}
    \caption{GGD (x-axis) vs GGD after \textbf{Edge Drop} perturbation (y-axis) for larger graphs.} \label{fig:per_6}
\end{figure}

\begin{figure}[H]
    \centering
    \begin{minipage}{0.32\textwidth}
        \centering
        \includegraphics[width=\linewidth]{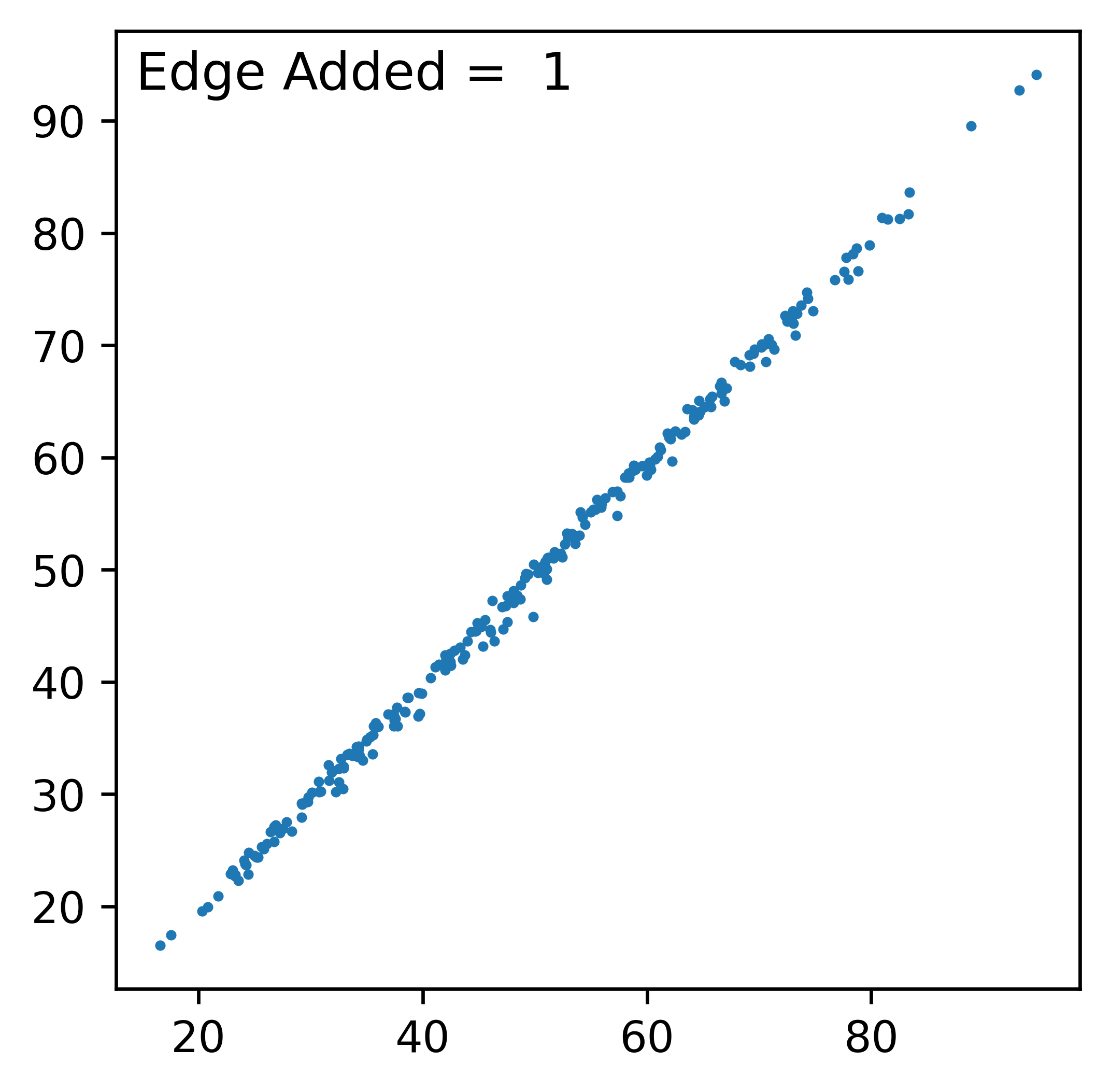}
    \end{minipage}
    \hfill
    \begin{minipage}{0.32\textwidth}
        \centering
        \includegraphics[width=\linewidth]{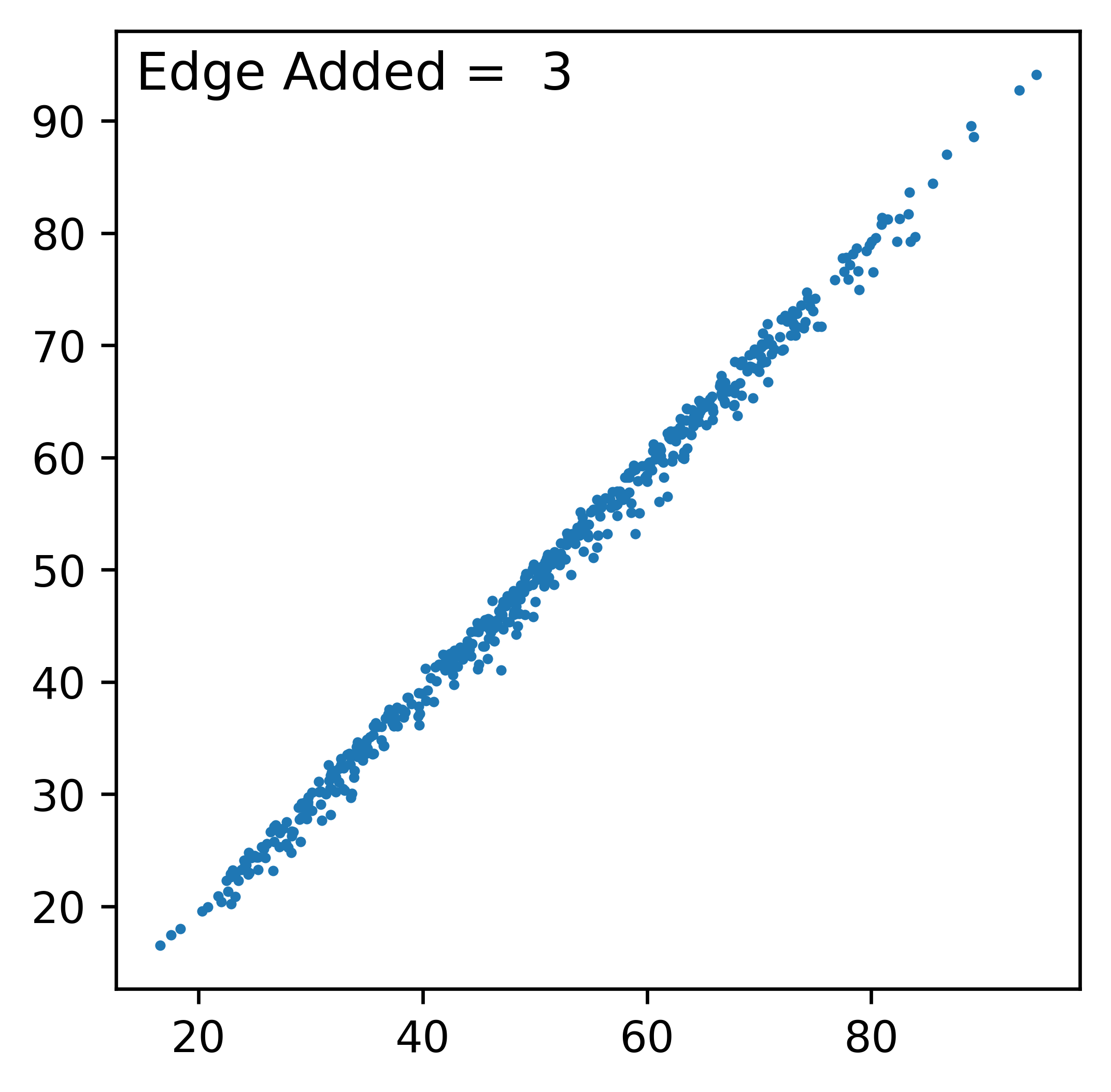}
    \end{minipage}
    \hfill
    \begin{minipage}{0.32\textwidth}
        \centering
        \includegraphics[width=\linewidth]{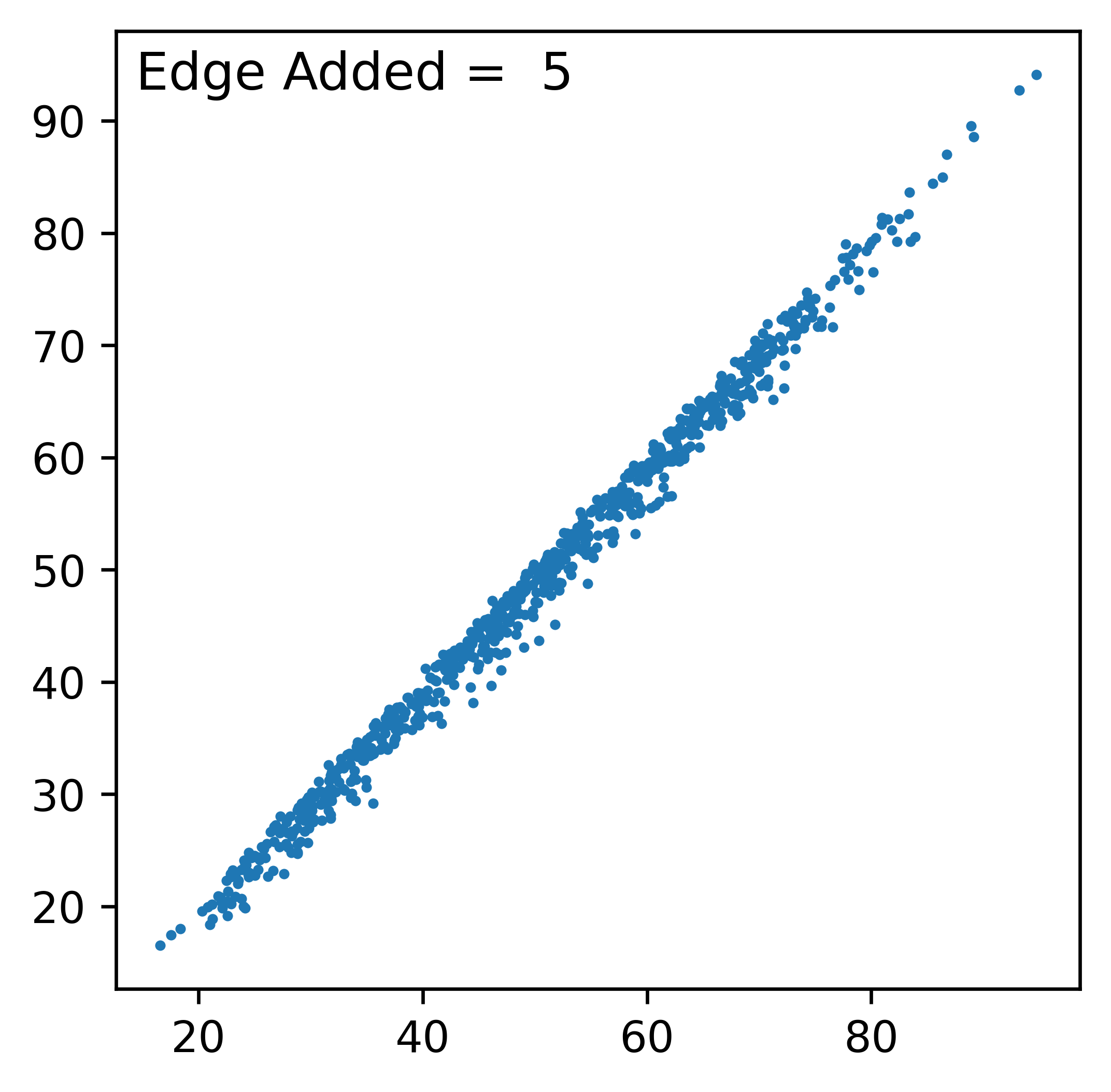}
    \end{minipage}
    \caption{GGD (x-axis) vs GGD after \textbf{Edge Addition} perturbation (y-axis) for smaller graphs.} \label{fig:per_7}
\end{figure}

\begin{figure}[H]
    \centering
    \begin{minipage}{0.32\textwidth}
        \centering
        \includegraphics[width=\linewidth]{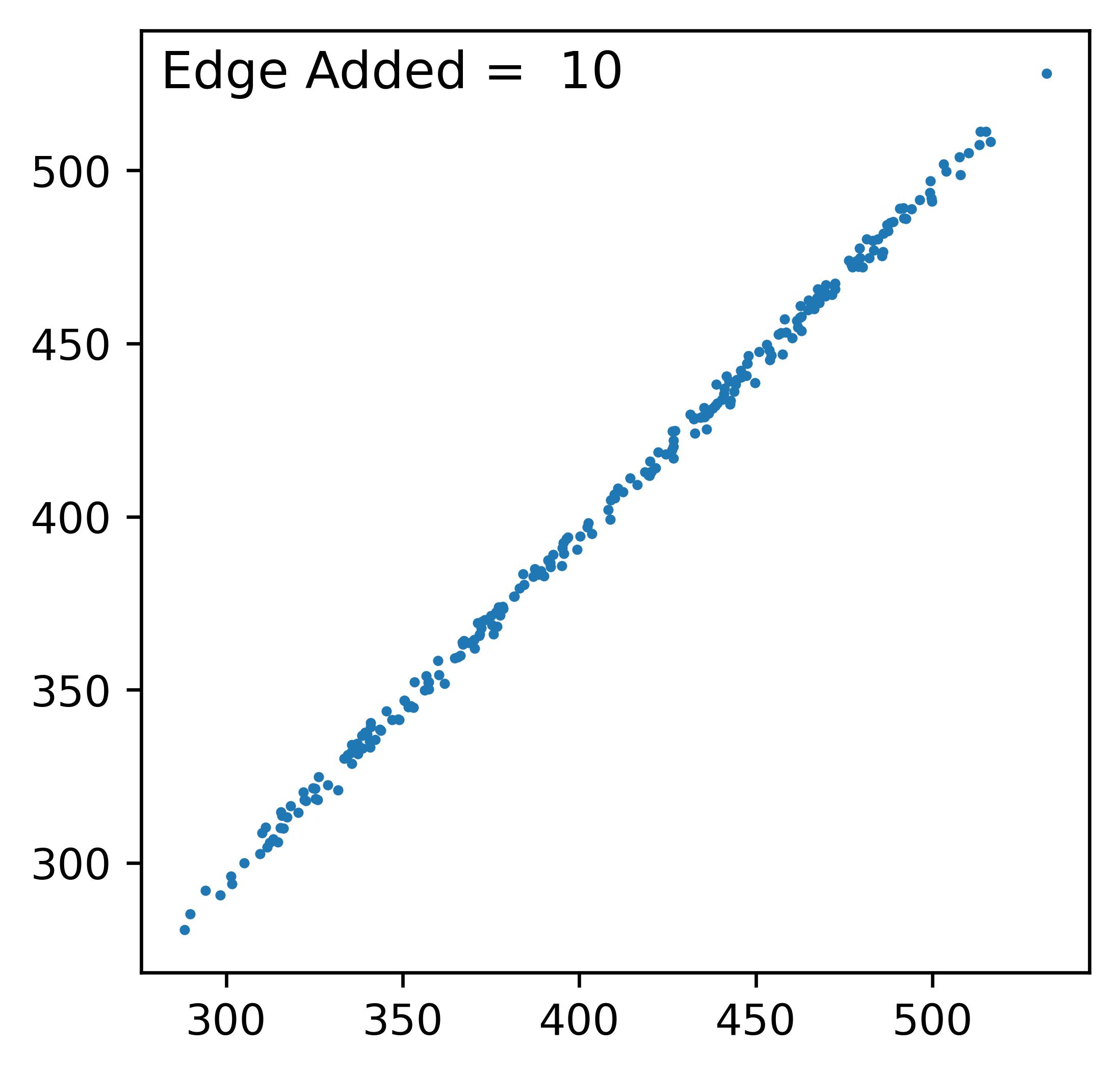}
    \end{minipage}
    \hfill
    \begin{minipage}{0.32\textwidth}
        \centering
        \includegraphics[width=\linewidth]{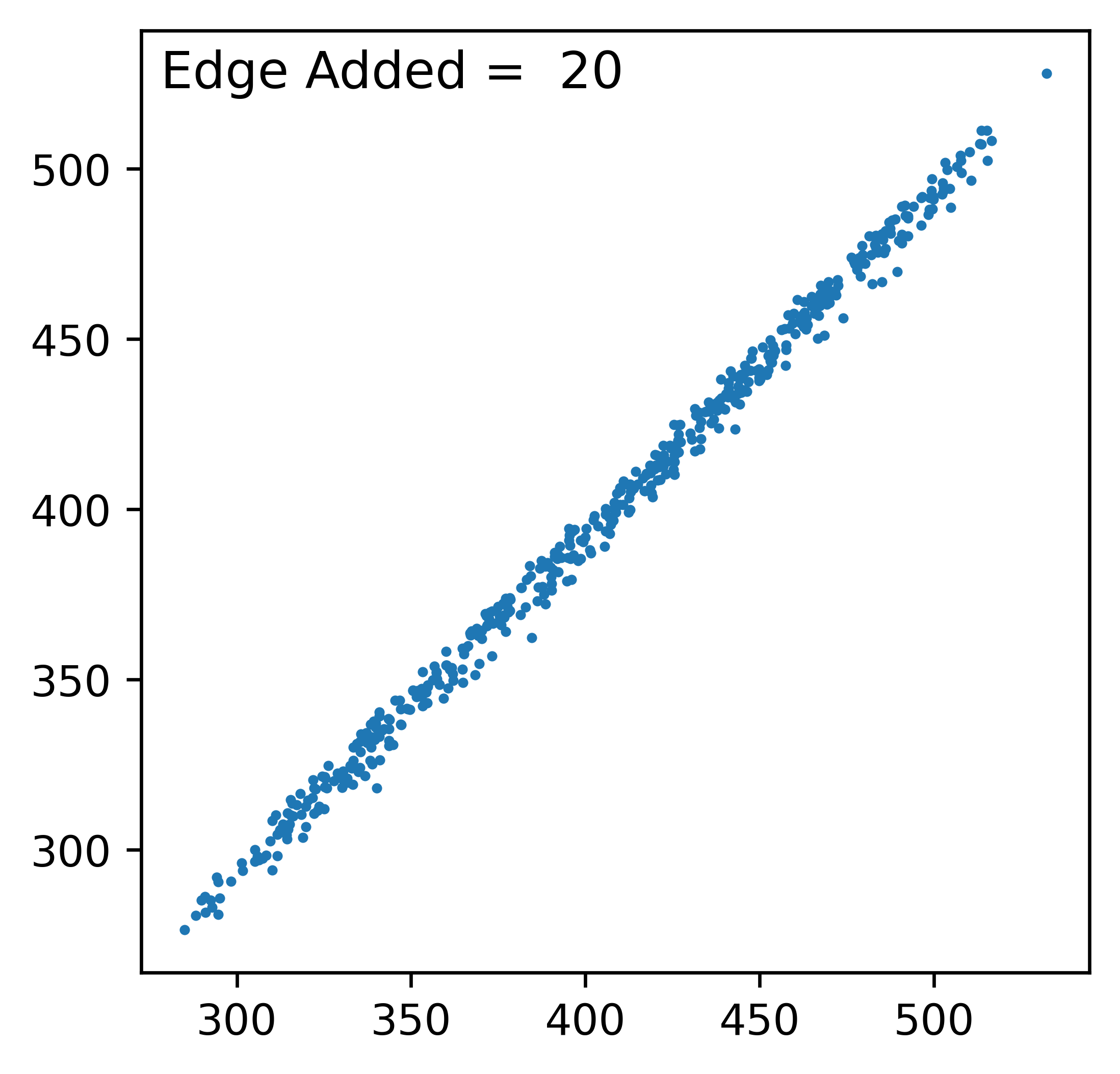}
    \end{minipage}
    \hfill
    \begin{minipage}{0.32\textwidth}
        \centering
        \includegraphics[width=\linewidth]{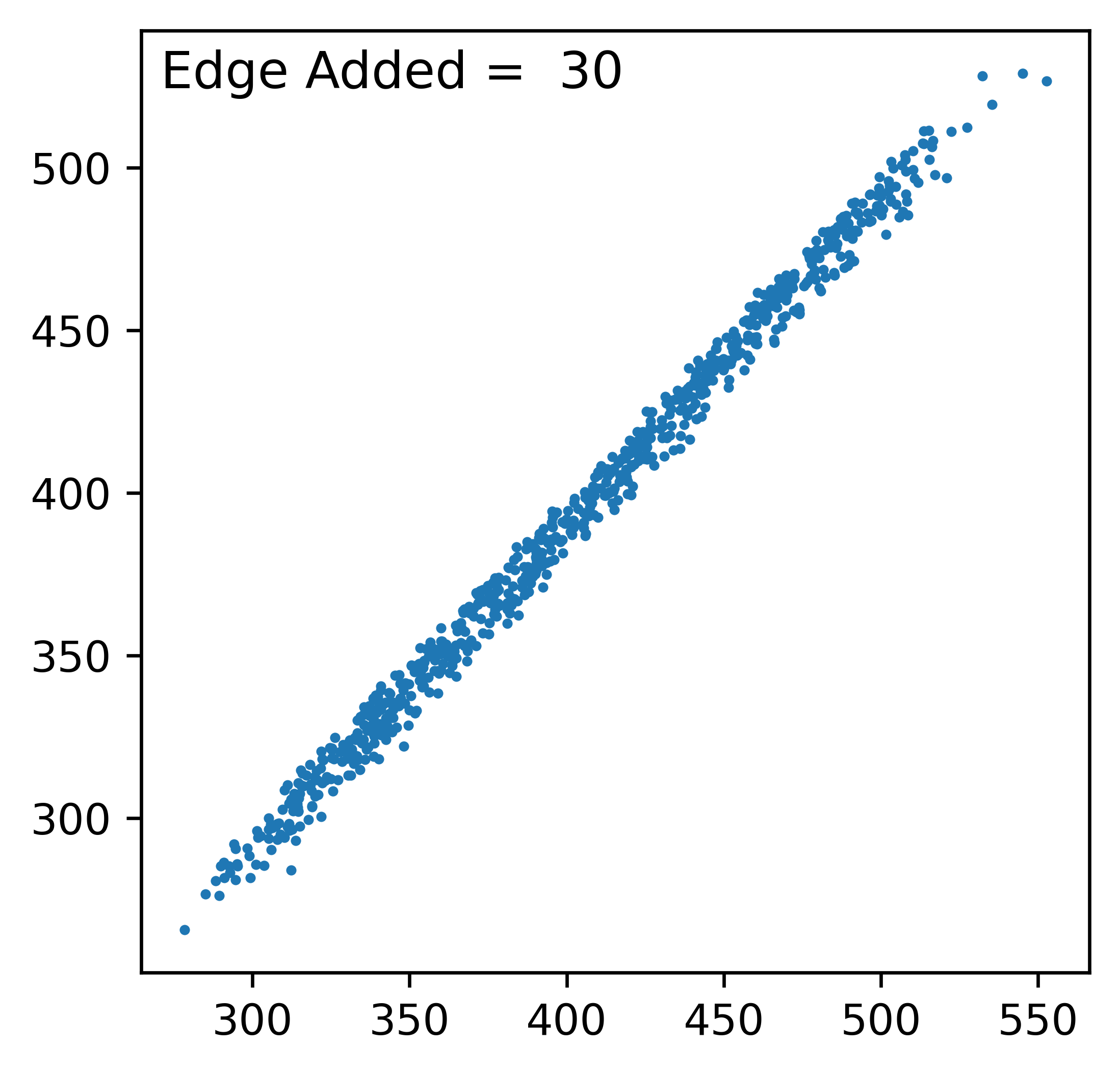}
    \end{minipage}
    \caption{GGD (x-axis) vs GGD after \textbf{Edge Addition} perturbation (y-axis) for larger graphs.} \label{fig:per_8}
\end{figure}

Our results show that node and edge additions typically have a minimal impact on the GGD values, suggesting that the metric is largely invariant to the inclusion of redundant elements. In contrast, node and edge drops exhibit noticeable effect. While a significant portion of these perturbations still yield GGD values comparable to the original graphs, some cases show some deviations. These deviations are associated with the removal of structurally important nodes or edges, which alters the graph topology. The quantitative results of these perturbation experiments are summarized Table \ref{table:perturb_small}, \ref{table:perturb_big} and Figure \ref{fig:per_1}, \ref{fig:per_2}, \ref{fig:per_3}, \ref{fig:per_4}, \ref{fig:per_5}, \ref{fig:per_6}, \ref{fig:per_7}, \ref{fig:per_8}.

% Due to the simplicity of GGD, it can be easily calculated between any two undirected graphs. We conducted additional experiments to show the effect of simple graph perturbations (node drop, edge drop, node addition, and edge addition) on both small and large graphs. From the results, we observe that node and edge addition have a minimal effect on the GGD values. In the case of node and edge drops, we observe some difference in the GGD values before and after the perturbations. However, the majority of perturbations result in almost the same GGD value. In many cases, node or edge drops significantly change the graph structure, leading to deviations in the GGD value, which is quite justified. These results are presented in Table \ref{table:perturb_small}, \ref{table:perturb_big} and Figure \ref{fig:per_1}, \ref{fig:per_2}, \ref{fig:per_3}, \ref{fig:per_4}, \ref{fig:per_5}, \ref{fig:per_6}, \ref{fig:per_7}, \ref{fig:per_8}.

\begin{table}
\centering
\caption{Correlation between GGD values before and after applying different perturbation methods on smaller graphs (20–50 nodes)}
 \label{table:perturb_small}
\begin{tblr}{
  cells = {c},
  hline{1,6} = {-}{0.08em},
  hline{2} = {-}{0.05em},
}
Amount of Perturbation & 1 & 2 & 3 & 4 & 5 \\
Nodes Dropped &         0.6934 &
0.6084 &
0.5896 &
0.5614 &
0.5368   \\
Nodes Added   &         0.9976	&
0.9959&
0.9943&
0.9922&
0.9900   \\
Edges Removed  &        0.8375 &
0.7890 &
0.6644 &
0.6537 &
0.6332   \\
Edges Added     &       0.9990&
0.9982&
0.9974&
0.9964&
0.9954   
\end{tblr}
\end{table}

% \begin{table}
% \centering
% \caption{Correlation between GGD values before and after applying different perturbation methods on larger graphs (200–500 nodes)}
% \begin{tblr}{
%   cells = {c},
%   hline{1,6} = {-}{0.08em},
%   hline{2} = {-}{0.05em},
% }
% Amount of Perturbation & 5 & 10 & 15 & 20 & 25 & 30 & 35 & 40 & 45 & 50\\
% Nodes Dropped &         0.8255 &
% 0.7734 &
% 0.6533 &
% 0.6116 &
% 0.5736 &
% 0.5563 &
% 0.5083 &
% 0.4729 &
% 0.4352 &
% 0.4162\\
% Nodes Added   &         0.9991&
% 0.9981&
% 0.9963&
% 0.9944&
% 0.9915&
% 0.9884&
% 0.9852&
% 0.9808&
% 0.9765&
% 0.9717\\
% Edges Removed  &        0.9137 &
% 0.8857&
% 0.8265&
% 0.7989&
% 0.7550 &
% 0.7226 &
% 0.7086&
% 0.6740&
% 0.6456&
% 0.6262\\
% Edges Added     &       .9996&
% .9992&
% .9986&
% .9979&
% .9971 &
% .9963&
% .9954&
% .9940&
% .9927&
% .9913
% \end{tblr}
% \end{table}

\begin{table} 
\centering
\caption{Correlation between GGD values before and after applying different perturbation methods on larger graphs (200–500 nodes)}
\label{table:perturb_big}
\begin{tblr}{
  cells = {c},
  hline{1,6} = {-}{0.08em},
  hline{2} = {-}{0.05em},
}
Amount of Perturbation & 5 & 10 & 15 & 20 & 25 & 30\\
Nodes Dropped &         0.8255 &
0.7734 &
0.6533 &
0.6116 &
0.5736 &
0.5563\\
Nodes Added   &         0.9991&
0.9981&
0.9963&
0.9944&
0.9915&
0.9884\\
Edges Removed  &        0.9137 &
0.8857&
0.8265&
0.7989&
0.7550 &
0.7226\\
Edges Added     &       0.9996&
0.9992&
0.9986&
0.9979&
0.9971 &
0.9963
\end{tblr}
\end{table}

% \begin{table}[H]
% \caption{Correlation between GGD values before and after node drop perturbation for smaller graphs (20-50 nodes)}
%     \centering
%     \begin{tabular}{c c c c c c}
                    
%         \hline
%         Nodes Removed & 1 & 2 & 3 & 4 & 5 \\
%         \hline
%         Correlation &
%         0.6934 &
% 0.6084 &
% 0.5896 &
% 0.5614 &
% 0.5368
%  \\
%         \hline
%     \end{tabular}
% \end{table}

% \begin{table}[H]
% \caption{Correlation between GGD values before and after node drop perturbation for larger graphs (200-500 nodes)}
%     \centering
%     \begin{tabular}{c c c c c c}
%         \hline
%         Nodes Removed & 5 & 10 & 15 & 20 & 25 \\
%         \hline
%         Correlation &
%         0.8255 &
% 0.7734 &
% 0.6533 &
% 0.6116 &
% 0.5736
%  \\
%         \hline
%         Nodes Removed & 30 & 35 & 40 & 45 & 50 \\
%         \hline
%         Correlation &
%         0.5563 &
% 0.5083 &
% 0.4729 &
% 0.4352 &
% 0.4162

%  \\
%  \hline
%     \end{tabular}
    
% \end{table}

% Table 3
% \begin{table}[H]
% \caption{Correlation between GGD values before and after node add perturbation for smaller graphs (20-50 nodes)}
%     \centering
%     \begin{tabular}{c c c c c c}
%         \hline
%         Nodes Added & 1 & 2 & 3 & 4 & 5 \\
%         \hline
%         Correlation &
%         0.9976	&
% 0.9959&
% 0.9943&
% 0.9922&
% 0.9900
%  \\
%         \hline
%     \end{tabular}
    
% \end{table}

% \begin{table}[H]
% \caption{Correlation between GGD values before and after node add perturbation for larger graphs (200-500 nodes)}
%     \centering
%     \begin{tabular}{c c c c c c}
%         \hline
%         Nodes Added & 5 & 10 & 15 & 20 & 25 \\
%         \hline
%         Correlation &
%         0.9991&
% 0.9981&
% 0.9963&
% 0.9944&
% 0.9915
%  \\
%         \hline
%         Nodes Added & 30 & 35 & 40 & 45 & 50 \\
%         \hline
%         Correlation &
% 0.9884&
% 0.9852&
% 0.9808&
% 0.9765&
% 0.9717
%  \\
%  \hline
%     \end{tabular}
    
% \end{table}

% Table 5
% \begin{table}[H]
% \caption{Correlation between GGD values before and after edge drop perturbation for smaller graphs (20-50 nodes)}
%     \centering
%     \begin{tabular}{c c c c c c}
%         \hline
%         Edges Dropped & 1 & 2 & 3 & 4 & 5 \\
%         \hline
%         Correlation &
%         0.8375 &
% 0.7890 &
% 0.6644 &
% 0.6537 &
% 0.6332
%  \\
%         \hline
%     \end{tabular}
    
% \end{table}

% \begin{table}[H]
% \caption{Correlation between GGD values before and after edge drop perturbation for larger graphs (200-500 nodes)}
%     \centering
%     \begin{tabular}{c c c c c c}
%         \hline
%         Edges Dropped & 5 & 10 & 15 & 20 & 25 \\
%         \hline
%         Correlation &
% 0.9137 &
% 0.8857&
% 0.8265&
% 0.7989&
% 0.7550

%  \\
%         \hline
%         Edges Dropped & 30 & 35 & 40 & 45 & 50 \\
%         \hline
%         Correlation &
% 0.7226 &
% 0.7086&
% 0.6740&
% 0.6456&
% 0.6262

%  \\
%  \hline
%     \end{tabular}
    
% \end{table}

% Table 5
% \begin{table}[H]
% \caption{Correlation between GGD values before and after edge addition perturbation for smaller graphs (20-50 nodes)}
%     \centering
%     \begin{tabular}{c c c c c c}
%         \hline
%         Edges Added & 1 & 2 & 3 & 4 & 5 \\
%         \hline
%         Correlation &
% 0.9990&
% 0.9982&
% 0.9974&
% 0.9964&
% 0.9954

%  \\
%         \hline
%     \end{tabular}
    
% \end{table}

% \begin{table}[h]
% \caption{Correlation between GGD values before and after edge addition perturbation for larger graphs (200-500 nodes)}
%     \centering
%     \begin{tabular}{c c c c c c}
%         \hline
%         Edges Added & 5 & 10 & 15 & 20 & 25 \\
%         \hline
%         Correlation &
% .9996&
% .9992&
% .9986&
% .9979&
% .9971

%  \\
%         \hline
%         Edges Added & 30 & 35 & 40 & 45 & 50 \\
%         \hline
%         Correlation &
% .9963&
% .9954&
% .9940&
% .9927&
% .9913
%  \\
%  \hline
%     \end{tabular}
    
% \end{table}

\subsection{Graph formation from Dataset using Probabilistic Graphical Models} \label{app:datasettograph}

We take similiar type of datasets such as MNIST, Fashion-MNIST, KMNIST, and USPS \citep{otdd} and convert them into connected graphs. To construct these graph structures, we use Probabilistic Graphical Models (PGMs), also known as Markov Random Fields (MRFs) \citep{cheng2024sagman}. PGMs are powerful tools in machine learning and statistical physics for representing complex systems with intricate dependency structures \citep{roy2009learning}. They encode the conditional dependencies between random variables through an undirected graph structure. Recent studies have shown that the graph structure learned via PGMs can exhibit resistance distances that encode the Euclidean distances between their corresponding data samples \citep{feng2021sgl}.  

We create a Feature Matrix (FM) $U$ from the dataset, where each row represents a data sample, and the row $U_p$ itself serves as the feature vector of that sample $p$, on this context- the pixel values. A dense $k$-nearest neighbor (k-NN) graph $G_{dense}$ is initially constructed using the FM. To obtain the final dataset graph $G_{dataset}$, spectral sparsification is applied by solving the convex optimization problem \citep{cheng2024sagman}:  
\begin{equation} \label{eq:pgm}
    \max _{\Theta} \quad F(\Theta)=\log \operatorname{det}(\Theta)-\frac{1}{k} \operatorname{Tr}\left(U^{\top} \Theta U\right)
\end{equation}  
where \( \Theta=L+\frac{1}{\sigma^2} I \). Here, \( L \) is the graph Laplacian, \( \operatorname{Tr}(\cdot) \) denotes the trace of a matrix, \( I \) is the identity matrix, and \( \sigma^2 > 0 \) represents the prior feature variance. To solve this the following lemma is used: 

\begin{lemma}  
Maximizing the objective function in Equation \ref{eq:pgm} can be achieved in nearly-linear time via the following edge pruning strategy equivalent to spectral sparsification of the initial dense
nearest-neighbor graph. Specifically, edges with small distance ratios  
\begin{equation}
\rho_{p, q}=\frac{R_{\mathrm{eff}}(p, q)}{d_{\mathrm{euc}}(p, q)}=w_{p, q} \cdot R_{\mathrm{eff}}(p, q)
\end{equation}
are pruned, where $R_{\mathrm{eff}}(p, q)$ denotes the effective resistance distance between nodes $p$ and $q$, $d_{\mathrm{euc}}(p, q) = \left\|U_p-U_q\right\|_2^2$ represents the data distance between the feature of nodes $p$ and $q$, and $w_{p, q}=\frac{1}{d_{\mathrm{euc }}(p, q)} $ is the weight of edge $(p, q)$ \citep{cheng2024sagman}.
\end{lemma}

Computing the edge sampling probability $\rho_{p, q}$ for each edge $(p, q)$ becomes computationally expensive for large graphs. To address this, an improved algorithm using a low-resistance-diameter (LRD) decomposition is proposed, extending the short-cycle decomposition \citep{chu2020graph} to weighted graphs. The method efficiently computes effective resistance to partition the graph into clusters, thereby enhancing the sparsification process. This results in a low-dimensional graph $G_{dataset}$ that retains important structural properties while reducing dimensionality \citep{cheng2024sagman}.

\subsection{Experimental Setup} \label{exp_set}

To evaluate the performance of the Graph Geodesic Distance (GGD) metric, we utilized graph datasets from the TUDataset collection \citep{morris2020tudataset}. For small graphs, we used datasets like MUTAG and BZR, and for larger graphs, we selected PC-3H and SW-620H, which present more sizable networks. Detailed information about the datasets used is provided in Table \ref{dataset_description}.

\begin{table}[h]
    \centering
    \caption{Brief description of graph datasets used.}
    \label{dataset_description}
    \begin{tabular}{l  c c c}
        \hline
        Dataset name & Number of graphs & Average number of nodes & Average number of edges \\
        % Name & Graphs & of Nodes & of Edges\\
        \hline
        MUTAG & 188 & 17.93 & 19.79 \\
        PC-3H & 27509 & 47.20 & 49.33 \\
        SW-620H & 40532 & 26.06 & 28.09 \\
        BZR & 405 & 35.75 & 38.36 \\
        \hline
    \end{tabular}
\end{table}

While Classification tasks, each dataset was split into 90\% training and 10\% testing sets to ensure an unbiased evaluation process. When assessing the correlation with GNN, we trained a three-layer GIN with 90\% of all graphs from MUTAG and validated with the rest 10\%. For the performance evaluation using graphs with partial node features, we took each dataset with node features and randomly removed a certain portion of features.

All experiments have been evaluated on a laptop with an Apple M1 chipset, featuring an eight-core CPU and a seven-core GPU.

\end{document}